\documentclass[journal]{IEEEtran}
\usepackage{cite}

\usepackage[pdftex]{graphicx}

\usepackage{amsmath,amssymb,amsfonts}

\usepackage{algorithmic}
\usepackage{float}
 \usepackage{multirow}
\usepackage{comment}
\usepackage{url}

\usepackage{xcolor}
\usepackage{caption}
\usepackage{subcaption}

\begin{document}
\title{Physical Adversarial Attacks for Surveillance: A Survey}

\author{Kien~Nguyen,
        ~Tharindu~Fernando,
        ~Clinton~Fookes,
        ~Sridha~Sridharan
\thanks{K. Nguyen, T. Fernando, C. Fookes, S. Sridharan are with 
Queensland University of Technology, QLD, 4000 Australia. 
E-mails: \{k.nguyenthanh,t.warnakulasuriya,c.fookes,s.sridharan\}@qut.edu.au.}
}

\markboth{IEEE Transactions on Neural Networks and Learning Systems}%
{Nguyen \MakeLowercase{\textit{et al.}}: Physical Adversarial Attacks for Surveillance: A Survey}
%

\maketitle

\begin{abstract}
Modern automated surveillance techniques are heavily reliant on deep learning methods. Despite the superior performance, these learning systems are inherently vulnerable to adversarial attacks - maliciously crafted inputs that are designed to mislead, or trick, models into making incorrect predictions. An adversary can physically change their appearance by wearing adversarial t-shirts, glasses, or hats or by specific behavior, to potentially avoid various forms of detection, tracking and recognition of surveillance systems; and obtain unauthorized access to secure properties and assets. This poses a severe threat to the security and safety of modern surveillance systems. This paper reviews recent attempts and findings in learning and designing physical adversarial attacks for surveillance applications. In particular, we propose a framework to analyze physical adversarial attacks and provide a comprehensive survey of physical adversarial attacks on four key surveillance tasks: detection, identification, tracking, and action recognition under this framework. Furthermore, we review and analyze strategies to defend against the physical adversarial attacks and the methods for evaluating the strengths of the defense. The insights in this paper present an important step in building resilience within surveillance systems to physical adversarial attacks.  
\end{abstract}

\begin{IEEEkeywords}
Physical Adversarial Attacks, Counter Biometric Surveillance, Adversarial Vulnerability, Safety and Security, Surveillance Systems, Adversarial Defense 
\end{IEEEkeywords}

%
\IEEEpeerreviewmaketitle

\section{Introduction}
\label{sec:introduction}
\IEEEPARstart{T}{he} proliferation of surveillance cameras across the world, with a projection of 1 billion surveillance cameras installed worldwide in 2021, gives law enforcement agencies the ability to quickly, accurately, uniquely detect, track and identify individuals through their physiological and behavioral traits. This capacity has been boosted by recent breakthroughs in computer vision and deep learning, leading to faster and more precise automated analysis of surveillance footage, and enabling highly accurate surveillance techniques.

Within the areas of automated surveillance and identification techniques, good performance has been achieved in key surveillance tasks using highly parameterized deep learning neural networks. 

\begin{itemize}
    \item \textit{Human Detection:} the state of the art human detectors, e.g. YOLO \cite{YOLOv4,ScaledYOLO} and Faster R-CNN \cite{FasterRCNN}, are able to detect persons on par with or even better than humans. For example, YOLO demonstrates superior detection accuracy on the challenging MS COCO object detection challenge \cite{MSCOCO} with an Average Precision (AP) of 56.0\%. 
    \item \textit{Human Tracking:} the state of the art human trackers, e.g. SiamRPN++ \cite{SiamRPNpp} and Ocean \cite{Ocean}, are able to accurately track moving humans across a wide range of background and occlusion. For example, Ocean is the most accurate visual object tracker on the VOT2018 visual object tracking dataset with an Expected Average Overlap (EAO) of 0.467.
    \item \textit{Human Identification:} the state of the art face recognisers, e.g. ArcFace \cite{ArcFace} and SV-X-Softmax \cite{SVXSoftmax}, achieve near optimal verification and identification on large scale datasets such as more than 98\% for the MegaFace dataset \cite{MegaFace} and 99\% for the LFW dataset \cite{LFW}. 
    \item \textit{Human Action Recognition:} the state of the art human action recognizers, e.g. I3D \cite{I3D} and R2+1D-BERT \cite{R21DBERT}, are able to accurately detect and recognize what action a person is performing. For example, R2+1D-BERT is the most accurate action recognizer on the UCF101 action recognition dataset \cite{UCF101} with an accuracy of 98.69\%.
\end{itemize}

Surveillance systems have always been the target for attackers to bypass and obtain unauthorized access to the secure properties or assets. For example, military camouflage has been long used by armed forces to decrease the danger of being targeted for an attack \cite{CamouflageHistory}. With the development of automatic computer-based detection and recognition systems, more sophisticated techniques have also been attempted to interfere and mislead the vision-based surveillance tasks by wearing face masks \cite{3DFaceMask}, LED glasses \cite{LEDglasses}, or even putting on special make-ups \cite{CVDazzle}. These human-designed accessories have managed to mislead modern human detection, face detection and face recognition to some extent.

Recently, a new form of counter surveillance has emerged via adversarial attacks, which exploit the vulnerability of modern deep learning and machine learning models to mislead or trick them into making incorrect predictions. These techniques employ optimization algorithms to search for adversarial examples - crafted inputs with well-designed adversarial perturbations which can thwart deep/machine learning models \cite{AdvSurvey}. Adversarial attacks can be either \emph{digital} or \emph{physical}. Digital attacks usually manipulate an image or video by adding perturbations which can be visually difficult to discern to manipulate the system output. This strategy can not be applied to physical attacks because the impact of the imperceptible perturbations quickly diminishes due to physical imaging conditions. In contrast, \emph{physical} attacks seek to manipulate the physical environment by purposefully creating or altering real-world objects/accessories to incorporate a physical manifestation of an adversarial perturbation or patch. Examples of adversarial accessories are: (1) an \textit{adversarial t-shirt} to thwart human detection \cite{UPC,AdvTshirtFB,AdvTShirtIBM}, (2) an \textit{adversarial poster} to thwart human tracking \cite{AdvTracking}, (3) an \textit{adversarial pair of glasses} to thwart face recognition \cite{advGlasses}, and (4) an \textit{adversarial flickering LED} to thwart human action recognition \cite{PhysicalAdvAction}.

Learning and designing these adversarial accessories or behaviors to be effective in the physical world of surveillance is more challenging than digital attacks due to the diversity of real-world imaging conditions, the dynamics of adversaries, and the multi-camera nature of surveillance systems that adversarial artifacts have to survive.

\vspace{3px}
\noindent \textbf{Scope:} Considering the security and safety of the existing surveillance systems and the real-world threat that physical adversarial attacks pose to proper functionality of those systems, we limit our discussion to physical adversarial attacks and human related surveillance tasks. We believe the specific scope of this study allows us to provide an in-depth and systematic analysis of physical adversarial attack and defense mechanisms under numerous practical surveillance tasks, associated challenges with respect to the surveillance tasks, the limitation of existing attack and defense mechanisms and the scope for future research, using a unified framework. 

\vspace{3px}
\noindent \textbf{Our Contributions:}
This paper investigates recent attempts and findings in learning and designing physical adversarial attacks for the surveillance application. We propose a framework to understand the state of the art approaches in generating and designing physical adversarial attacks. Under this framework, we provide a comprehensive survey of modern physical adversarial attacks on four key surveillance tasks: detection, identification, tracking and action recognition. In addition, this paper also discusses interesting development of physical adversarial attacks beyond the popular visible domain, i.e., infrared, LiDAR and multispectral spectra. We also review and discuss the methods introduced for defending human detection attacks, human identification attacks and human action recognition attacks. Based on the review and analysis, we identify the challenges and provide our perspective on the next steps for this research topic.

Although there exists several recent survey articles \cite{wei2022physically, yuan2019adversarial, akhtar2018threat, wiyatno2019adversarial} on adversarial attacks, to the best of our knowledge there is no comprehensive review of physical adversarial attack methods that are related to human surveillance. The unique and specific perspective enables a systematic review of surveillance tasks, ranging from detection, identification, tracking to action recognition. Specifically, for each surveillance task, our paper discusses the state-of-the-art approaches that have been introduced for that particular surveillance task,  then summaries numerous physical adversarial attacks approaches have managed to fool these state-of-the-art approaches, and the ones that are still standing strong. Utilizing the proposed framework as the guideline we outline the similarities and differences between the physical adversarial attack methods.

Moreover, this paper provides a comprehensive overview of existing adversarial defense mechanism that can defend a surveillance system against a physical adversarial attack. Our analysis is not limited to human surveillance in the visible spectrum but also spans to infrared, LiDAR and multispectral spectra. Finally, this review details the challenges of generating successful physical adversarial attacks within the surveillance perspective, the limitations of the state-of-the-art adversarial attack and defense methods and key future research directions.

\vspace{3px}
\noindent \textbf{Organization:}
The rest of this paper is structured as follows. Section~\ref{sec:AdvAttacks} introduces adversarial attacks including both digital and physical-world attacks, and introduces the proposed framework that we utilize to analyze these attack methods. Section~\ref{sec:humandesigned} reviews human-design adversarial attacks which are conventional methods that attempt to thwart the surveillance process by interfering the
imaging process or obfuscate the subjects body, face, etc. with artifacts. Section~\ref{sec:machinelearned} reviews modern physical adversarial attacks which are learned by deep learning. Specifically, attacks on human detection, identification, tracking and action recognition are discussed. Furthermore, we analyze methods proposed to thwart human surveillance systems that operates beyond the visible spectrum. In addition evaluation methods that evaluates the success of the attack Section~\ref{sec:Defence} reviews strategies to defend against the physical adversarial attacks and the methods for evaluating the strength of the defense.
Section~\ref{sec:Conclusions} concludes the survey.

\section{Practical Adversarial Attacks in Human Surveillance Framework}
\label{sec:AdvAttacks}
In this section we illustrate the method for generating practical and effective adversarial attacks to thwart human surveillance. We first introduce digital adversarial attack methods and discusses their intriguing properties that make them potentially effective in the physical domain. We then compare and contrast physical and digital adversarial attacks. We focus on the key factors that make physical adversarial attacks challenging. Then we introduce the proposed framework for analyzing physical adversarial attacks. 

\vspace{-9px}
\subsection{Digital adversarial attacks}
In 2014, Szegedy \emph{et al.} \cite{AdvOriginal} first reported an intriguing property of neural networks whereby imperceptible perturbations added to an input can result in a trained deep network making incorrect predictions. Critically, perturbations that lead to erroneous classifications can be systematically found in all machine learning algorithms and neural networks. Such a model can be represented as a function, $h$, characterized by a parameter set, $\theta$, that predicts an output, $y$ from the input, $x$ such that $y=h_{\theta}(x)$. A minimal perturbation, $\sigma$, can be learned by an adversarial algorithm to alter the input and result in an incorrect prediction,
\vspace{-9px}
\begin{align}
\hat{y} = h_{\theta}( & x+\sigma)    \\
\textrm{such that: } & \hat{y} \neq y  \textrm{  and:} ||\sigma||_p < \eta
\end{align}
where the $L_p$ distance metric of $\sigma$ is constrained by a predefined value, $\eta$. Dangerously, for humans, the adversarial sample, $\hat{x} = x+\sigma$, can be indistinguishable from the original input, $x$, or in other words the perturbation is imperceptible to human naked eyes. The famous example of digital modification of a panda image with a small perturbation to mislead the classifier to classify the perturbed image as another class, i.e. gibbon, with a high confidence. 
The perturbations can be systematically found by optimization approaches such as L-BFGS \cite{LBFGS}, FGSM \cite{FGSM}, DeepFool \cite{Deepfool}, C\&W \cite{CandW}. Readers are referred to \cite{AdvSurvey2,AdvSurvey} for a deeper insight in the digital adversarial attack approaches. 

Adversarial attacks and adversarial perturbations have intriguing properties which make them potentially effective in the physical domain:
\begin{itemize}
    \item \emph{Universality:} adversarial examples exist in all machine learning and deep learning models \cite{AdvExistence,AdvExistence2}. This means, all modern surveillance tasks based on deep learning in detection, identification, tracking, and action recognition are vulnerable \cite{AdvSurvey}.
    \item \emph{Transferability:} adversarial examples learned to attack one model can also be effective against a different, potentially unknown, model \cite{AdvTransferability,AdvTransferability2}. This makes blackbox attacks a real physical threat. 
    \item \emph{Shape and Size:} adversarial examples can be present in a wide range of shapes and sizes, from a single pixel \cite{OnePixelAttack} to a rectangular and circular patches \cite{AdvPatches,CircularAdvPatches}. The arbitrary shapes and sizes can allow the adversarial examples to be applied to different accessories. 
\end{itemize}

\begin{figure*}
    \centering
    \includegraphics[width=2\columnwidth]{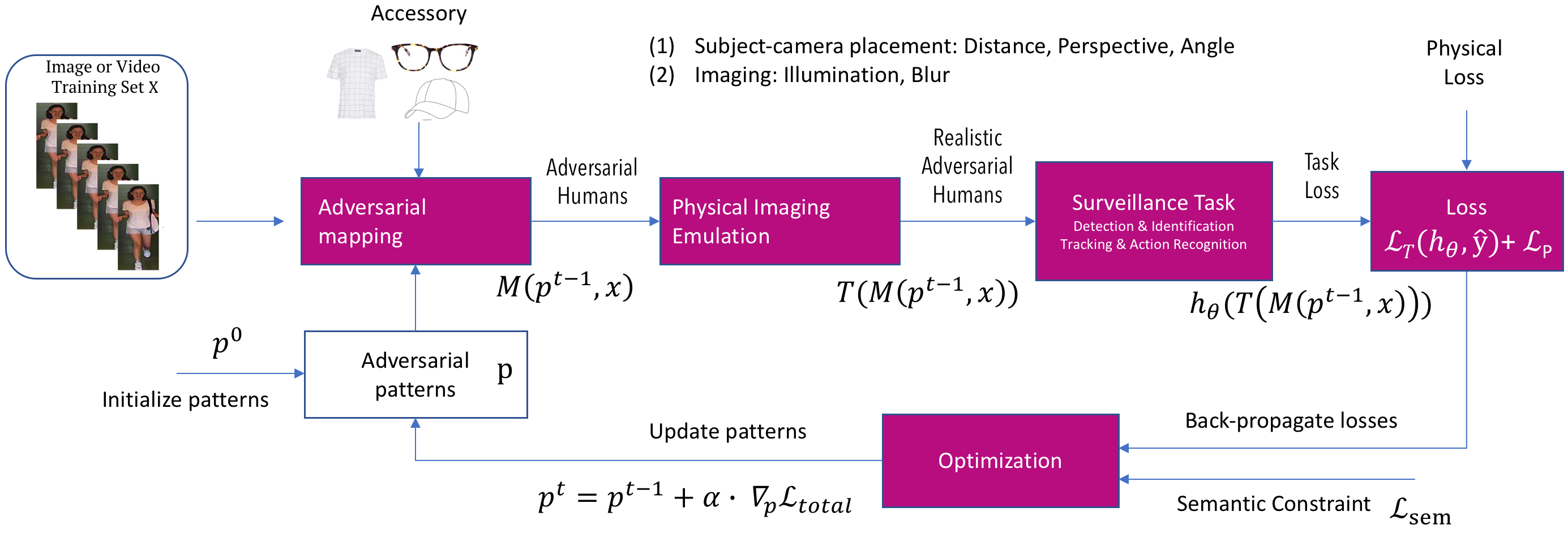}
    \caption{The proposed framework to investigate modern physical adversarial attacks, including both frame-based detection and identification and video-based tracking and action recognition tasks. While all approaches can be observed under the same framework, we can easily compare and contrast to see where the contributions are and what are missing to build effective adversarial physical attacks. }
    \label{fig:Framework}
\end{figure*}

\vspace{-9px}
\subsection{Physical adversarial attacks}
\label{sec:PhysicalFramework}
In 2016, Kurakin \emph{et al.} \cite{CellphoneAttacks} first demonstrated that threats of adversarial attacks are feasible in the physical world. They printed adversarial images and took snapshots from a cell-phone camera. These images were fed to TensorFlow Camera Demo app that uses Google’s Inception model for object classification. It was shown that a large fraction of images were misclassified even when perceived through the camera.

\vspace{3px}
\noindent\textbf{Physical vs. Digital}\\
Despite the potential of digital adversarial attacks to transfer to the physical domain, they usually fail due to the challenges in the imaging conditions, the dynamics of adversaries, and the multi-camera nature of surveillance systems. The impact of the imperceptible patterns learned in digital adversarial attacks approaches are easily diminished due to these challenges. Hence different from digital adversarial attacks which aim to learn a minimal and imperceptible perturbation, $\theta$, physical adversarial attacks seek to learn a realizable and perceptible pattern, $p$, which can be printed as a physical accessory for an adversary to carry or wear to change their appearance to mislead a detector, a tracker, an identifier or an action recognizer. The adversarial pattern $p$ is formulated in various forms depending on the physical items to be employed such as rectangular or circular forms in t-shirts, hats and stickers; or arbitrary forms in glasses and lighting, etc. In contrast to the constraint on minimality and imperceptibility of digital perturbations, physical adversarial patterns have to be \emph{printable} and \emph{survive physical real-world conditions} together with their challenges in the diversity of imaging conditions and the dynamics of adversaries.

\vspace{3px}
\noindent\textbf{What make physical adversarial attacks challenging?}\\
Compared with digital adversarial attacks, physical adversarial attacks in surveillance exhibit a new set of challenges due to the ``in the wild'' nature of the scene and subjects.
\begin{itemize}
    \item \emph{Fabrication of the adversarial accessories:} adversarial changes to digital images can be made at very fine pixel granularity, however the printing of real-world adversarial accessories can loose fine details. 
    \item \emph{Imaging conditions:} the unconstrained 
    physical environment introduces unpredictable changing and non-linear lighting/illumination. In addition, each imaging device has its own limitations in imaging capability (e.g. resolutions, color depth, lens quality, focal length, etc.)
    \item \emph{Non-rigid nature of adversaries:} compared with rigid and flat objects such as a traffic sign, a human adversary with different body components joined could cause severe distortion.
    \item \emph{Dynamics of adversaries:} an adversary is not stationary due to action dynamics, which changes his appearance in terms of angle, distance, occlusion, motion and activities to a surveillance camera.
    \item \emph{Multi-camera nature of surveillance systems:} an adversary may be imaged from a multitude of cameras, which may have different parameters (e.g. resolutions, fixed/zoom, visible/near-infrared, etc.) and configurations (e.g. subject-camera distance, angle, lighting).
    \item \emph{Multimodal recognition:} in reality, multiple modalities can be used for detection and recognition. For example, both face and gait can be simultaneously employed to recognize a human. 
\end{itemize}
Additional design criteria have to be considered in designing physical adversarial attacks to cater for this unpredictability.

\vspace{3px}
\noindent\textbf{Framework}\\
Physical adversarial attacks drawing heavily on the fact that one can learn an adversarial pattern through minimizing the physical surveillance detection or identification task loss when mapping the adversarial pattern on an accessory which an adversary can wear or carry with \cite{AdvTshirtFB,advGlasses,AdvHat}. We propose a framework to understand modern physical adversarial attacks in the literature as shown in Fig.~\ref{fig:Framework}. While all approaches can be observed under the same framework, we can easily compare and contrast to see where the contributions are and what are missing to build effective adversarial physical attacks.

The framework illustrates how to learn an adversarial pattern, $p$, to allow an adversary, while wearing and carrying it, to walk freely through the surveillance system without being detected or recognized by the surveillance task, $h_{\theta}$. Please note that for clarity of the illustration, we illustrate it only for object detection tasks but it is applicable to both frame-based detection and identification tasks and video-based tracking and action recognition tasks. Therefore, the model $h_{\theta}$ is task-specific, representing the object detection model in case of the object detection task, representing the object tracking model in case of surveillance object tracking task, and representing the action recognition model in case of surveillance action recognition task. Once learned, the adversarial pattern can be printed on an adversarial accessory, e.g. t-shirt, eyeglass, hat, and the adversary can wear or carry the adversarial accessory to mislead the surveillance task. 
\begin{itemize}
    \item In case of detection, the adversary aims to not being detected or detected as any other class except human.
    \begin{itemize}
        \item $h_{\theta}(\hat{x}) = c$ and $c \neq y$: if the adversarial accessory misleads the detection as a specific target class, $c$, which is different from the human class, $y$, it is called a target attack. 
        \item $h_{\theta}(\hat{x}) \neq y$: if the adversarial accessory misleads the detection as any target class as long as not the human class, it is called a non-target attack. 
    \end{itemize}
    
    \item In case of identification, the adversary aims to not being not recognized as his/her true identity.
    \begin{itemize}
        \item $h_{\theta}(\hat{x}) = c$ and $c \neq y$: if the adversarial accessory misleads the identification as a specific target identity, $c$, which is different from the true identity, $y$, it is called a target attack. 
        \item $h_{\theta}(\hat{x}) \neq y$: if the adversarial accessory misleads the identification as any target identity as long as not the true identity, it is called a non-target attack. 
    \end{itemize}
\end{itemize}

\begin{figure*}
     \centering
     \begin{subfigure}[b]{0.23\textwidth}
         \centering
         \includegraphics[width=\textwidth]{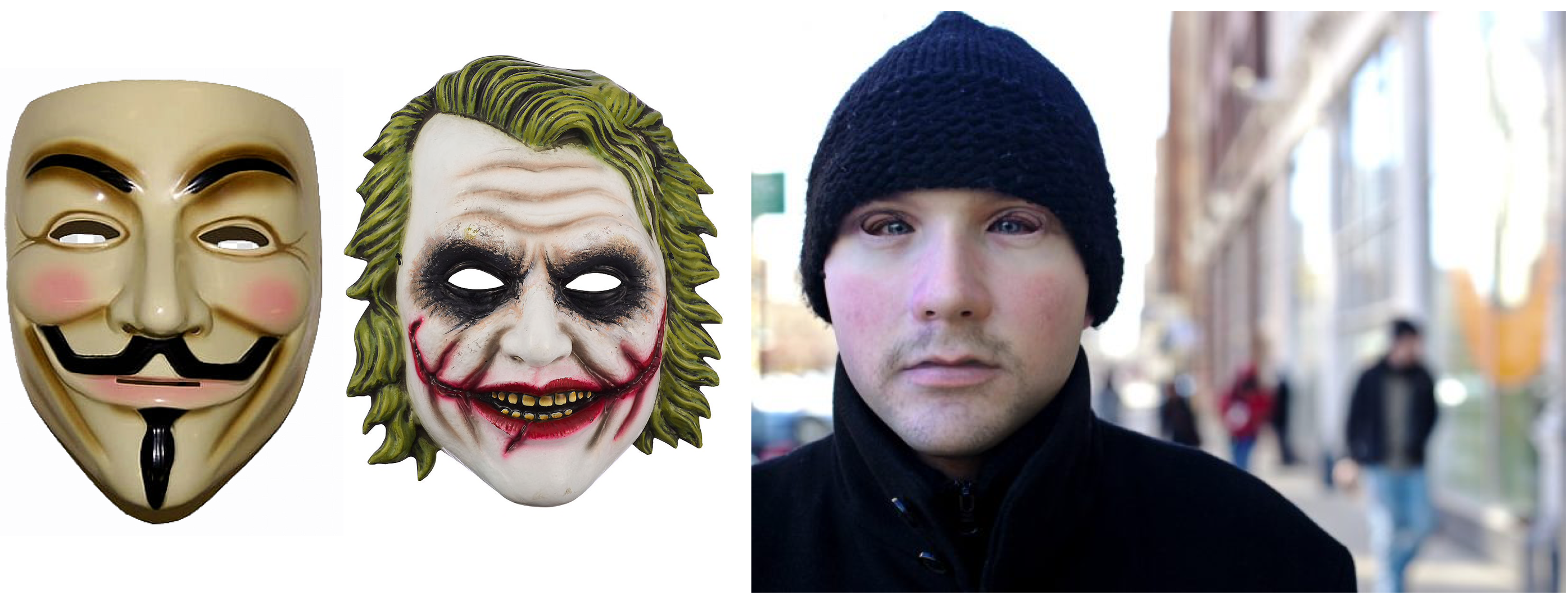}
         \caption{3D face masks \cite{3DFaceMask}.}
         \label{fig:3DFaceMask}
     \end{subfigure}
     \hfill
     \begin{subfigure}[b]{0.23\textwidth}
         \centering
         \includegraphics[width=\textwidth]{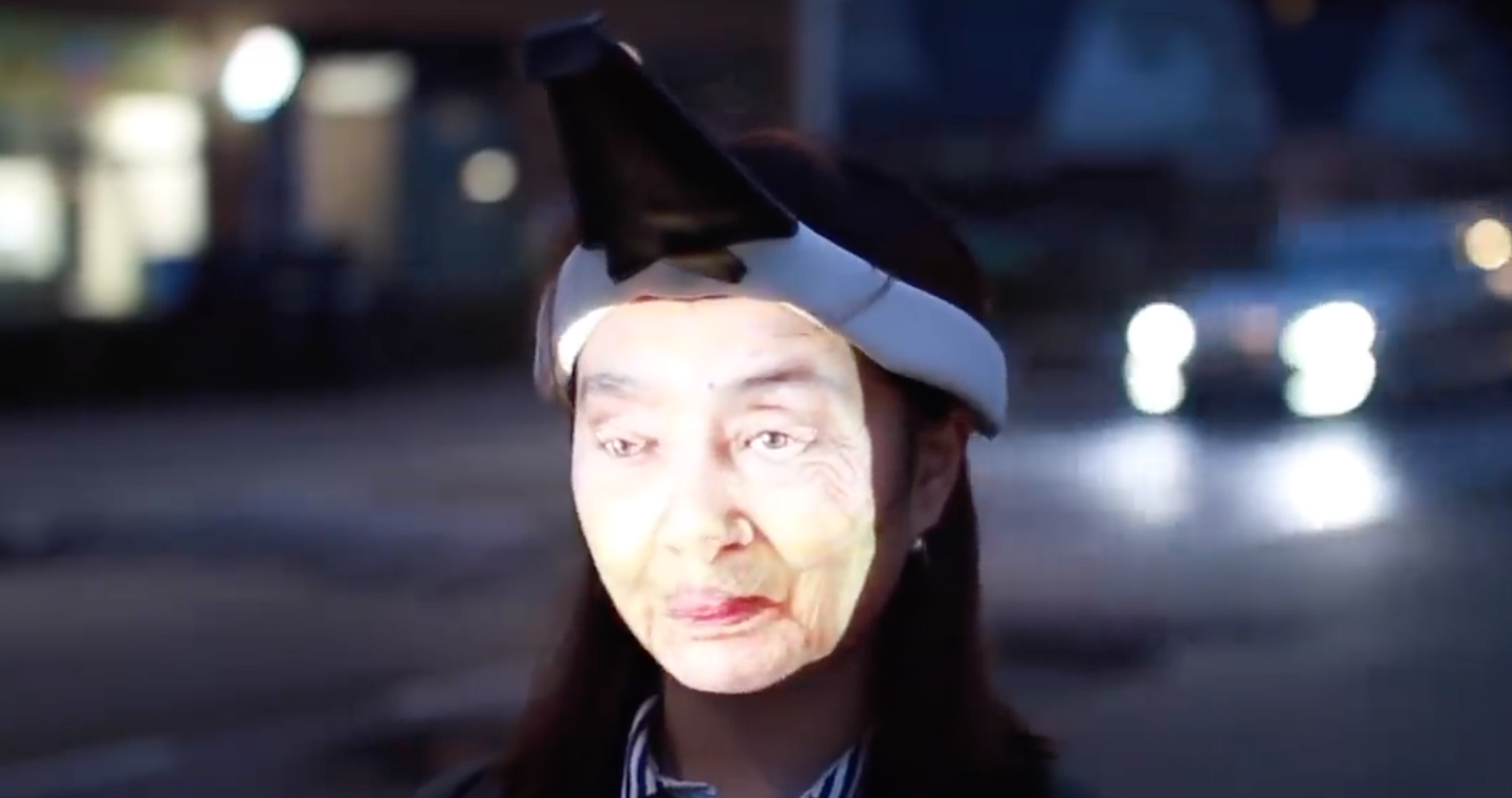}
         \caption{Face projector \cite{FaceProjector}.}
         \label{fig:FaceProjector}
     \end{subfigure}
     \hfill
          \begin{subfigure}[b]{0.23\textwidth}
         \centering
         \includegraphics[width=\textwidth]{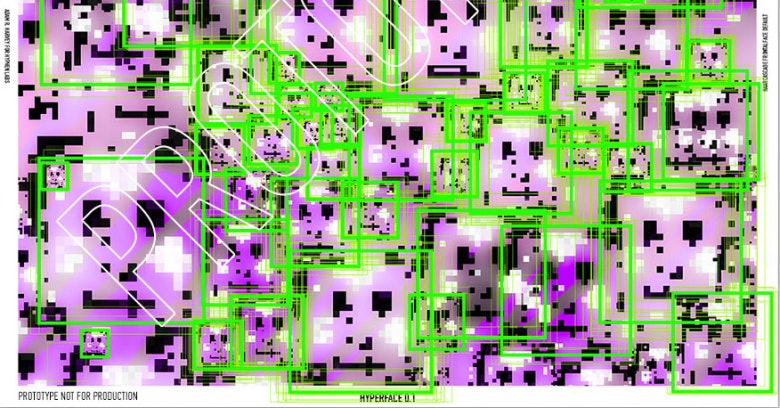}
         \caption{``Face'' patterns scarf \cite{Scarf}.}
         \label{fig:Scarf}
     \end{subfigure}
    \hfill
     \begin{subfigure}[b]{0.23\textwidth}
         \centering
         \includegraphics[width=\textwidth]{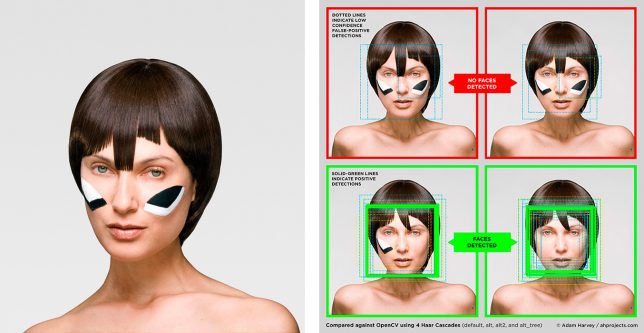}
         \caption{ Camouflage make-ups \cite{CVDazzle}.}
         \label{fig:CVDazzle}
     \end{subfigure}
     \\
    \begin{subfigure}[b]{0.23\textwidth}
         \centering
         \includegraphics[width=\textwidth]{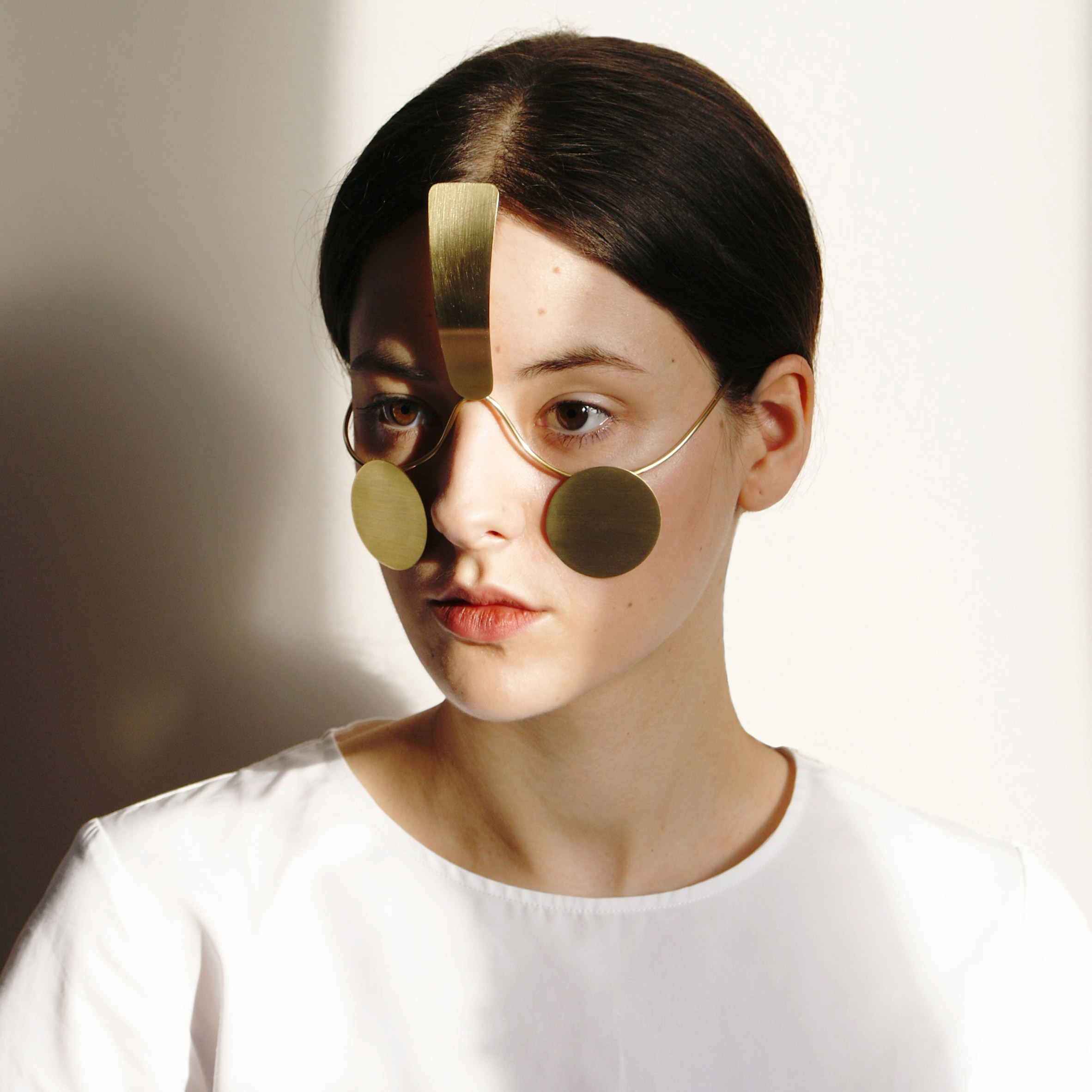}
         \caption{Incognito facial artifact \cite{FaceJewellery}.}
         \label{fig:FaceJewellery}
     \end{subfigure}
     \hfill
     \begin{subfigure}[b]{0.23\textwidth}
         \centering
         \includegraphics[width=\textwidth]{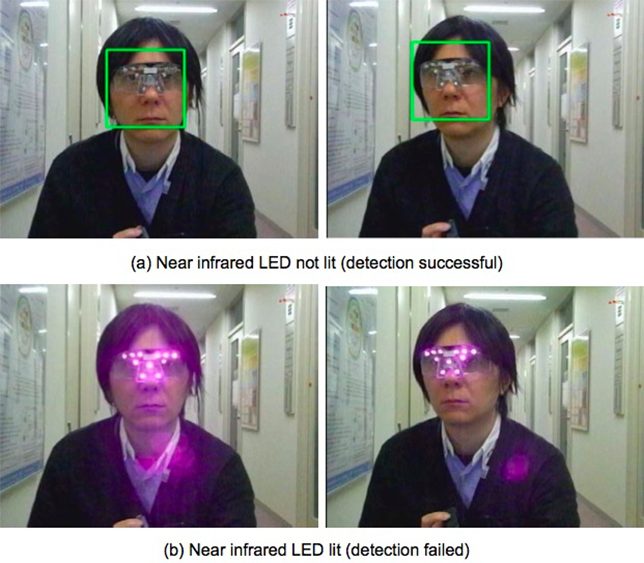}
         \caption{Adversarial LED glasses \cite{LEDglasses}.}
         \label{fig:LEDglasses}
     \end{subfigure}
     \hfill
     \begin{subfigure}[b]{0.23\textwidth}
         \centering
         \includegraphics[width=\textwidth]{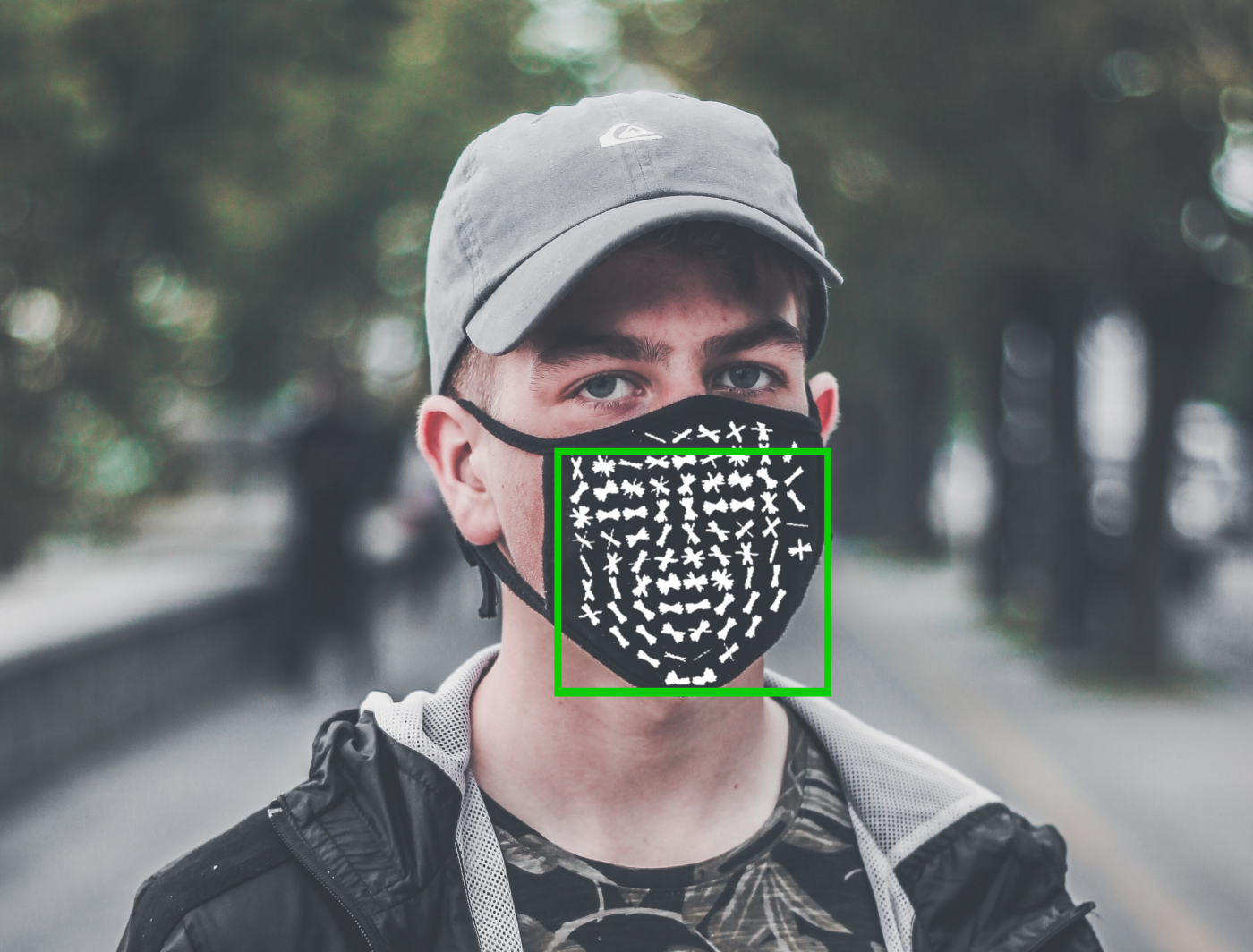}
         \caption{Face mask with  simulated annealing patterns \cite{AdvFaceMask}.}
         \label{fig:FaceMask}
     \end{subfigure}     
    \hfill
     \begin{subfigure}[b]{0.23\textwidth}
         \centering
         \includegraphics[width=\textwidth]{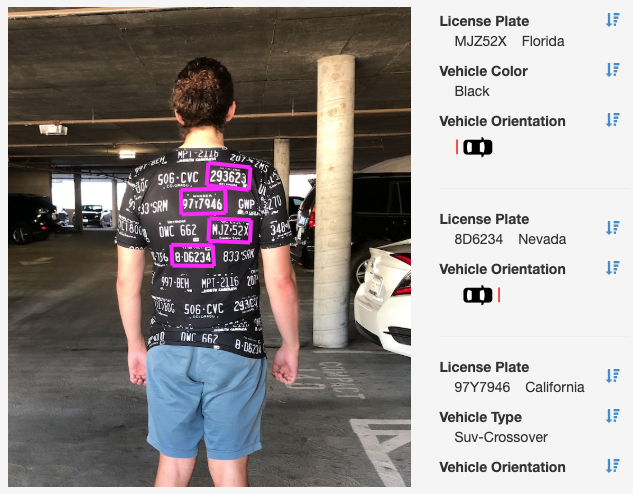}
         \caption{ Adversarial fashion \cite{AdversarialFashion}.}
         \label{fig:AdversarialFashion}
     \end{subfigure}

        \caption{Examples of human-designed physical adversarial attacks to thwart detection and identification.}
        \label{fig:humandesigned}
\end{figure*}
The learning of an adversarial pattern, $p$, is formulated as an optimization problem which minimize the expectation of the task loss and physical loss over the adversarial version of the training dataset $\hat{X}$,
\begin{equation}
\underset{p}{argmin} \underset{\hat{x} \in \hat{\chi}}{E}(\mathcal{L}_{T}) + \mathcal{L}_{P}, 
\end{equation}
where 
\begin{itemize}
    \item $\mathcal{L}_{T}$ denotes the task loss. This term penalizes the correct detection, identification, tracking or action recognition and encourages mis-detection, mis-identification, wrong tracking or mis-action-recognition.
    \item $\mathcal{L}_{P}$ denotes the physical loss. This term forces the adversarial pattern to be printable, to look smooth and natural.
\end{itemize}
These losses are calculated over the adversarial version of each image from the original training dataset $X$. The adversarial image, $\hat{x}$, is generated by
\begin{itemize}
    \item Adversarial Mapping $M(p^{t-1},x)$: map the current adversarial pattern, $p^{t-1}$, onto the accessory to make adversarial accessory. Then map the adversarial accessory onto the image, $x$.
    \item Physical Imaging Emulation $\hat{x}=T(M(p^{t-1},x))$: apply various transformation such as random change illumination/brightness or random blur to emulate imaging; or random resizing and affine to emulate subject-camera placement. 
\end{itemize}
This optimization can be solved by iteratively back-propagating the losses to update the adversarial pattern to make it converge to a solution where it can mislead the detection or identification task. 

Physical adversarial attack approaches discussed later will be analyzed under this framework, with variations in the form of framework components (where $\mathcal{L}_{T}$,  $\mathcal{L}_{P}$, $M(.)$ and $T(.)$ dominate). This framework allows us to compare existing approaches, see where their contributions are, and see what are missing to build an efficient learning system to physically attack surveillance systems.

\section{Human-designed adversarial attacks}
\label{sec:humandesigned}
For a long time, attackers have attempted to interfere the imaging process or obfuscate the subjects, e.g. body and face, with artifacts to hide their true identities and avoid being recognized. These interfere/obfuscation artifacts range from simple items to hide a part of the body part targets such as 3D face masks and facial jewels, and sophisticated patterns on scarfs or face masks or t-shirts to overwhelm surveillance software, to adjustable noise such as LED glasses and make-ups to target weaknesses in surveillance software. As heavily relying on human knowledge to design and optimize adversarial artifacts, these conventional approaches are usually limited to simple detection and identification algorithms or software, not the state of the art approaches. In addition, temporal surveillance tasks such as human tracking and action recognition are usually complicated and not easily intepretable for humans to investigate. Notice that all approaches implement non-target attacks.

\vspace{3px}
\noindent\textbf{3D face mask:}
Wearing a 3D printed mask of someone's face is the most obvious choice to avoid being identified. The most popular choices are the joker mask and the vendetta mask, which are widely used in armed robberies. These masks hide the identities of the attackers for dodging attacks. However, a 3D face mask of a specific person can also be used for impersonation attacks. For example, the 3D printed prosthetic image of artist Leo Selvaggio, the creator of the project URME Surveillance \cite{3DFaceMask}, is on sale for \$200. As illustrated in Fig.~\ref{fig:3DFaceMask}, even though the 3D face is creepy looking – especially the way the wearer’s eyes tend not to line up correctly with the eye holes, this mask will mostly allow an attacker to slip past biometric scanners without revealing your true identity.

\vspace{3px}
\noindent\textbf{Face projector:}
A wearable projector has been used to project faces and images onto the wearer’s own face \cite{FaceProjector}, which can confuse video surveillance cameras as illustrated in Fig.~\ref{fig:FaceProjector}.

\vspace{3px}
\noindent\textbf{Scarf:}
The team in Hyphen-Labs created a scarf that scrambles computer-vision algorithms. The purple material is packed with glitchy splodges—a camouflage for the 21st century that swaps out fake foliage for ghost faces, spamming the camera with potential matches \cite{Scarf}. The design scrambles those recording efforts by using ``maximally activated false faces based on ideal algorithmic representations of a human face''. That throws off the software’s ability to distinguish and recognize faces. There are 1,200 different “faces” on the scarf as depicted in Fig.~\ref{fig:Scarf}, so a lot of people wearing the same scarf would overload a surveillance system.

\vspace{3px}
\noindent\textbf{Make-ups:}
CV Dazzle explores how make-ups can be used as camouflage from face-detection technology \cite{CVDazzle}. For now, facial detection software still is not quite advanced enough to recognize human faces that significantly depart from the usual symmetric arrangement of features. That’s where the ‘anti-face’ comes in, a way of altering your appearance via hairstyles and cosmetics to fool computers into thinking they’re looking at something other than a face. The CVDazzle project explores this idea, as in Fig.~\ref{fig:CVDazzle}, with a series of six style tips for reclaiming privacy, explaining how to foil detection via makeup, obscuring the nose bridge and eyes as well as the elliptical shape of your face, modifying contrast and avoiding symmetry.

\vspace{3px}
\noindent\textbf{Artifacts:}
Artifacts can be stuck or painted to human faces to avoid detection software. For example, a Polish designer, Ewa Nowak, has developed a mask called Incognito that makes the wearer's face undetectable to facial recognition algorithms used in public surveillance cameras \cite{FaceJewellery}. Described by the designer as ``face jewellery'', the main structure of the mask-like accessory consists of a long piece of brass that has been shaped to fit the contours of the face, curving behind the ears like a pair of glasses. It features three prominent elements that work together to make the wearer's face ``unrecognisable'' to cameras. These are two brass circles that sit below the wearer's eyes, and a rectangular element positioned between the eyes as illustrated in Fig.~\ref{fig:FaceJewellery}.

\vspace{3px}
\noindent\textbf{Glasses:}
The LED glasses \cite{LEDglasses} designed by Japan’s National Institute for Informatics have a specific arrangement of LED lights around its eyes and nose thwarts face detection software at any distance. The light creates ‘noise’ that confuses the surveillance software to fool facial detection software. Chicago-made glasses, called Reflectacles, are designed to thwart facial recognition software that use infrared for illumination or 3D mapping/scanning by extreme reflecting or blocking the infrared light of security cameras \cite{Reflectacleglasses}. The glass frames are designed to be extremely reflective and the lens are designed to block infrared light.

\vspace{3px}
\noindent\textbf{Face Masks:}
Observing that a Histogram of Oriented Gradients (HOG) based face detector can be fooled by line patterns arranged as a face, Bruce \cite{AdvFaceMask} applied simulated annealing with random optimization to search for the pattern that can mislead most faces in the training data. The face mask designed, as illustrated in Fig.~\ref{fig:FaceMask}, succeed in manipulating a HOG-based face detector by successfully hiding the actual face and falsely detecting the face mask as a face. 

\vspace{3px}
\noindent\textbf{Adversarial fashion:}
Adversarial Fashion design specific patterns on the clothing to trigger Automated License Plate Readers, injecting junk data in to the systems used by the State and its contractors to monitor and track civilians and their locations \cite{AdversarialFashion}. The patterns, illustrated in Fig.~\ref{fig:AdversarialFashion}, were generated by testing a series of modified license plate images with commercial ALPR APIs, working to generate aesthetic fabric patterns that read in to devices and services as if they were real plates.

\section{Machine-learned adversarial attacks}
\label{sec:machinelearned}
Different to the human-designed nature of conventional approaches in Section~\ref{sec:humandesigned}, modern approaches seek to use learning algorithms to learn adversarial artifacts. This is to deal with the increasingly complicated machine learning algorithms that modern surveillance systems are employing. The learning capability allows machine-learned approaches to be more effective while multiple imaging conditions can be simulated in the training process. While human-designed adversarial attacks: (i) can only mislead simple detectors, e.g. Viola-Jones face detector, and (ii) can only perform non-target attacks; learning algorithms enables better learning which can learn over numerous conditions and can even emulate physical imaging conditions, hence: (i) can mislead state of the art detectors, identifier, trackers, action recognizers, and (ii) can perform both target and non-target attacks.

\subsection{Surveillance Human Detection Attacks}
Detecting humans accurately and reliably in a visual surveillance system is the first and crucial task before any further processing can be employed. Compared with object detection, human detection is more challenging due to the non-rigid nature of human body and the dynamics of humans in actions. 

\vspace{6px}
\noindent\textbf{State-Of-The-Art Detectors:} Modern detectors in the literature can be categorized into three groups: (i) one-stage detectors such as the YOLO series \cite{YOLOv2,YOLOv3,YOLOv4,ScaledYOLO}; (ii) two-stage detectors such as the R-CNN series \cite{FastRCNN,FasterRCNN,MaskRCNN}; and (iii) multiple-stage detectors such as Cascade R-CNN \cite{CascadeRCNN}. The detectors from all three groups have been considered in physical adversarial attack papers.

Researchers from KU Leuven have shown that a specifically-designed patch as small as $40cm\times40cm$ can successfully fool a state of the art human detector into thinking that a person is not a person -- or able to hide persons from AI-based security camera systems \cite{AdvPatches}. These adversarial patches have been printed on papers to be carried with or on outfit such as t-shirts to be worn to challenge detection in the physical world. Recent attempts in the literature have focused on different components of the framework proposed in Section~\ref{sec:PhysicalFramework} to make adversarial patches and adversarial t-shirts effective in the physical world.




\vspace{6px}
\noindent\textbf{Non-overlap Adversarial Patches for Object Detection:}
Liu \emph{et al.} \cite{DPatch} extended this idea to object detection by designing an adversarial patch which can be placed anywhere in an image to cause all existing objects in the image to be missed entirely by the detector. They simultaneously attack the bounding box regression and object classification. They proposed DPatch, which is an adversarial patch of a 40 by 40 pixel size that when added to any image could degrade the mAP of Faster R-CNN and YOLO from 75.10\% and 65.7\% down to below 1\%, respectively. DPATCH shows great transferability across detectors and datasets.
\begin{itemize}
    \item Detector Transferability: adversarial patches trained on YOLO manage to downgrade the mAP of Faster R-CNN from 75.10\% to 1.72\%. Adversarial patches trained on YOLO manage to downgrade the mAP of YOLO from 65.70\% to 0.02\%.
    \item Dataset Transferability: adversarial patches learned on the MS COCO dataset can degrade the mAP of Faster R-CNN trained on the VOC dataset from 75.10\% to 28.00\% and the mAP of YOLO trained on the VOC dataset from 65.70\% to 24.34\%.
\end{itemize}
Lee \emph{et al.} \cite{DPatch2} proposed to clip the patch to $[0,1]$ to correspond to actual perturbed images. They demonstrated effective physical attacks on YOLO by placing the learned adversarial patch printed on paper anywhere in the image, causing all existing objects in the image to be missed entirely by the detector, even those far away from the patch itself. 

\begin{figure}
    \centering
    \includegraphics[width=\columnwidth]{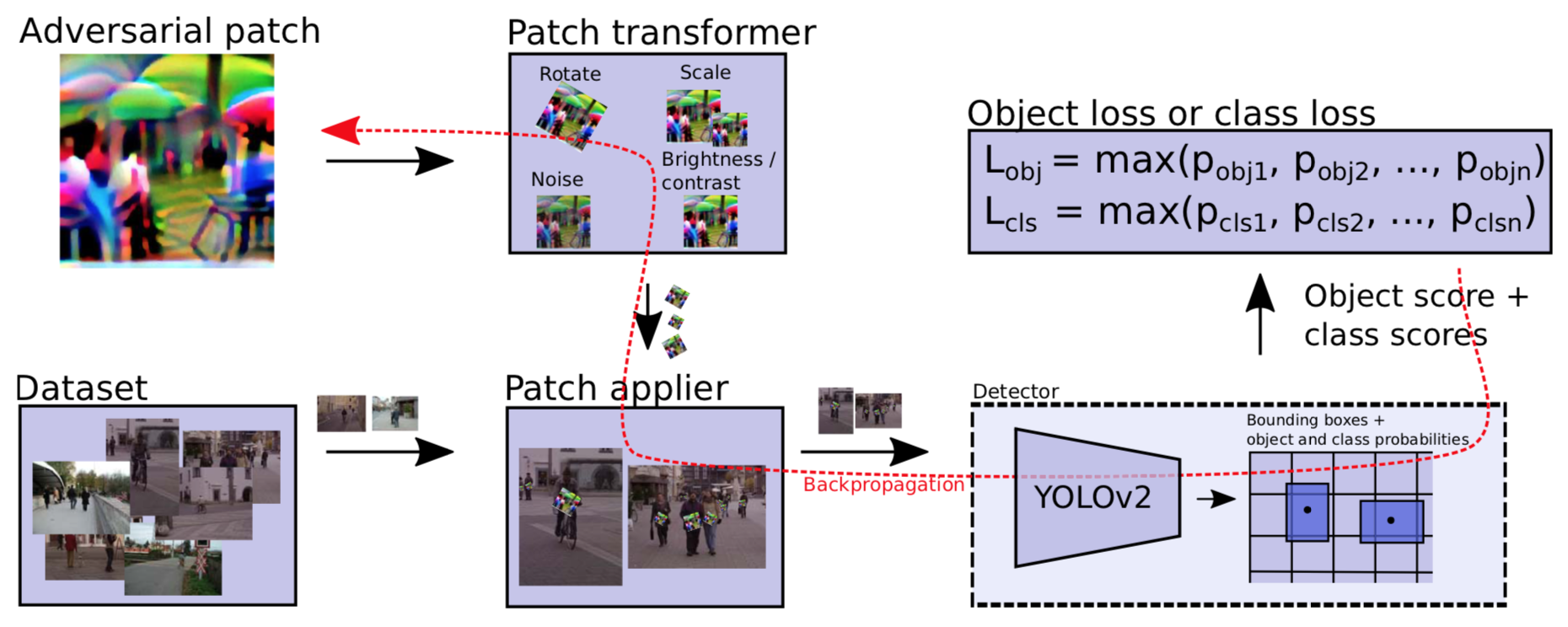}
    \caption{The pipeline in \cite{AdvPatches} to train an adversarial patch to attack the YOLO detectors.}
    \label{fig:Thys}
\end{figure}

\begin{figure*}
     \centering
     \begin{subfigure}[b]{0.3\textwidth}
         \centering
         \includegraphics[width=\textwidth]{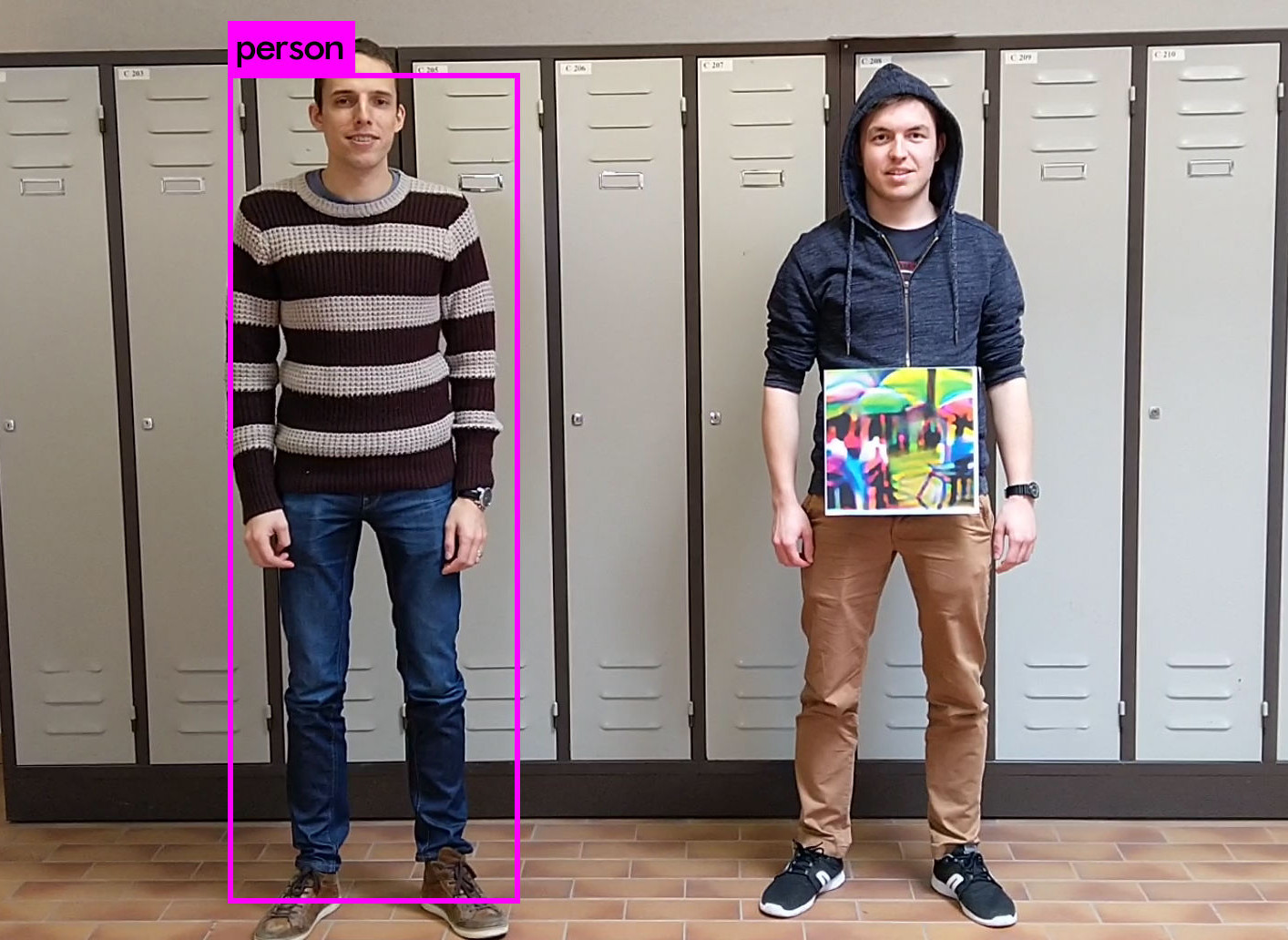}
         \caption{Adversarial Patch \cite{AdvPatches}.}
         \label{fig:ThysExamples}
     \end{subfigure}
     \hfill
     \begin{subfigure}[b]{0.3\textwidth}
         \centering
         \includegraphics[width=\textwidth]{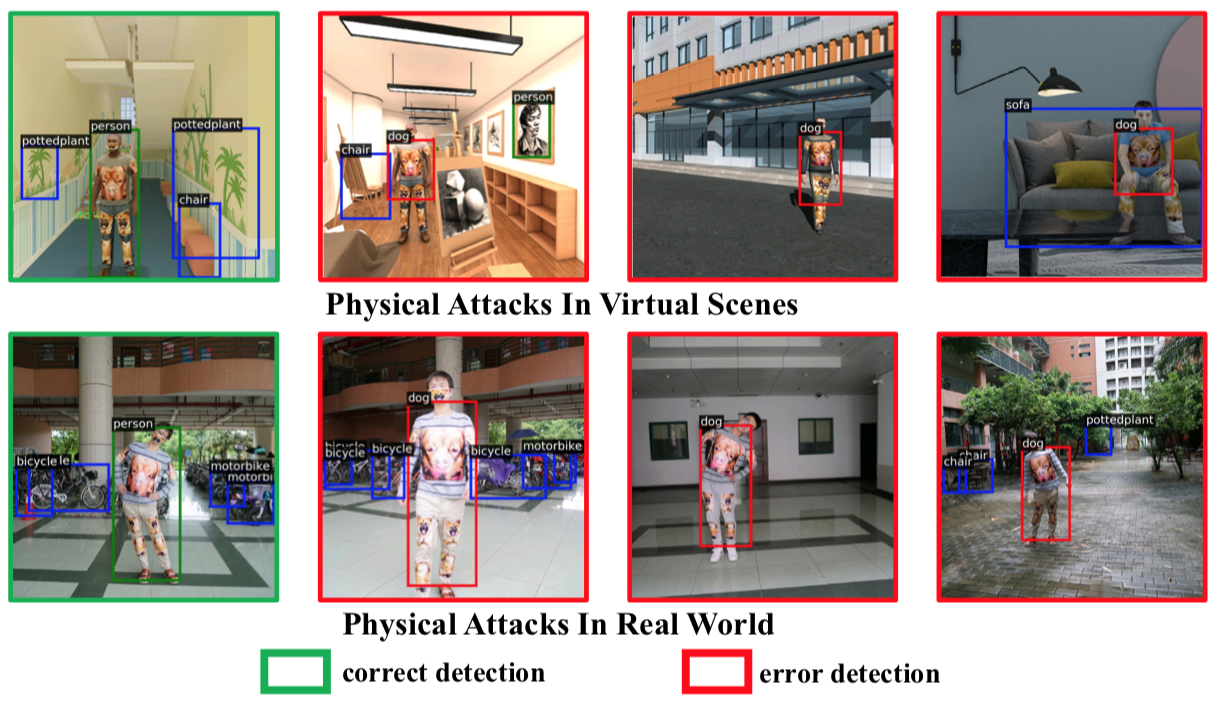}
         \caption{Adversarial T-shirt \cite{UPC}.}
         \label{fig:UPCexamples}
     \end{subfigure}
     \hfill
          \begin{subfigure}[b]{0.3\textwidth}
         \centering
         \includegraphics[width=\textwidth]{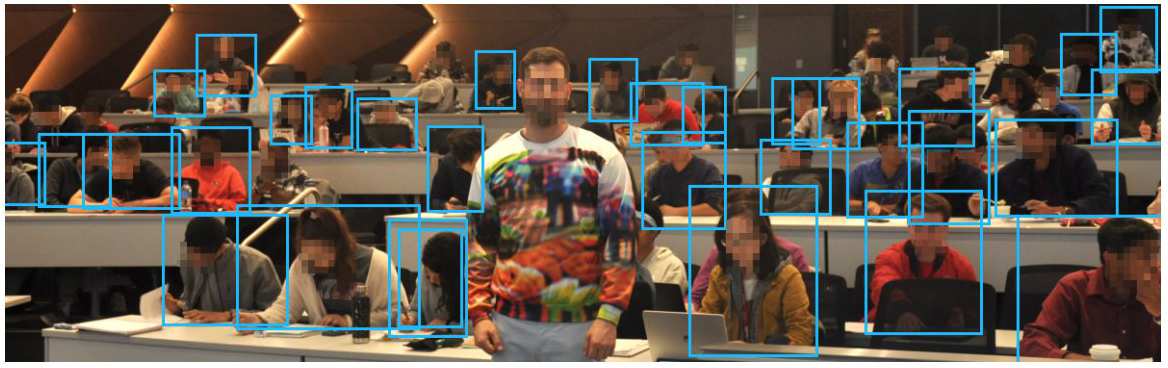}
         \caption{FB Invisibility Cloak \cite{AdvTshirtFB}.}
         \label{fig:AdvTshirt}
     \end{subfigure}
    \hfill
     \begin{subfigure}[b]{0.3\textwidth}
         \centering
         \includegraphics[width=\textwidth]{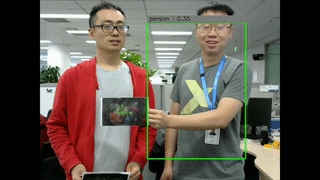}
         \caption{Baidu Invisible Cloak \cite{BaiduAdvTShirt}.}
         \label{fig:BaiduAdvPatches}
     \end{subfigure}
     \hfill
    \begin{subfigure}[b]{0.3\textwidth}
         \centering
         \includegraphics[width=\textwidth]{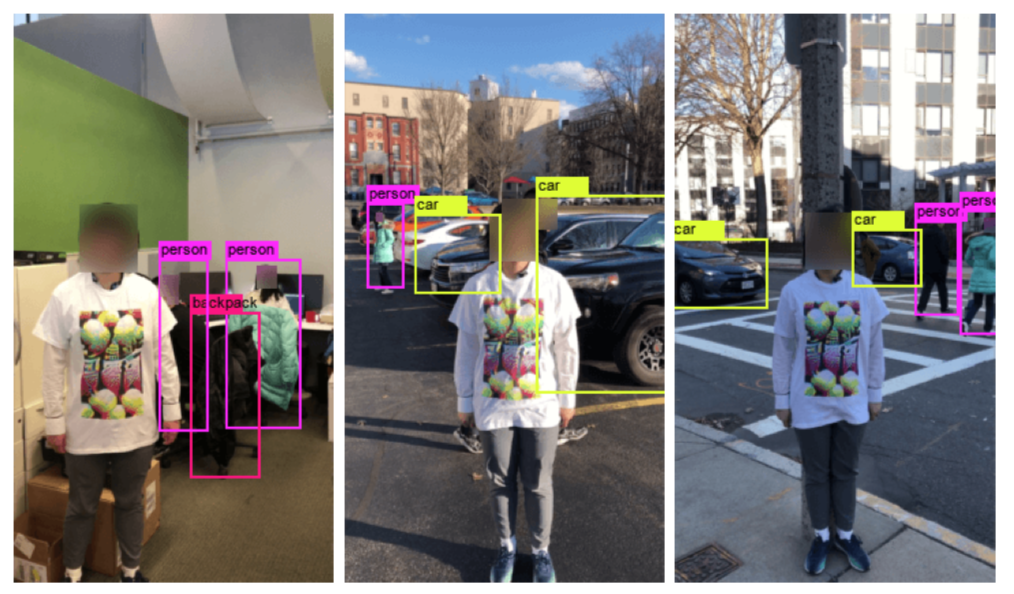}
         \caption{IBM Adversarial T-Shirt \cite{AdvTShirtIBM}.}
         \label{fig:IBMAdvTShirt}
     \end{subfigure}
     \hfill
     \begin{subfigure}[b]{0.3\textwidth}
         \centering
         \includegraphics[width=\textwidth]{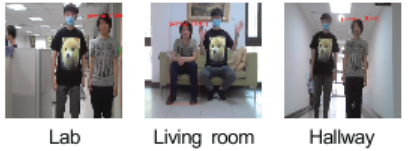}
         \caption{Naturalistic Adversarial Patch \cite{hu2021naturalistic}.}
         \label{fig:naturalistic_examples}
     \end{subfigure}
     \hfill

        \caption{A wide range of adversarial patches have specially designed to mislead the state of the art human detection algorithms such as the YOLO and R-CNN detectors to hide a person from being detected.}
        \label{fig:TShirtAdvPatches}
\end{figure*}

\vspace{6px}
\noindent\textbf{Thys:} Thys \emph{et al.} \cite{AdvPatches} proposed an approach to learn adversarial patches to attack person detection to fool automated surveillance cameras. The adversarial patches are learned by iteratively applying to images in the dataset and updated based on the objectness and class loss of the YOLO detector as shown in Fig.~\ref{fig:Thys}. The patch transformer contains random rotation, scaling, adding noise and brightness/contrast to make the adversarial patches better survive diverse imaging conditions in the physical world. The approach aims to minimize the total loss as follow.
\begin{equation}
\mathcal{L} = \mathcal{L}_{obj} + \mathcal{L}_{cls} + \alpha \mathcal{L}_{nps} + \beta \mathcal{L}_{TV} 
\end{equation}
where $\mathcal{L}_T = \mathcal{L}_{obj} + \mathcal{L}_{cls}$ is the detection loss of the YOLO detector and $\mathcal{L}_P = \alpha \mathcal{L}_{nps} + \beta \mathcal{L}_{TV}$ is the physical loss. $\mathcal{L}_{nps}$ is the non-printability score to make sure the patch is printable with common printers. The $\mathcal{L}_{TV}$ is the total variation to force smooth colour transitions and prevents noisy images. When printed on papers, the adversarial patches successfully hides persons from the YOLO detector as shown in Fig.~\ref{fig:ThysExamples}. 

\vspace{6px}
\noindent\textbf{UPC:} Huang \emph{et al.} \cite{UPC} proposed an approach to learn universal physical camouflage, which is an adversarial patch, designed to effectively attack all instances of humans. The setup targets two-stage detectors in the RCNN series by simultaneously attacking the region proposal network (RPN) and the classification (cls) network. The total loss is as follows,
\begin{equation}
\mathcal{L} = \mathcal{L}_{rpn} + \lambda_1 \mathcal{L}_{cls} + \lambda_2 \mathcal{L}_{reg} + \mathcal{L}_{TV}
\end{equation}
where $\mathcal{L}_T = \mathcal{L}_{rpn} + \lambda_1 \mathcal{L}_{cls} + \lambda_2 \mathcal{L}_{reg}$ is the detection loss of the Faster RCNN detector and $\mathcal{L}_P = \mathcal{L}_{TV}$ is the physical loss. Interestingly, they tested with multiple patches and patches for the face, arms, legs are also available.

The adversarial patches are tested in both virtual scenes and real world. In virtual scenes using 3DS Max, the adversarial patches are mapped on the body of 3D human models can reduce the detection accuracy from 100\% to 17\% using 8 patches. These patches also show great transferability. For example, they reduce the detection accuracy of the YOLO detector from 100\% to 69\% and of the SSD detector from 75\% to 13\%. In the real world evaluation, when printing these patches on t-shirts, the adversarial t-shirts successfully hides persons from the Faster RCNN detector as shown in Fig.~\ref{fig:UPCexamples}.

\begin{figure}
    \centering
    \includegraphics[width=\columnwidth]{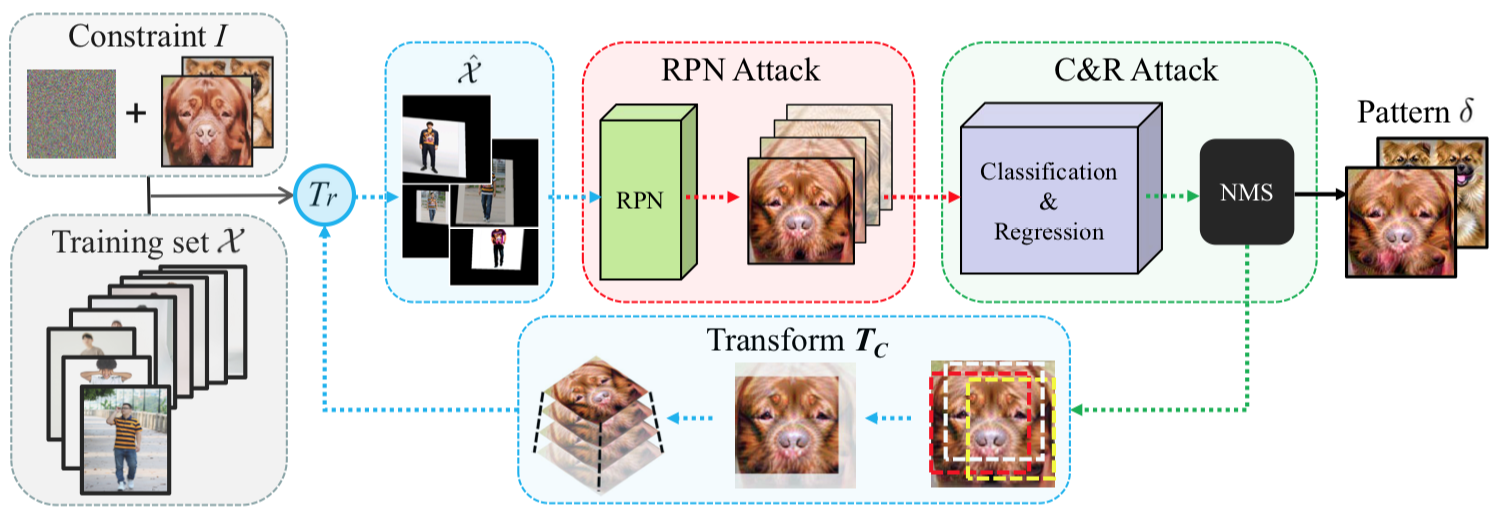}
    \caption{The pipeline in Universal Physical Camouflage \cite{UPC} to train multiple adversarial patches to attack Faster R-CNN.}
    \label{fig:UPC}
\end{figure}

\begin{figure}
    \centering
    \includegraphics[width=\columnwidth]{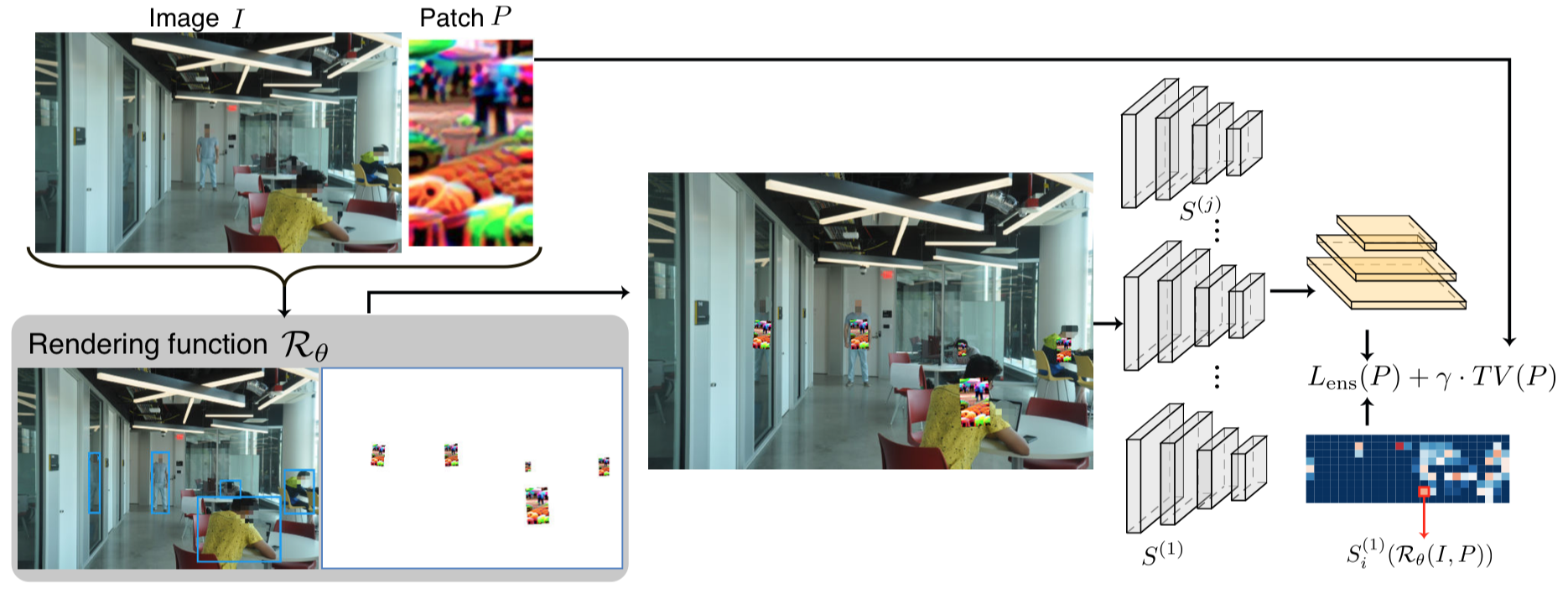}
    \caption{The pipeline of Facebook's work \cite{AdvTshirtFB} in training an adversarial patch to attack both YOLO and Faster R-CNN detectors.}
    \label{fig:FBAIAdv}
\end{figure}

\vspace{6px}
\noindent\textbf{Facebook Invisibility Cloak:}
Researchers from Facebook have learned and printed the patches into the t-shirt design to successfully fool many modern human detectors \cite{AdvTshirtFB}. Their approach minimizes the popular objectiveness loss and TV loss as follows,
\begin{equation}
L_\text {ens}(p) = \mathbb {E}_{\theta ,I} \sum _{i,j}\max \{\mathcal {S}_i^{(j)}(\mathcal R_\theta (I,p))+1, \, 0\}^2, 
\end{equation}
where $p$ is the adversarial patch, $I$ is the clean image, $R_\theta (I,p)$ is the transform function which applies a composition of brightness, contrast, rotation, translation, and sheering transforms that help make patches robust to variations caused by lighting and viewing angle that occur in the real world.
A detector network takes the adversarial patched image $R_\theta (I,p)$ as its input, and outputs a vector of objectness scores,  $S(R_\theta (I,p))$  one for each prior. These scores rank general objectness for a two-stage detector, and the strength of the “person” class for a one-stage detectors. The framework is illustrated in Fig.~\ref{fig:FBAIAdv}. As illustrated in Fig.~\ref{fig:AdvTshirt}, the learned t-shirt can render you invisible to human detection software, which is YOLOv2 in this case.

\vspace{6px}
\noindent\textbf{Baidu Invisible Cloak:}
Researchers from Baidu \cite{BaiduAdvTShirt} proposed an extended version of Expectation over Transformation (EoT) to design an adversarial sticker for physical attack as illustrated in Fig.~\ref{fig:BaiduAdvPatches}. Traditional EoT only transforms, rotates and scales digital images in 2D space. Researchers from Baidu considered the transformation of the stickers in real 3D world while generating these digital transformations. They built up an pin hole camera model to simulate the stickers’ transformation in the physical world. The original sticker is a US-Letter sized paper, they sampled the transformation by translating and rotating the photo with physical sticker each in three directions in the camera coordinate system. Then they got the projection of the photo with transformed stickers on the pin hole camera focal plane as an EOT digital transformed sample.

The loss for training is a typical combination of a detection loss, a regulization loss and a smoothness loss as,
\begin{equation}
\mathcal{L} = \mathcal{L}_{cls} + \lambda_1 \mathcal{L}_{reg} + \lambda_2 \mathcal{L}_{TV}
\end{equation}
where $\mathcal{L}_{reg} = ||tanh(\hat{x})-x||_2$
is the regulization loss as in the C\&W optimization method \cite{CandW}. In the physical attack experiment setting, they also experimented by displaying the adversarial sticker on the screen of a Macbook Retina 15-inch Laptop, and then let a person hold it and take photos in different scenarios. The experimental result shows that the precision of a Tiny YOLO model decreases from 1.0 to 0.28, achieving a success rate of 72\% of thwarting human detection.

\begin{figure}
    \centering
    \includegraphics[width=\columnwidth]{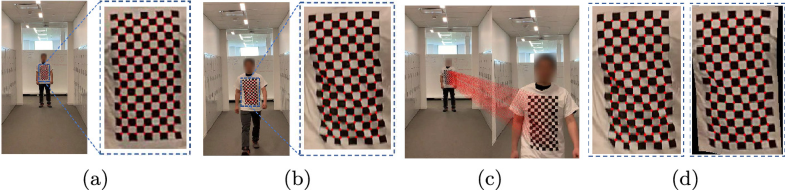}
    \caption{Researchers from IBM model 3D deformation of a t-shirt in realworld to learn a Thin Plate Spline deformation matrix via the checkerboard printed on a t-shirt to make the adversarial t-shirt effective when the subject is moving \cite{AdvTShirtIBM}. }
    \label{fig:TPS}
\end{figure}

\vspace{6px}
\noindent\textbf{IBM T-shirt:} Xu \emph{et al.} \cite{AdvTShirtIBM} showed that performance of existing adversarial t-shirts degraded drastically in actual physical world due to the movement of the wearers and the non-rigid nature of a human body. To deal with these issues, they explored Thin Place Spline (TPS) mapping to model the possible deformation encountered by a moving and non-rigid object. A checkerboard is printed on a t-shirt. A video of a subject walking while wearing the checkerboard t-shirt is recorded. The checkerboard enables learning the TPS deformation matrix from a 2D patch of the checkerboard to 3D physical appearance of the checkerboard in action as illustrated in Fig.~\ref{fig:TPS}. Once learned, the TPS deformation matrix allows adversarial mapping of the adversarial patch onto the body of the subject precisely with actual deformation such as wrinkles as shown in Fig.~\ref{fig:IBMAdvTShirt}.

Another notable contribution of this paper is that they showed physical ensemble attack of both YOLO and Faster RCNN can be designed from the perspective of min-max optimization, which yields much higher worst-case attack success rate than the averaging strategy over multiple detectors. Given $N$ object detectors, the physical ensemble attack is cast as,
\begin{equation}
\displaystyle \mathop {\mathrm{min}}\limits_{p}, \mathop {\mathrm{max}}\limits_{\mathbf{w}} \sum_{i=1}^N w_i \phi _i (p) - \frac{\gamma}{2} \Vert \mathbf{w} - \mathbf{1}/ N \Vert _2^2 + \lambda g(p),     
\end{equation}
where $p$ is the adversarial patch, $\mathbf{w}=\{w_i:i=1..N\}$ are known as domain weights that adjust the importance of each object detector during the attack generation, $\phi _i(p)$ is the EoT loss in each detector. The solution for this min-max optimization results in adversarial t-shirts that are robust to physical deformations, successfully attacked YOLOv2 74\% and 57\% in digital and physical worlds.

\begin{figure}
    \centering
    \includegraphics[width=\columnwidth]{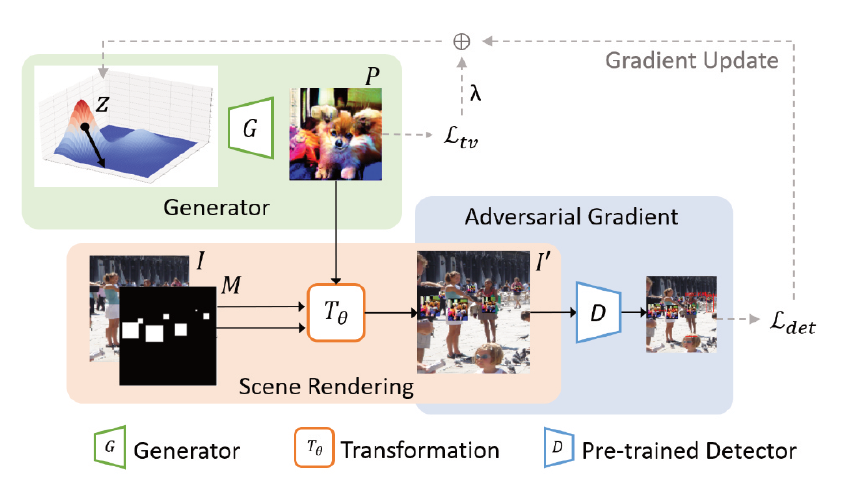}
    \caption{Overview of the naturalistic adversarial patch generation framework \cite{hu2021naturalistic}.}
    \label{fig:naturalistic}
\end{figure}

\begin{figure*}
    \centering
    \includegraphics[width=2\columnwidth]{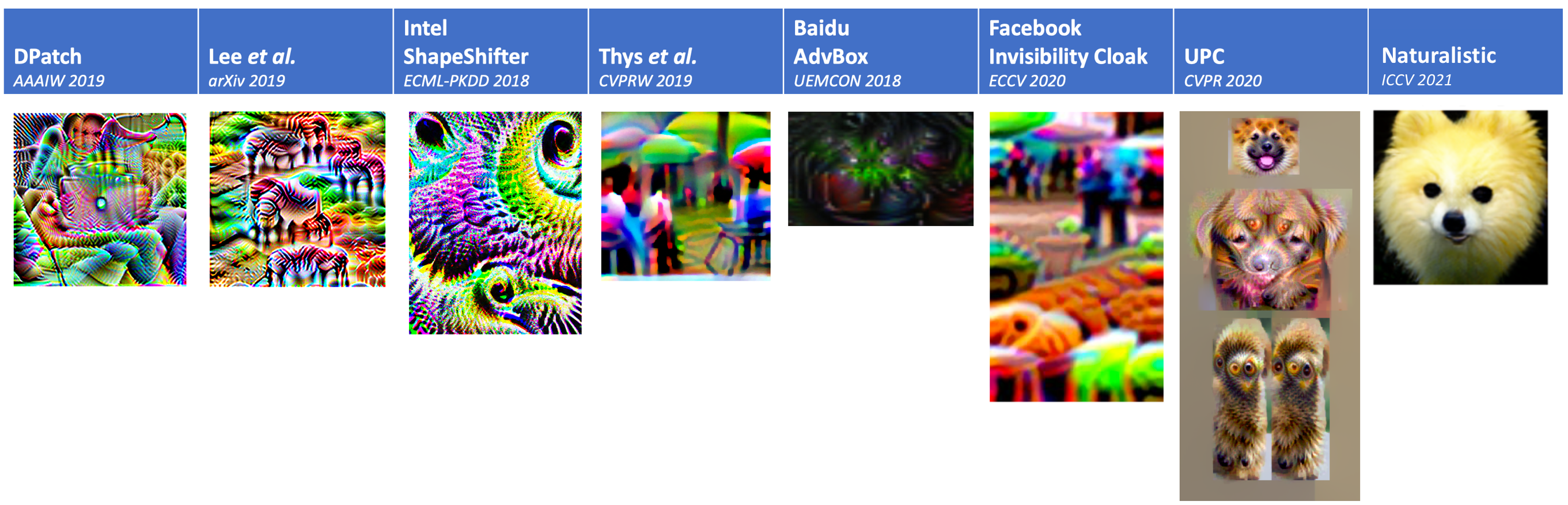}
    \caption{Summary of popular adversarial patches to thwart detection in the literature. While most adversarial patches have random patterns, recent approaches such as UPC \cite{UPC}, ShapeShifter \cite{ShapeShifter} and Naturalistic \cite{hu2021naturalistic} focus more on meaningful object-like patterns, making them to better survive the degradation of physical imaging conditions and avoid human attention. }
    \label{fig:AdvPatchesSummary}
\end{figure*}

\vspace{6px}
\noindent\textbf{Naturalistic Physical Adversarial Patch \cite{hu2021naturalistic}:} aims to generate effective physical adversarial patches while maintaining naturalistic characteristics. The authors show that the arbitrary way that the adversarial perturbations are generated in existing algorithms could lead to the generation of conspicuous and attention-grabbing patterns. As such, 
instead of optimizing for an adversarial patch in the pixel space, the authors propose to optimize it in the learned image manifold of a pre-trained Generative Adversarial Network (GAN). This GAN has been pre-trained using natural real-world images, as such, the adversarial perturbation generated from the generator, $G$, lies within the learned latent space. Specifically, let the learned latent space be denoted by $z \in \mathbb{R^d}$ and the initial adversarial patch, $P \in \mathbb{R^{H \times W \times 3}}$, can be estimated using $P = G(z)$, then the objective to update the latent vector $z$ can be written as,
\begin{equation}
\mathcal{L}_{total} = \mathcal{L}_{det} + \lambda_{TV}\mathcal{L}_{TV},
\end{equation}
where $\mathcal{L}_T = \mathcal{L}_{det}$ is the loss of the detector and $\mathcal{L}_P = \mathcal{L}_{TV}$ is the total variation loss introduced to encourage the smoothness. Hyper-parameter $\lambda_{TV}$ is used to balance the contribution from both loss terms. In \cite{hu2021naturalistic} the authors have incorporated the objectness and class, probabilities of the YOLO object detector as the $\mathcal{L}_{det}$ where,

\begin{equation}
 \mathcal{L}_{det} = \frac{1}{N}\sum_{i=1}^{N}\max_{j}[D^j_{obj}(I'_i)D^j_{class}(I'_i)].
\end{equation}
here $I'_i$ is the $i^{th}$ image in a mini-batch of size $N$ and $D^j_{obj}$ is the objectness probability of $j^{th}$ object while $D^j_{class}$ is the class probability of $j^{th}$ object. To balance the realism and attack performance, the authors used a threshold $\tau$ and ensured that the latent vector $z$ will not have a norm greater than $\tau$. Via adjusting the value of this threshold $\tau$, the authors show that the realism and attack performance objectives can be controlled. 
An overview of this naturalistic adversarial patch generation framework is given in Fig.~\ref{fig:naturalistic} and Fig.~\ref{fig:naturalistic_examples} show some physical adversarial examples generated using this framework.
This concept is extended in \cite{wang2021towards} where the authors have included an additional objective to ensure that the generated patterns are printable in the physical world.

\textbf{ \textit{Summary of patch-based adversarial designs:}} A figure comparing adversarial patches in the literature is illustrated in Fig.~\ref{fig:AdvPatchesSummary}. One the main drawback of the patch-based adversarial designs is that they are effective only when the adversarial patches face the camera frontally. They will easily fail at non-frontal viewing angles, as the camera may only capture a segment of the heavily deformed patch.




\vspace{6px}
\noindent\textbf{Adversarial Texture: \cite{AdvTexture}} offers a solution to this issue by proposing adversarial textures that can cover clothes with arbitrary shapes. The authors illustrate that textures can be generated in arbitrary sizes and can cover any cloth in any size. In addition, they can be placed in any local area of the clothing such that numerous local areas caught by the camera can attack the person detectors. Specifically, a novel two-stage generative attack method named Toroidal-Cropping-based Expandable Generative Attack (TC-EGA) is proposed. At the first stage a fully convolutional network (FCN) generates textures via sampling from random latent variables. The FCN allows the generated textures to be in multiple sizes. In the next stage, the best local pattern of the latent variable is searched using the Toroidal Cropping technique. These local patterns are tiled together and fed to the FCN in order to get the finally get adversarial texture. The authors have optimized their model using the following loss,
\begin{equation}
    \frac{1}{N}\sum_{i=1}^N[U(G_{\varphi}(z_i))],
\end{equation}
where $G_{\varphi}$ is the generator and $U$ is a combination of the task function (in an object detection task it denotes the confidence scores of the boxes predicted by the attacked object detector) and the physical loss. The authors have utilised total variation (TV) as the physical loss. $z_i$ denotes the sampled latent variables and $N$ denotes the total number of the samples. Furthermore, to maximizing the mutual information between $z$ and the adversarial path, the authors have used an auxiliary objective function which they name as the information objective function.

This approach is extended in \cite{hu2023physically} where  the authors propose to craft adversarial texture for clothes using 3D modeling of humans. Specifically, the authors illustrate that the simple adversarial textures are less effective in evading human detectors at multiple viewing angles and propose to perform the adversarial parameter optimization via 3D modeling of the humans. To generate realistic adversarial textures the authors get inspiration from camouflage patterns and adapt a soft version of the Voronoi diagram to generate the cluster regions of the camouflage pixels. Therefore, each polygon within the Voronoi diagram is assigned a color, location, and shape in the proposed differentiable soft version of the Voronoi diagram they have assigned a sampling probability of color which describes the probability of coloring a particular polygon using a particular color selected from a set of discrete colors. To improve the robustness of the adversarial texture to physical transformations the authors have modeled the physical warps and movements of the clothes in the 3D mesh space and constrain the adversarial augmentations to topologically plausible projections, geometrically plausible projection and 3D Thin Plate Spline (TPS) \cite{tang2019augmentation}. Additional argumentation was also proposed to calibrate the digital color to the physical color.  In the final objective function, the confidence scores of the object detector are used as the task loss and a concentration loss which penalizes too small polygons is used as the physical loss. 

\begin{figure*}
     \centering
     \begin{subfigure}[b]{0.3\textwidth}
         \centering
         \includegraphics[width=\textwidth]{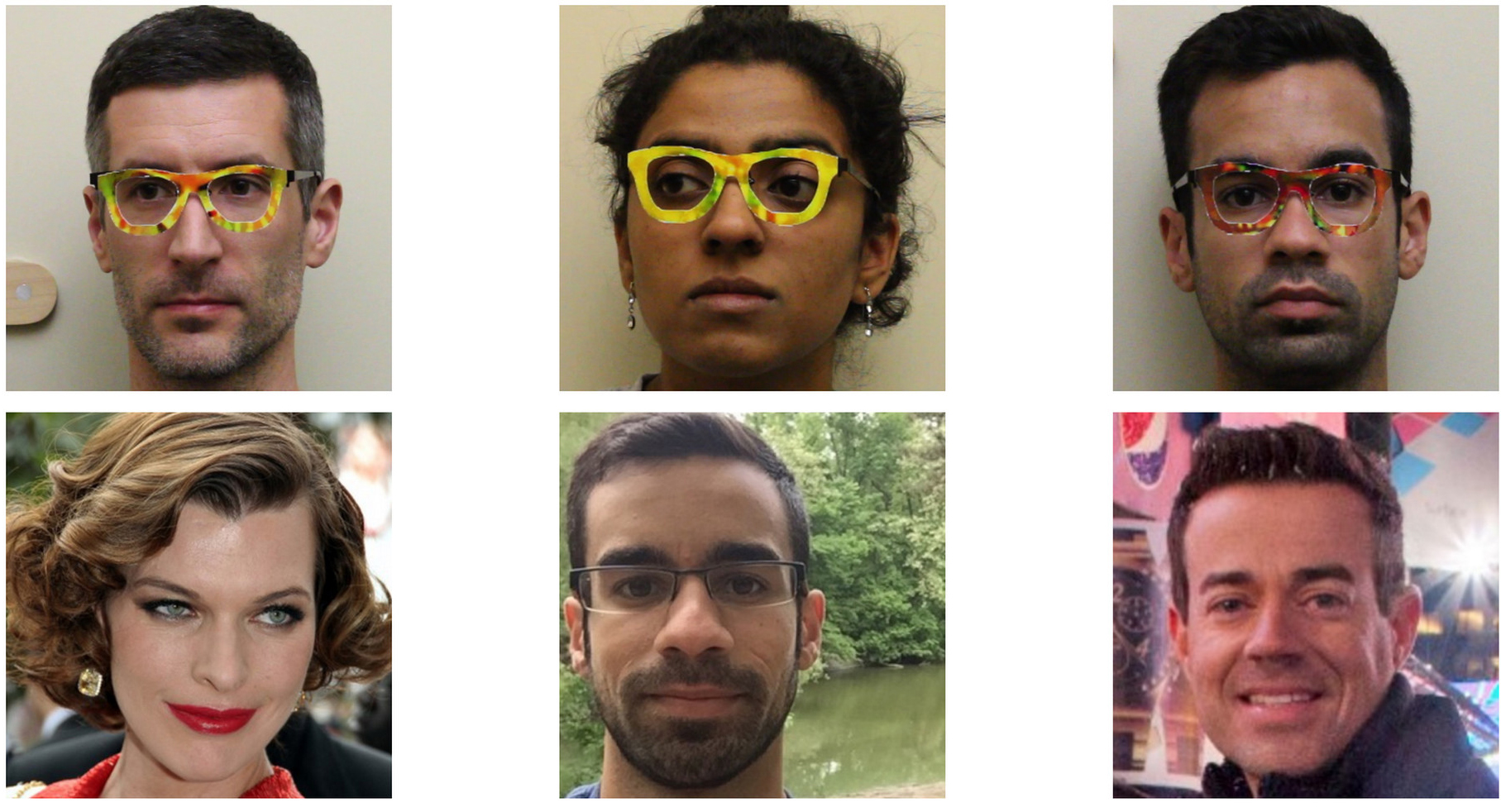}
         \caption{Adversarial Glasses \cite{advGlasses}.}
         \label{fig:AdvGlasses}
     \end{subfigure}
     \hfill
     \begin{subfigure}[b]{0.3\textwidth}
         \centering
         \includegraphics[width=\textwidth]{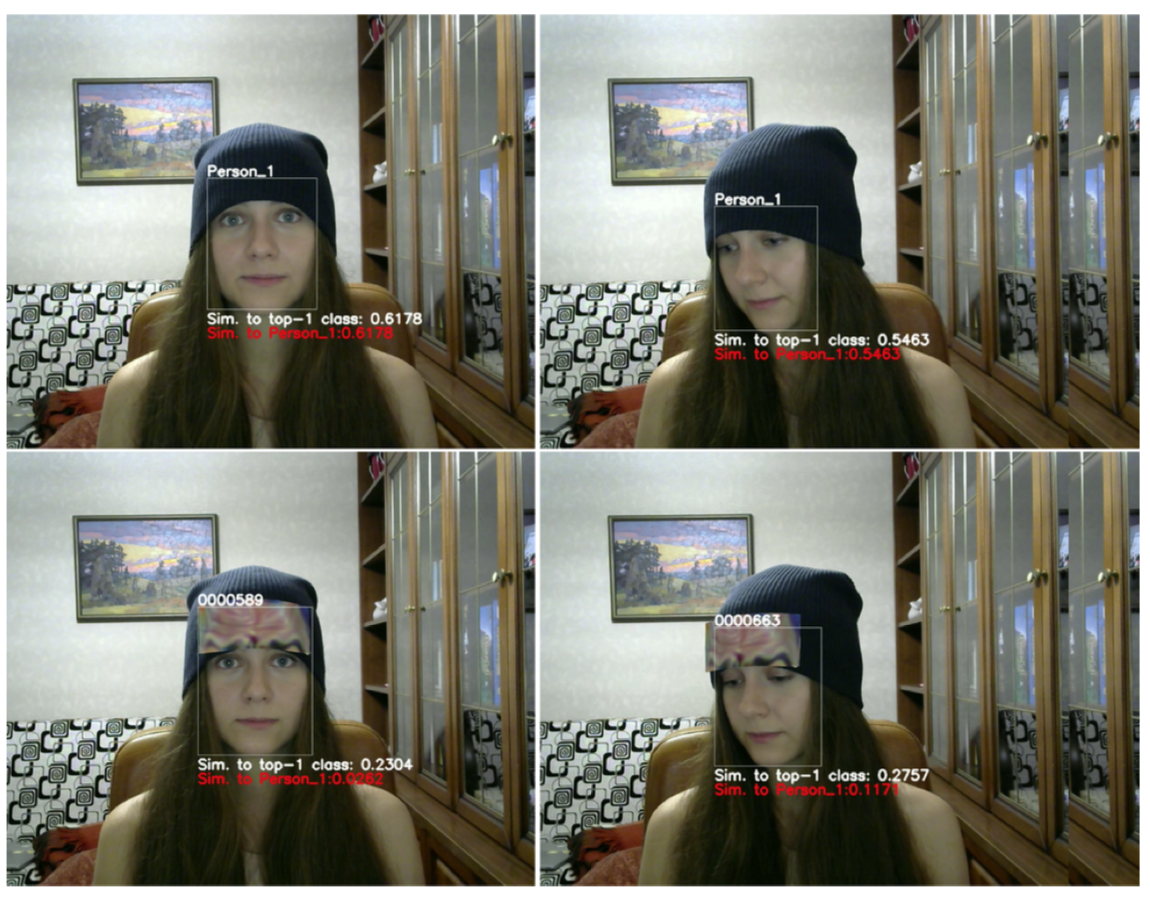}
         \caption{Adversarial Hat \cite{AdvHat}.}
         \label{fig:AdvHat}
     \end{subfigure}
     \hfill
          \begin{subfigure}[b]{0.3\textwidth}
         \centering
         \includegraphics[width=\textwidth]{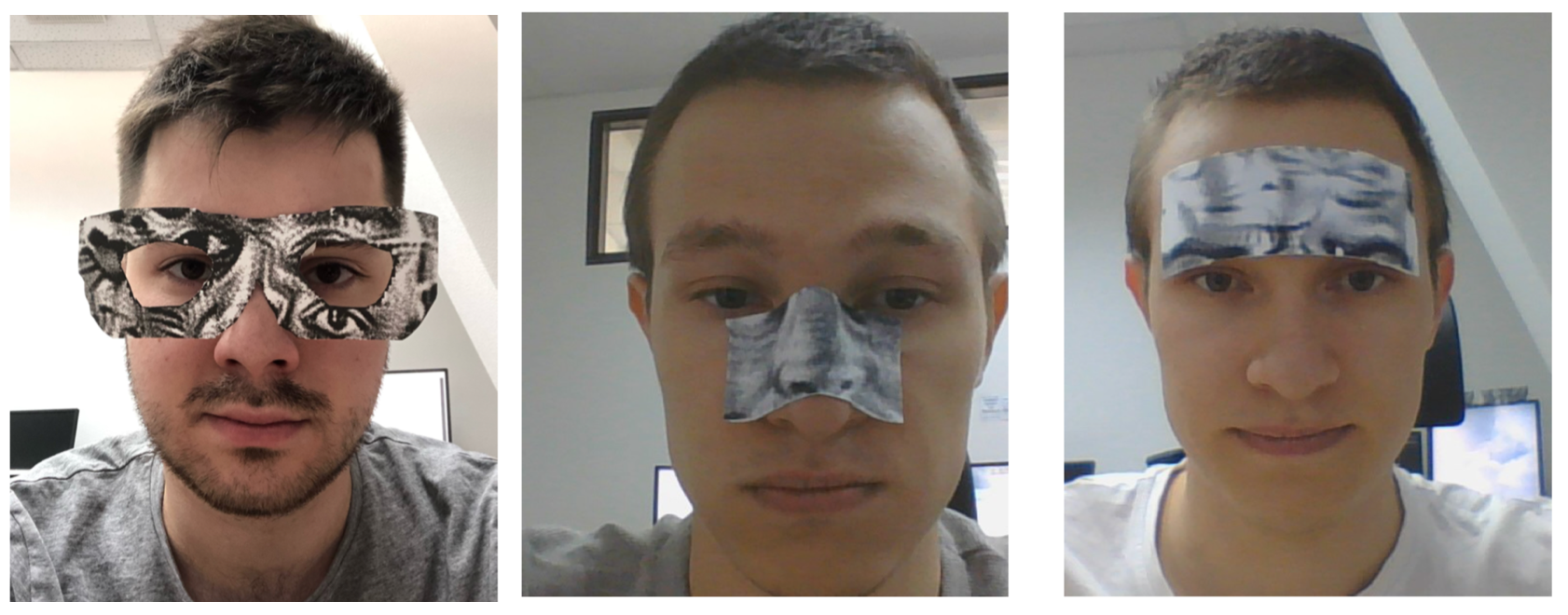}
         \caption{Adversarial Patch \cite{Scarf}.}
         \label{fig:Scarf}
     \end{subfigure}
    \hfill
     \begin{subfigure}[b]{0.3\textwidth}
         \centering
         \includegraphics[width=\textwidth]{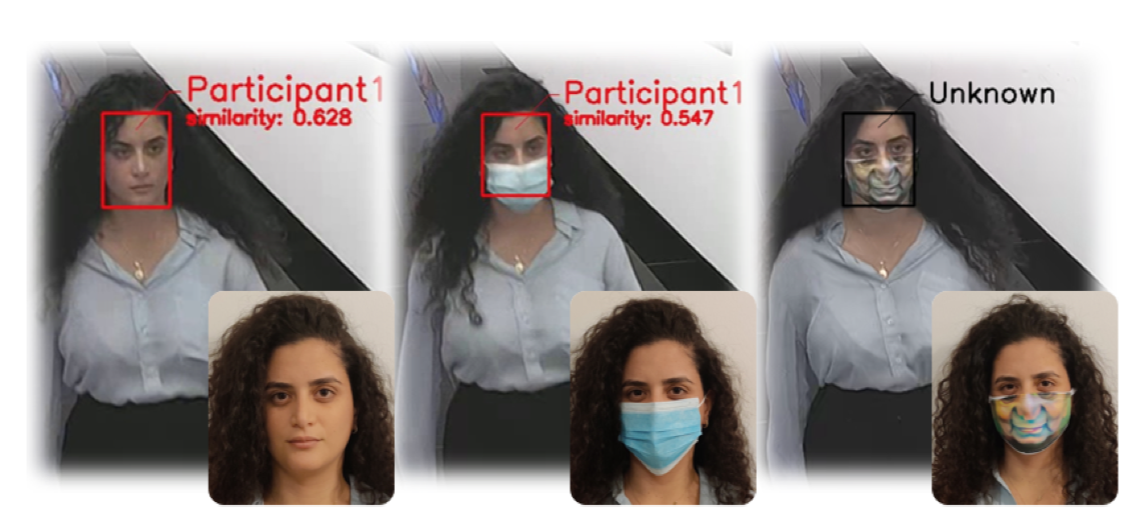}
         \caption{Adversarial Mask \cite{AdvMask}.}
         \label{fig:AdvMask}
     \end{subfigure}
     \hfill
    \begin{subfigure}[b]{0.3\textwidth}
         \centering
         \includegraphics[width=\textwidth]{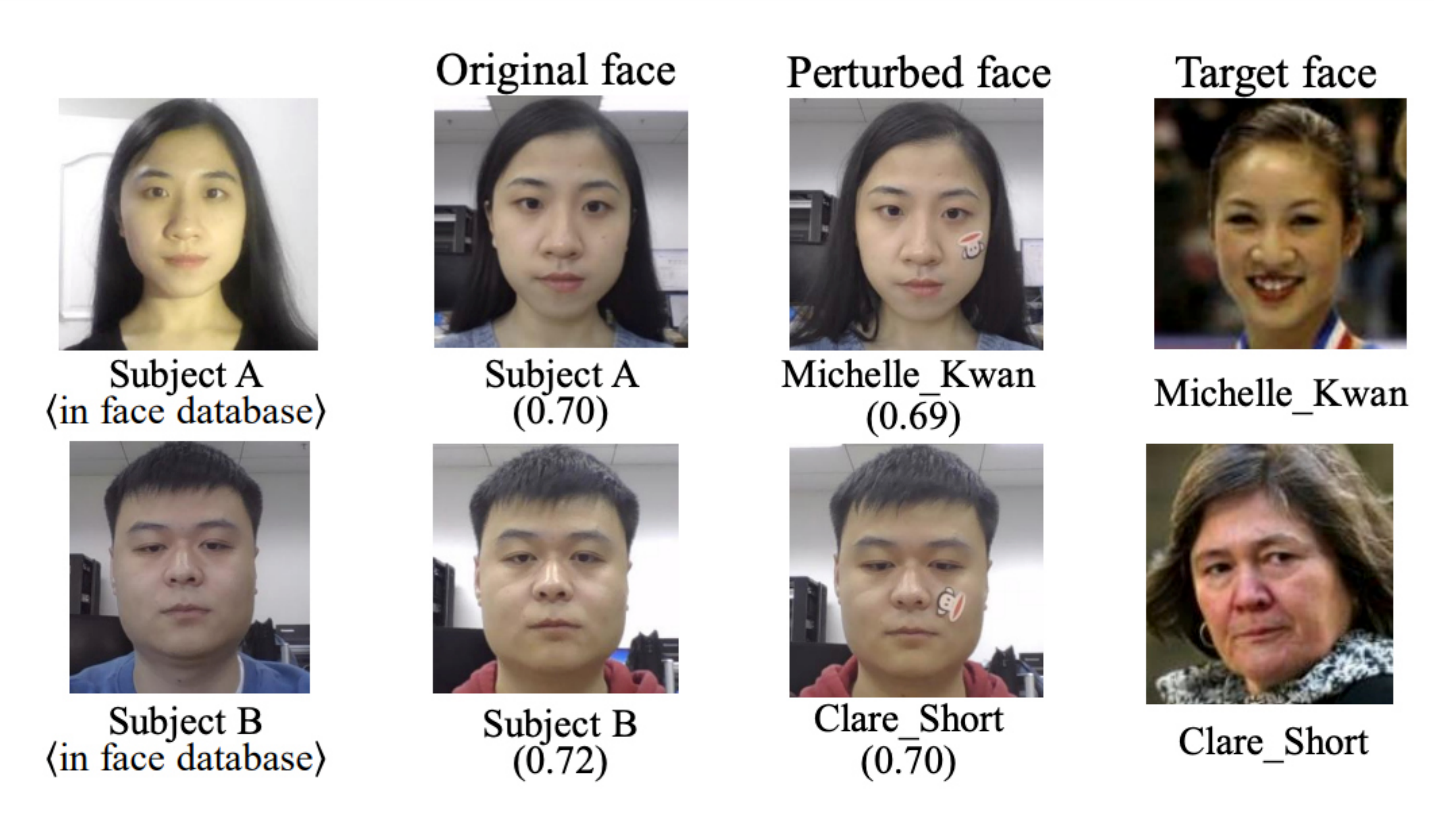}
         \caption{Adversarial Sticker \cite{guo2021meaningful}.}
         \label{fig:AdvSticker}
     \end{subfigure}
     \hfill
     \begin{subfigure}[b]{0.3\textwidth}
         \centering
         \includegraphics[width=\textwidth]{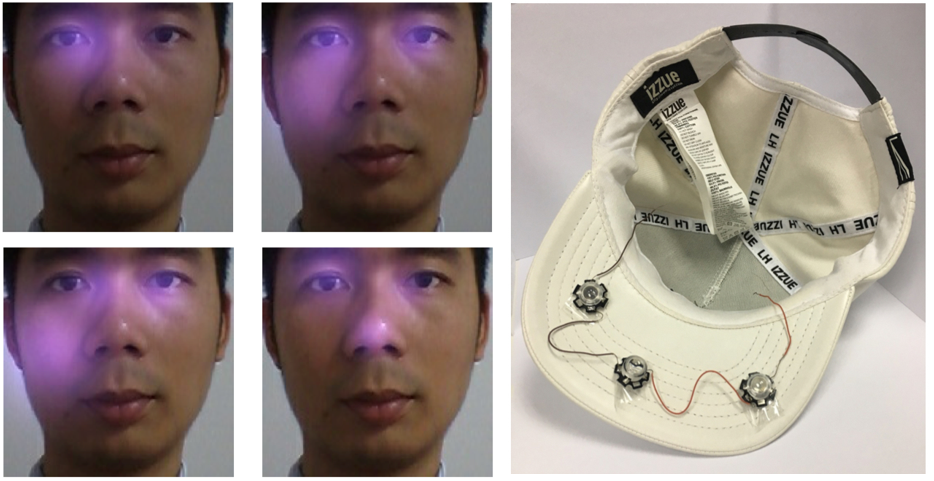}
         \caption{Adversarial Cap \cite{AdvCap,QUTAdvLED}.}
         \label{fig:AdvCap}
     \end{subfigure}
     \hfill

        \caption{Adversarial items such as glasses, hat, patch, mask, sticker, cap which are specially designed, have been shown to manage to mislead the state of the art facial recognition ArcFace to wrongly recognize a person with another identity.}
        \label{fig:AdvItemsFace}
\end{figure*}

\subsection{Surveillance Human Identification Attacks}
Three identification modalities have been investigated for physical adversarial attacks in the literature: face and person Re-ID. This section will delve deep into each modality.

\vspace{6px}
\subsubsection{Face Recognition} \hfill

\noindent{Face recognition aims to identify or verify the identity of an individual using their face. Face recognition is one of the most popular forms of human identification in surveillance due to the visibility and availability of human faces \cite{DFR_survey}.}

\vspace{6px}
\noindent\textbf{State-Of-The-Art Face Recognizers:} Modern face recognizers can be categorized into two groups based on the losses used: (i) Euclidean-distance-based losses such as FaceNet \cite{FaceNet} and CenterFace \cite{CenterFace}; (ii) Angular/cosine-margin-based losses such as ArcFace \cite{ArcFace}, CosFace \cite{CosFace} and AdaptiveFace \cite{AdaptiveFace}. The recognizers from both groups have been considered in physical adversarial attack papers.


\vspace{6px}
\noindent\textbf{Adversarial Glasses:}
Researchers from Carnegie Mellon University \cite{advGlasses} have shown that specially designed spectacle frames can fool even state-of-the-art facial recognition ArcFace \cite{ArcFace}. Not only can the glasses make the wearer essentially disappear to such automated systems, it can even trick them into thinking you’re someone else \cite{advGlasses}. The adversarial pattern on the glasses can be learned similar to the proposed framework in Section~\ref{sec:PhysicalFramework}. The pattern is adversarially mapped on the subjects' faces through aligned faces. The identification loss is back-propagated to update the pattern iteratively. However, \cite{advGlasses} did not directly optimize the pattern, but employed a GAN architecture with a generator to learn the distribution of adversarial patterns. The loss to train the generator $G$ is defined as,
\begin{equation}
\mathcal{L} = \mathcal{L}_G(Z,D) - \kappa \sum_{z \in Z} \mathcal{L}_F(x+G(z)),
\end{equation}
where $\mathcal{L}_G(Z,D)$ aims to generate real-looking (i.e., inconspicuous) outputs that mislead the discriminator, and $\mathcal{L}_F(x+G(z))$ is the identification loss.

Similarly to human detection systems we apply to our framework to describe the components of this loss formulation. We identify $\mathcal{L}_{T} = \mathcal{L}_F(x+G(z))$ as the task loss and $\mathcal{P} = \mathcal{L}_G(Z,D)$ as the physical loss.
The definition of the loss depends on whether the attacker aims to achieve an untargeted misclassification or a targeted one. For untargeted attacks, it is defined as,
\begin{equation}
    \mathcal{L}_F(x+G(z)) = \sum_{i \neq x} F_{c_i}(x+G(z)) - F_{c_x}(x+G(z)).
\end{equation}
For targeted attacks, the identification loss is defined as,
\begin{equation}
    \mathcal{L}_F(x+G(z)) = F_{c_t}(x+G(z))-\sum_{i \neq t} F_{c_i}(x+G(z)).
\end{equation}

By designing an algorithm to iteratively learn the glass pattern to target different output, scientists were able to assume one another’s identities or make the software think they were looking at celebrities. And far from looking like the kind of goofy disguises individuals might have worn to avoid being recognized in the past, these eyeglasses also appear completely normal to other people.

\vspace{6px}
\noindent\textbf{Adversarial Hat:}
Komkov \emph{et al.} \cite{AdvHat} from from Lomonosov Moscow State University and Huawei Moscow Research Center designed a rectangular paper sticker to be stuck on a hat to trick ArcFace in multiple shooting conditions. The same strategy as the framework proposed in Section~\ref{sec:PhysicalFramework}. For adversarial mapping, the sticker is off-plane bent and pitch rotated to match the hat in each image. Subsequently a Spatial Transformer Layer (STL) \cite{STN} is employed to project the obtained sticker on the image of the face.

A batch of adversarial images are employed to train the adversarial sticker iteratively using the following loss function.
\begin{equation}
\label{eq:AdvHatLoss}
\mathcal{L} = L_{sim} (x, a) + \lambda \mathcal{L}_{TV} (x),
\end{equation}
where $\mathcal{L}_T = L_{sim} (x,a)$ is a cosine similarity between the embeddings of $x$ and $a$. $\mathcal{L}_P = \mathcal{L}_{TV} (x)$ is the loss to encourage the smoothness.

\begin{figure}
    \centering
    \includegraphics[width=\columnwidth]{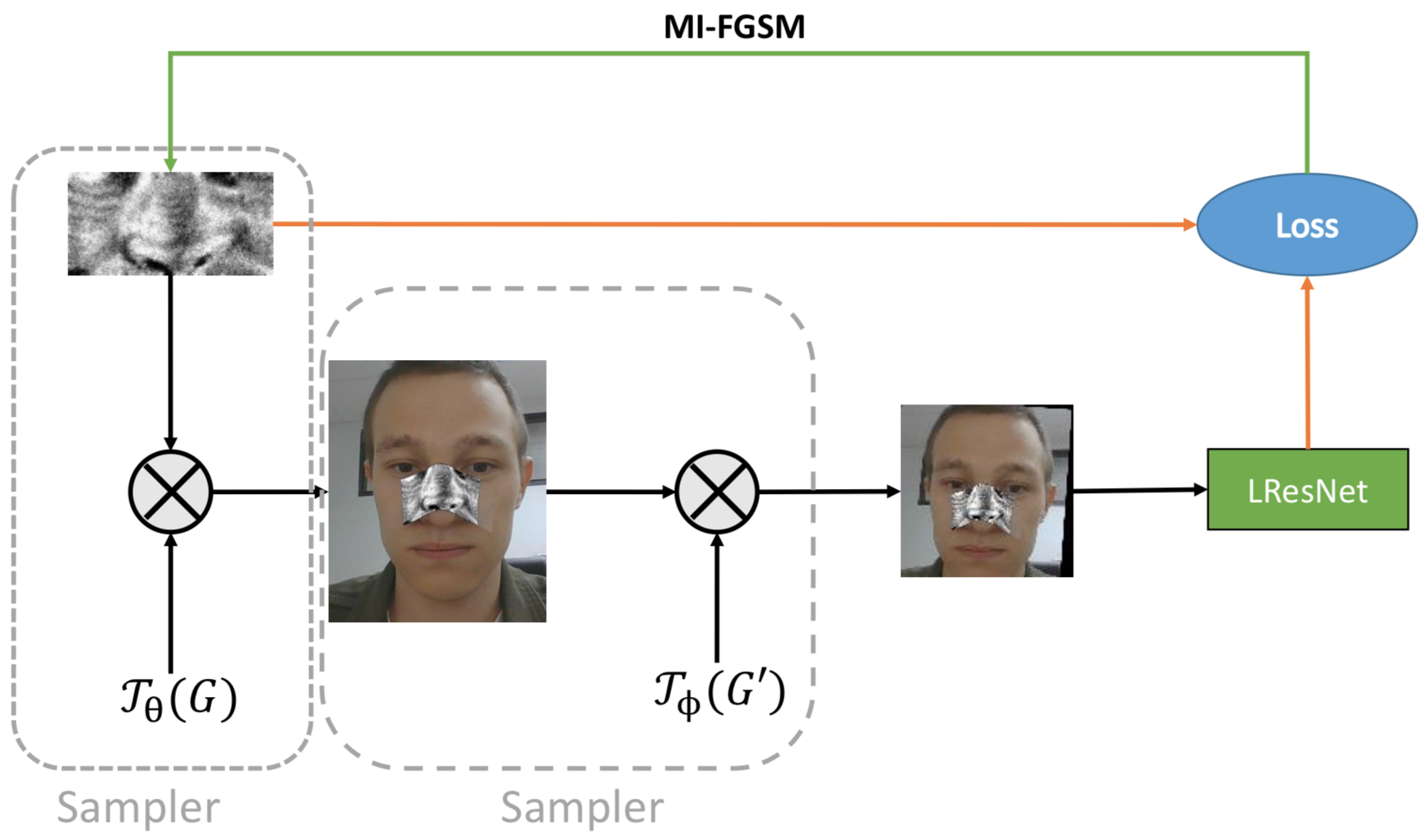}
    \caption{The pipeline in \cite{AdvPatches} to design an printable nose patch to mislead the state of the art ArcFace face recognition system.}
    \label{fig:AdvPatches_Face_Algorithm}
\end{figure}

\vspace{6px}
\noindent\textbf{Adversarial Patches:}
Pautov \emph{et al.} \cite{AdvFacePatches} applied the same strategy as \cite{AdvHat} to generate a broad class of adversarial patches. To learn adversarial mapping, the authors employed a checkboard at the sticker location to model a 2D nonlinear projective transformation $\mathcal{T}_{\theta}(G)$ to map from a 2D sticker to the sticker ``in action'' in the face image. The same loss function with Equation~\ref{eq:AdvHatLoss}. The pipeline is illustrated in Fig.~\ref{fig:AdvPatches_Face_Algorithm}.

More recently, an adversarial 3D patch-based attack is proposed in \cite{yang2023towards} for attacking face recognition systems. Specifically, the authors propose designing adversarial textured 3D mesh which can be 3D printed in order to generate a more elaborate attack. Instead of optimizing the higher dimensional mesh space, which is computationally exhaustive and limits the black-box transferability of the attack, the authors propose the representation of the input face in low-dimensional manifold using 3D Morphable Model coefficients \cite{deng2019accurate} that denotes the identity, expression, illumination, camera position, and texture, generated from a CNN regression model. These coefficients are used during the adversarial optimization and the authors have controlled the local topology such that adversarial generation can be restricted to a desired region such as eyes, eye and nose, etc.

\vspace{6px}
\noindent\textbf{Adversarial Masks:}
Zolfi \emph{et al.} \cite{AdvMask} used 3D face reconstruction to digitally apply an adversarial mask to a facial image. They relied on an UV space, a concept from the 3D mesh domain, to record the position information of the 3D face and provide dense correspondence to the semantic meaning of each point in the UV space. This method allowed them to achieve near-real approximation of the mask on the face, which is essential to the creation of a practical patch. Examples of the adversarial masks are illustrated in Fig.~\ref{fig:AdvMask}.

\begin{figure}
    \centering
    \includegraphics[width=\columnwidth]{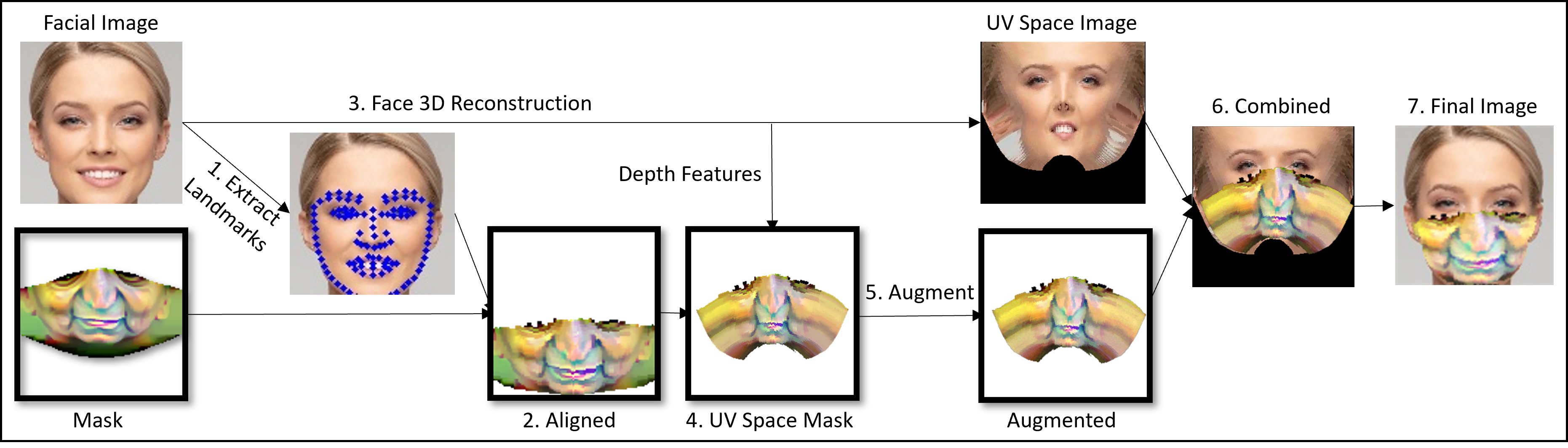}
    \caption{The pipeline to design adversarial masks to attack face recognition \cite{AdvMask}.}
    \label{fig:AdvMask_pipeline}
\end{figure}

As shown in Fig.~\ref{fig:AdvMask_pipeline}, the pipeline of the mask's placement is as follows: given a facial image, they first find the landmark points in order to align the mask with the correct location on the original facial image.
Then, they input the facial image to the 3D face reconstruction model. The output of the model is used for two purposes: (a) to transfer the original image to the UV space, and (b) to extract the face's depth features to transfer our mask to the UV space. Moreover, to improve the robustness of the patch, they randomly apply location- and color-based transform augmentations:
\begin{itemize}
    \item Location-based - Add random translation and rotation to simulate possible distortions in the mask placement on the face in the real world.
    \item Color-based - Add random contrast, brightness, and noise to simulate changes in the appearance of the patch due to various possible factors (e.g., lighting, noise or blurring caused when the camera captures the image).
\end{itemize}
\noindent These transformations are parameterized by $\theta$. Finally, they apply the UV space mask to the UV space facial image and reconstruct the combined image resulting in a masked face image.

\vspace{6px}
\noindent\textbf{Adversarial Makeup:}
The motivation \cite{yin2021adv} is to synthesize imperceptible yet effective adversarial perturbations using eye shadow over the orbital region of faces. The authors argue that the existing adversarial attack generation methods are impractical and ineffective under physical scenarios, and the generated samples are noticeable. Furthermore, they point out that generated examples are model specific, hence, not transferable. As such, the authors have selected adding eye shadow on the source identity as the synthesizing procedure. To alleviate the style and content differences between the source faces and generated eye shadows, they propose a makeup blending method. Better-generalized adversarial perturbations are generated through a novel fine-grained meta-learning attack strategy. 


This framework consists of makeup generation, makeup blending, and makeup attack stages.Specifically, the makeup generation contains a generator, $G$, that synthesizes eye shadow, and a discriminator, $D$, that criticizes the realism of the generated images. It should be noted that $G$ only synthesizes the orbital region of the face and the discriminator arbitrarily receives the real cosmetic orbital images and the synthesized orbital images. The direct overlay of the synthesized orbital region to the source image yields obvious style differences and other noticeable artifacts in the images. Hence a  makeup blending stage is introduced to generate the imperceptible attacks. The authors have introduced a special constraint which shifts the color of the orbital region to match the original image. Furthermore, two VGG-16 based losses, namely, content loss and style loss, are defined to enhance the integration of style and content of the source image to the synthesized orbital region.

In the adversarial attack generation stage the authors of \cite{yin2021adv} have used a series of victim face recognition systems, and a meta-learning strategy is employed to perform impersonating attacks targeting each of these models.  

\vspace{6px}
\noindent\textbf{Adversarial Stickers:}
In a different line of work, an adversarial sticker based perturbation strategy is proposed in \cite{guo2021meaningful}. Most importantly, the authors have used existing (real) stickers and in the attack generation stage they only manipulate the pasting parameters of stickers on the face. Therefore, there is no perturbation design stage in this pipeline. The generation of pasting parameters, is formed into an optimization problem
which they have solved using an evolution algorithm, namely, the Region based Heuristic Differential Algorithm. Fig.~\ref{fig:AdvSticker} shows some sample attacks for the face identification task. 

\begin{figure}
    \centering
    \includegraphics[width=\columnwidth]{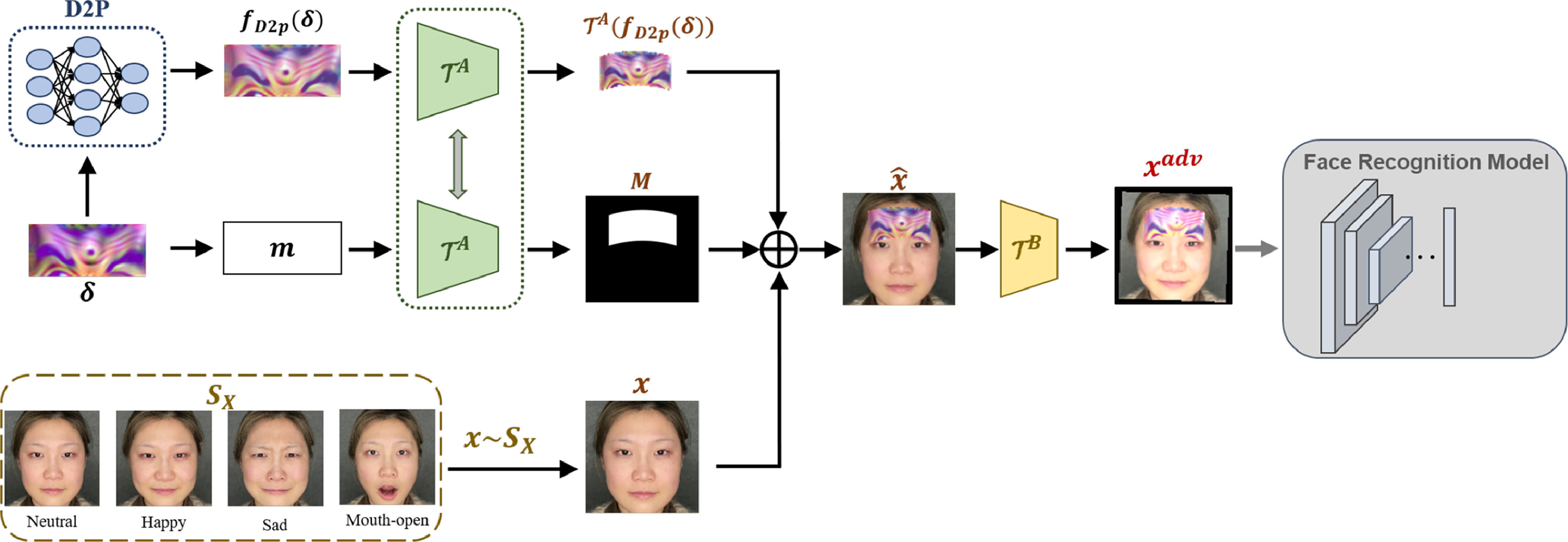}
    \caption{Robust physical adversarial attack generation method of PadvFace \cite{zheng2022robust}. }
    \label{fig:AdvSticker_2}
\end{figure}

More recently, another sticker-based physical attack method, dubbed PadvFace, is proposed in \cite{zheng2022robust}. In particular, this method explores various physical-world conditions and propose a novel curriculum adversarial attack algorithm that is capable of optimizing the model to cope with different attack difficulties. The proposed framework is shown in Fig.~\ref{fig:AdvSticker_2}. First, the synthesized sticker will go through a Digital-to-Physical (D2P) module which will mimic the chromatic aberration induced by printers and cameras. The authors also incorporate two transformation modules, namely, $\tau^A$ and $\tau^B$. Specifically, $\tau^A$ induces parabolic transformation, rotation, and translation that can occur when pasting the printed sticker on a real-world face. In contrast, $\tau^B$ simulates environmental variations such as different poses and lighting conditions, that would impact the visibility of the adversarial sticker when capturing it through a camera. Hence the final adversarial image is the output after sequentially passing the synthesized patch through those transformation modules and digitally pasting that image on the attacker's face.

The authors indicate that due to the complexity with respect to physical attacks and the high nonlinearity of DNNs a non-convex optimization problem can be rendered and the model could get stuck in local minima. As such, a novel curriculum learning algorithm is proposed that gradually optimizes adversarial stickers from easy to complex physical-world conditions. Specifically, the authors denote an adversarial sticker as $\delta$, k loss of the adversarial attack under the condition $k_i$ as $\mathcal{L}_{sim, k_i}$, then the authors propose to assign a learnable weight parameter $p_i \in [0, 1]$ for each physical world transformation, $k_i$. Then they formulate the objective of their curriculum learning algorithm as, 

\begin{equation}
    \min_{\delta,p_i \in [0, 1]} \frac{1}{n} \sum_{k_i \in K} [p_i \mathcal{L}_{sim, k_i} + \lambda g(p_i)] + \alpha \mathcal{L}_{TV}(\delta),
\end{equation}
where $g(p_i) = \frac{1}{2}p_i^2 - p_i$ is a regularizer and $\lambda > 0$ is a curriculum parameter. Similar to prior works $L_{TV}$ is introduced to enhance the smoothness of the generated sticker. When describing the components in this loss we identify $\mathcal{L}_T = \mathcal{L}_{sim, k_i}$ as the task loss and $\mathcal{L}_P =  alpha \mathcal{L}_{TV}(\delta)$ as the physical loss.

The evaluation results indicate that this proposed method is resilient to complex physical-world variations and can generate effective dodging and impersonation physical attacks. 

\vspace{6px}
\noindent\textbf{3D Face Recognition Attacks \cite{Adv3Dface}:} Li \emph{et al. } propose adversarial illumination based attack structured-light-based 3D face recognition system. The authors demonstrate that the perturbed light will be shifting the point cloud and can cause dodging or impersonation attacks. Specifically, the the projected patterns are modified to pollute the captured 3D data indirectly. The loss of the proposed framework can be defined as,
\begin{equation}
    L = L_{adv} + \lambda Sen  \cdot |\phi_a' - \phi_a|,
\end{equation}
where $ \cdot $ is Hadamard products. Based on the observation that humans are more sensitive to perturbation in the central and flat areas of the human face, $Sen$ is regularisation term that punish perturbations in areas of high sensitivity. $L_{adv}$ is the loss of the face recognition model. The authors have successfully attacked both point-cloud-based and depth-image based 3D face recognition models. 


\vspace{6px}
\noindent\textbf{Non-texture Attacks - Adversarial Cap:}
Different from popular approaches in learning texture of adversarial patches and accessories, scientists at China’s Fudan University \cite{AdvCap} are learning the positions on a human face on which infrared LEDs can be projected to trick the facial recognition FaceNet \cite{FaceNet}. The researchers found that they can learn the infrared patterns to trick FaceNet into thinking they were Moby, which is the perfect disguise to baffle your boss – so long as you’re prepared to be mistaken for a musician who claimed to have dated Natalie Portman but was then obliged to retract and apologize for the suggestion \cite{AdvCap}. Frearson \emph{et al.} \cite{QUTAdvLED} employed a Light Perturbation Optimizer (LPO) algorithm to search for the optimal positions of the light spots on the attacker’s face.

\vspace{6px}
\subsubsection{Person Re-ID}\hfill

\noindent{Person Re-Identification (Re-ID)} is the task of retrieving a person of interest across multiple non-overlapping cameras. Given a query person-of-interest, the goal of Re-ID is to determine whether this person has appeared in another place at a distinct time captured by a different camera, or even the same camera at a different time instant \cite{PersonReIDsurvey}.

\vspace{6px}
\noindent\textbf{State-Of-The-Art Person Re-Identifiers:} Modern person re-identifiers employ various losses to train backbone networks. For example, CTL \cite{CTL} employed a centroid loss, FAT \cite{TripletLossReID} employed a triplet loss, stReID \cite{stReID} and OSNet \cite{OSNetPAMI} employed classification losses. Both triplet-loss and classification-loss approaches have been attacked in physical adversarial attack papers.

\begin{figure}
    \centering
    \includegraphics[width=\columnwidth]{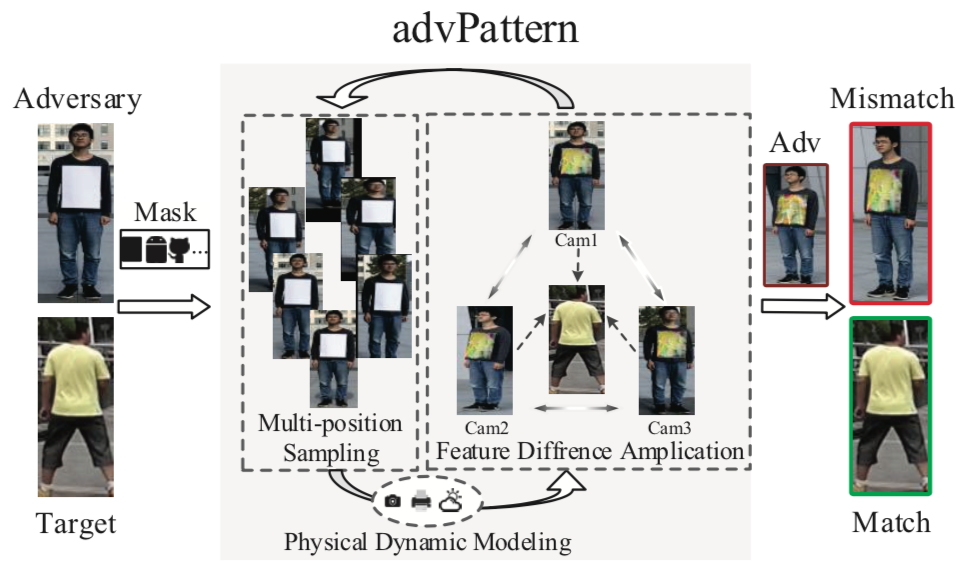}
    \caption{The pipeline to train an adversarial patch to attack real-world person re-identification systems \cite{AdvPersonReID}. }
    \label{fig:AdvPersonReID}
\end{figure}

\vspace{6px}
\noindent\textbf{AdvPatterns:} Wang \emph{et al.} \cite{AdvPersonReID} designed an adversarial patch to attack person re-identification systems. The adversarial patch is adversarial mapped onto the body of each subject and learned by minimizing the Re-ID task loss and physical loss similar to the proposed framework in Section~\ref{sec:PhysicalFramework}. The task loss aims to minimize the similarity between two images of the same subject from two different cameras and maximize the similarity between two images of the same subject from the same camera. Total Variation (TV) is used as the physical loss. The optimization problem of non-target attacks is formulated as,
\begin{eqnarray}
\underset{p}{argmin} \underset{(x_k^o,x_k^-,x_k^+)}{E} f_{\theta} (\varphi(x_k^o)',\varphi(x_k^{-})') \\
- \beta f_{\theta} (\varphi(x_k^o)',\varphi(x_k^{+})') + \kappa TV(p),
\end{eqnarray}
with $(x_k^o,x_k^-,x_k^+)$ is the triplet where $x_k^o$ and $x_k^+$ are two images of the same person from the same camera, and $x_k^-$ is the same person from a different camera. The adversarial mapping is performed through a pre-defined mask for each image. The function $\varphi(.)$ acts as physical imaging emulation by randomly transforming the image. The pipeline is illustrated in Fig.~\ref{fig:AdvPersonReID}. 

\begin{figure}
    \centering
    \includegraphics[width=\columnwidth]{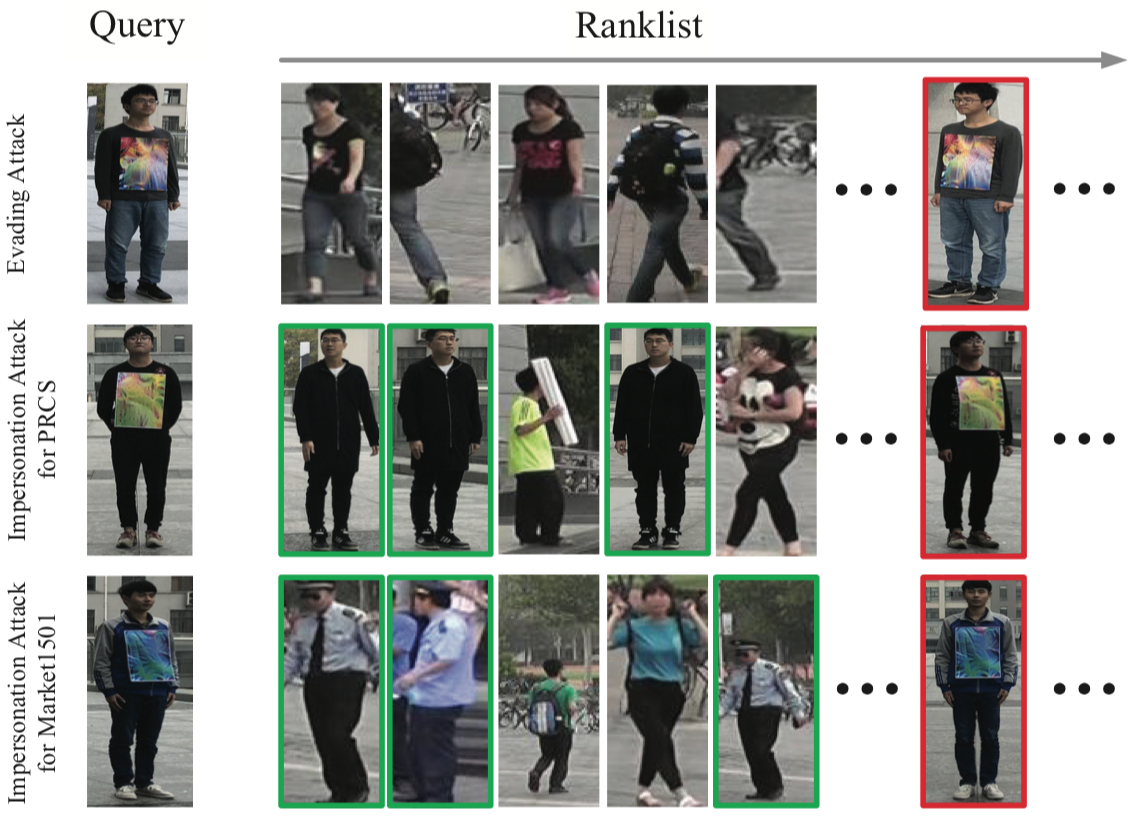}
    \caption{The adversarial patch successfully misled the state of the art person re-identification systems \cite{AdvPersonReID,zheng2016discriminatively}. }
    \label{fig:AdvPersonReID_Examples}
\end{figure}

The experimental results show that the rank-1 accuracy of Re-ID models \cite{zheng2016discriminatively} for matching the adversary decreases from 87.9\% to 27.1\% under Evading Attack. Furthermore, the adversary can impersonate a target person with 47.1\% rank-1 accuracy and 67.9\% mAP under Impersonation Attack. Examples of the adversarial patch functions in real world are illustrated in Fig.~\ref{fig:AdvPersonReID_Examples}.

\vspace{6px}
\noindent\textbf{Differential multi-shot sampling:} While most digital approaches aims to add minimal perturbations to mislead person Re-ID models \cite{DigitalAdvPersonReID,DigitalAdvPersonReID2,DigitalAdvPersonReID3}, they are not transferable to the physical domain due to no control of the adversary on the digital images. However, there is one digital approach that allows to control the number of malicious pixels by using differential multi-shot sampling \cite{TCIAdvPersonReID}. 
Ability to control the number of malicious pixels enables turning existing digital adversarial attacks approaches into physical attacks by limiting the number of pixels similar to adversarial patches or conform to the shape of adversarial accessories, e.g. bags, shirts, pants, to be implemented.

\subsection{Surveillance Human Tracking Attacks}\hfill

\noindent{Human tracking aims at locating and following single or multiple humans over time in a single camera \cite{TrackingSurvey}. In surveillance, human tracking is critically important to maintain performance of surveillance systems against the adverse imaging condition and the dynamics of subjects in the scene.}

\vspace{6px}
\noindent\textbf{State-Of-The-Art Trackers:} Modern object trackers can be categorized into two groups based on their objective functions: (i) classification-based such as SiamRPN++ \cite{SiamRPNpp,SiamMaskE} and Ocean \cite{Ocean}; and (ii) regression-based such as GOTURN \cite{GOTURN}.

\begin{figure}
    \centering
    \includegraphics[width=\columnwidth]{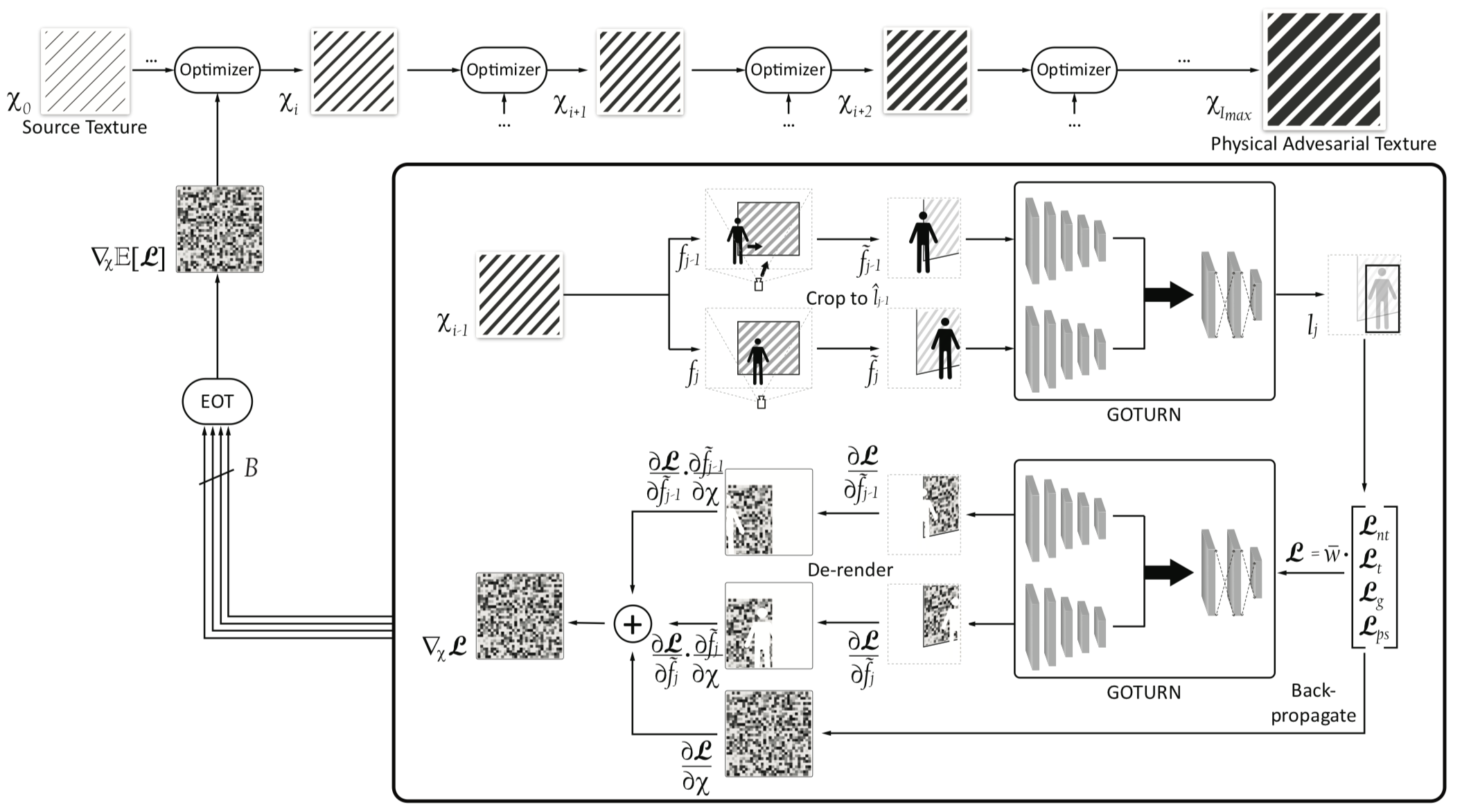}
    \caption{The pipeline to train an adversarial poster to attack human tracking \cite{AdvTracking}. }
    \label{fig:AdvPosters_PAT}
\end{figure}

\vspace{6px}
\noindent\textbf{PAT - Adversarial Posters:} Wiyatno \emph{et al.} \cite{AdvTracking} proposed to learn physical adversarial textures (PAT) to be printed on posters or displayed on TV on the background to confuse realworld tracking systems such as GOTURN \cite{GOTURN}. The proposed adversarial learning process is similar to the framework proposed in Section~\ref{sec:PhysicalFramework}. The PAT is learned through iteratively updated by back-propagating the tracking loss from the tracker as illustrated in Fig.~\ref{fig:AdvPosters_PAT}.

\begin{figure}
    \centering
    \includegraphics[width=\columnwidth]{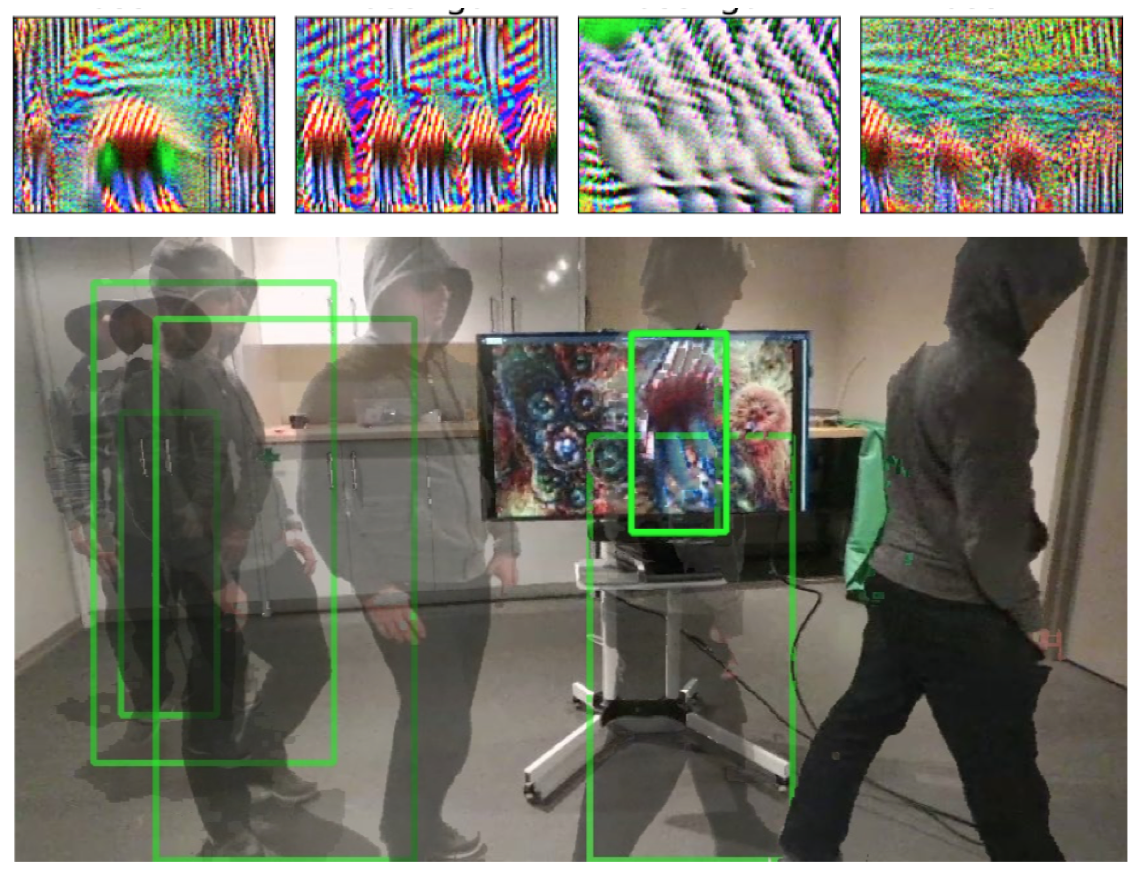}
    \caption{Displaying the adversarial poster on a TV screen can successfully mislead the human tracker GOTURN \cite{AdvTracking}.}
    \label{fig:AdvPosters_PAT_Results}
\end{figure}

The losses are calculated based on EOT to extend the diversity of transformation and environment conditions. 
\begin{equation}
\mathcal{L} = [\mathcal{L}_{nt},\mathcal{L}_{t},\mathcal{L}_{g},\mathcal{L}_{ps}]    
\end{equation}
where 
\begin{itemize}
    \item $\mathcal{L}_{nt}$ is the non-targeted loss which maximizes the victim model's training loss, thus causing it to become generally confused (e.g. FGSM \cite{FGSM});
    \item $\mathcal{L}_{t}$ is the targeted loss which also applies to the victim model’s training loss, but to minimize the distance to an adversarial target output (e.g. JSMA \cite{JSMA});
    \item $\mathcal{L}_{g}$ is the guided loss to regulate specific adversarial attributes rather than strict output values, analogous to misclassification onto a set of output values \cite{CellphoneAttacks};
    \item and $\mathcal{L}_{ps}$ is a Lagrangian-relaxed loss to enforce perceptual similarity.
\end{itemize}
Examples of the patches learned and how they successfully thwarted the GOTURN tracker are depicted in Fig.~\ref{fig:AdvPosters_PAT_Results}.

\vspace{6px}

\noindent\textbf{Universal Physical Attacks on Single Object Tracking \cite{ding2021towards}: } In this work, the authors look at adversarial attacks on single object tracking from a different perspective. The authors identify that the core objective of single object tracking lies in the feature matching between the search image and templates, and propose to design novel perturbations using Maximum Textural Discrepancy (MTD). Specifically, let $z^t_{\delta}$ denote the exemplar image, $x^{(t, s)}_{\delta}$ denote the search image and $\varphi$ denote the feature extractor of Siamese network. Then, the MTD loss is defined as,  
\begin{equation}
\begin{split}
    \mathcal{L}_{MTD} (z^t_{\delta}, x^{(t, s)}_{\delta}) & = - \frac{1}{D} \sum_{d \in D} || g(\varphi_d(z^t_{\delta})) \\
    - & g(\varphi_d(\omega(x^{(t, s)}_{\delta})))||_F,
\end{split}
\end{equation}
where $F$ is the Frobenius norm, $g$ represents the Gramian matrix operator and $d = [1, \ldots, D]$ is the layer id within the Siamese network. This attack framework is further extended using a shape attack objective. Specifically, the authors show that motion model penalization schemes within SiamMask and SiamRPN++ penalize unstable position predictions. Due to this penalty, it is challenging to misguide trackers and as a solution, a shape attack objective is introduced. This loss, $\mathcal{L}_{sha}$, is written as,
\begin{equation}
\begin{split}
\mathcal{L}_{sha} (z^t_{\delta}, x^{(t, s)}_{\delta}) & = \frac{1}{K} \sum_{k=1}^K max(|\tilde{h}_k^{(s, t)} - {h'}|_1 \\
& + |\tilde{w}_k^{(s, t)} - w'|_1, m_{\tau}),
\end{split}
\end{equation}
where $K$ is the selected top-K bounding boxes, $(\tilde{h}_k^{(s, t)}, \tilde{w}_k^{(s, t}))$ denotes the shape of the selected bounding box, $(h', w')$ is the shape of the targeted bounding box, and $m_{\tau}$ is the regression margin. To ensure the physical feasibility the authors have added the total variation-based smoothness loss, $\mathcal{L}_{TV}$, to their overall objective. Now the final objective can be written as,
\begin{equation}
  \mathcal{L}(z^t_{\delta}, x^{(t, s)}) =  \alpha \mathcal{L}_{MTD} (z^t_{\delta}, \beta x^{(t, s)}_{\delta}) + \gamma \mathcal{L}_{sha} (z^t_{\delta}, x^{(t, s)}_{\delta}) + L_{TV}, 
\end{equation}
where $\alpha, \beta$, and $\gamma$ are the weights for the respective loss functions. Within this framework $\mathcal{L}_T = \mathcal{L}_{MTD} (z^t_{\delta}, \beta x^{(t, s)}_{\delta}) + \gamma \mathcal{L}_{sha} (z^t_{\delta}, x^{(t, s)}_{\delta})$ is the task loss while physical loss $\mathcal{L}_P$ is $\mathcal{L}_{TV}$.

The proposed framework has been vigorously tested in different real-world testing settings and the experimental results demonstrate that this framework can significantly degrade the visual trackers' performances.

\vspace{6px}
\noindent\textbf{Hijacking with Adversarial Patches: } A number of approaches in the literature \cite{AdvMOT,HijackingTracker} proposed to hijack modern trackers by adding one adversarial patch on one frame \cite{AdvMOT} or several frames \cite{HijackingTracker}. While these approaches only investigated digital attacks, they can be extended to physical attacks by allowing the adversary to putting on the adversarial patch for a period of time.

\subsection{Surveillance Action Recognition Attacks}
\noindent{Human action recognition} aims to recognize activities from a series of observations on the actions of subjects and the environmental conditions \cite{ActionRecognitionSurvey}. The goal is to analyze a video to identify the actions taking place in the video. In addition to the spatial content in each frame, the temporal information is of essence to an effective recognition approach.

\vspace{6px}
\noindent\textbf{State-Of-The-Art Action Recognizers: } Modern action recognizers can be categorized into two groups based on how they process temporal information: (i) Two-stream approaches such as  ST-GCN \cite{STGCN}; and (ii) 3D CNNs approaches such as R2+1D-BERT \cite{R21DBERT}, and I3D \cite{I3D}. Both two-stream and 3D-CNN approaches have been attacked in physical setting.

\begin{figure}
    \centering
    \includegraphics[width=\columnwidth]{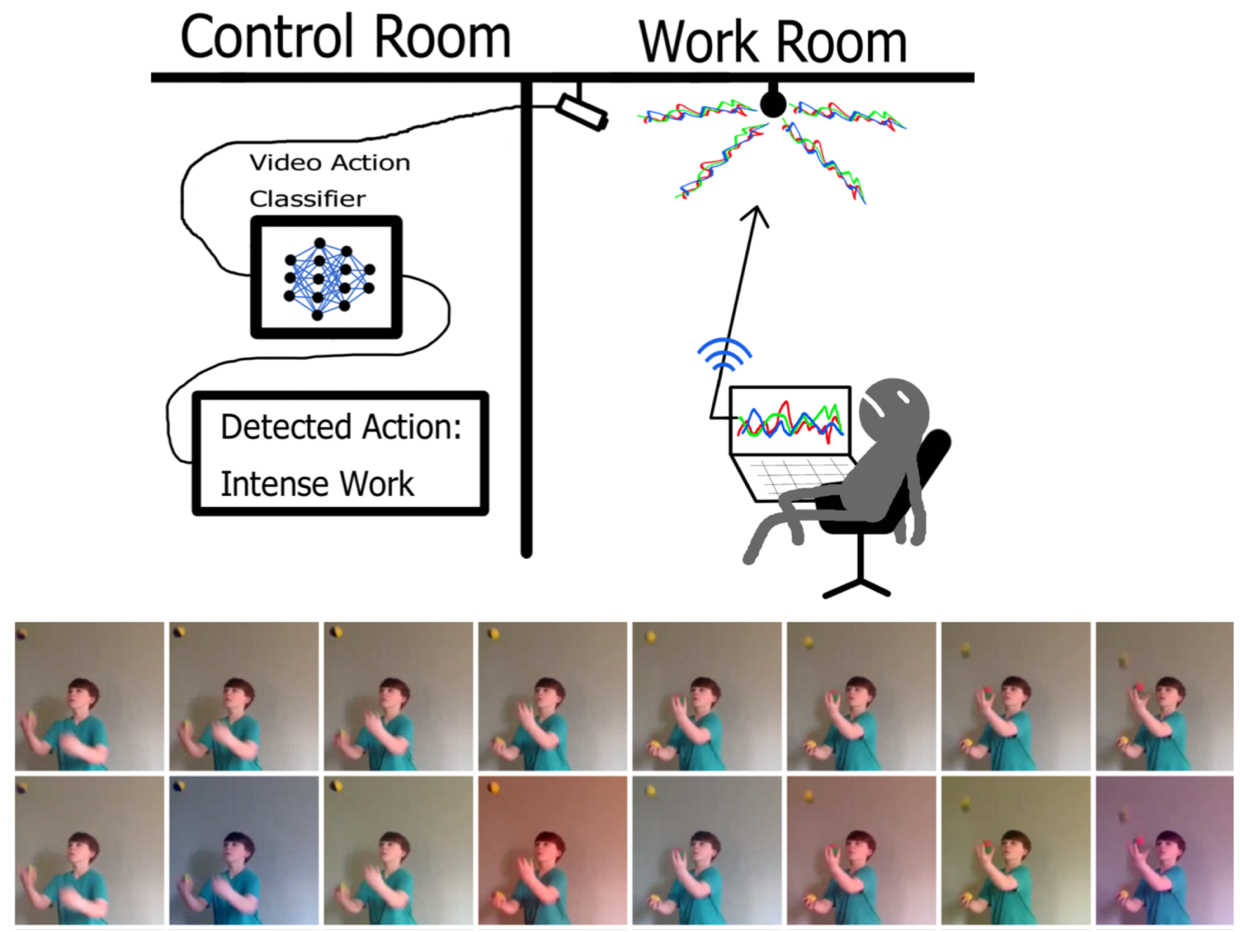}
    \caption{Projecting adversarial temporal patterns by flickering a RGB LED bulb can mislead at a 93\% fooling ratio over multiple state of the art action recognizers such as I3D, R3D and MC3 \cite{PhysicalAdvAction}.}
    \label{fig:FlickeringPerturbation}
\end{figure}

\vspace{6px}
\noindent\textbf{Over-the-Air Adversarial Flickering Attacks: } There exists only one work in recent literature which manages to perform adversarial attacks on action recognition models in the physical world. Pony \emph{et al.} \cite{PhysicalAdvAction} proposed ``flickering perturbations''. A flickering perturbation is a series of uniform RGB perturbations, each for one frame in the video, thus constructing a temporal adversarial pattern. Each uniform RGB perturbation which is applied to a single frame does not contain any spatial information other than a constant offset. The constant offset is practically unnoticed by the human observer. In the physical world, the flickering perturbations can be implemented by a RGB led bulb controlled over Wifi as shown in Fig.~\ref{fig:FlickeringPerturbation}.

The flickering perturbation can be learned by the following objective function,
\begin{equation}
    \underset{\sigma}{argmin} \frac{1}{N} \mathcal{L}(F_0(X_n+\sigma),t_n) + \lambda \sum_{j} \beta_j D_j(\sigma)
\end{equation}
where the first terms is the adversarial loss and the second terms are regularization terms. $N$ is the number of training videos, $X_n$ is the $n_{th}$ video, $F_0(.)$ is the action classifier output, and $t_n$ is the targeted label in targeted attacks or any label except the genuine label in untargeted attacks. The authors have used a set of regularization terms $D_j(.)$ to control the distortion introduced by the perturbations and ensure that the adversarial noise is imperceptible to a human observer. As such, we can denote $\mathcal{L}_T = \mathcal{L}(F_0(X_n+\sigma),t_n)$ as the task loss and $ \mathcal{L}_P =\lambda \sum_{j} \beta_j D_j(\sigma)$ as the physical loss term. 

\subsection{Adversarial Attacks beyond Visible}
While most of existing work has been focused on visible images, images from other spectrum have also been used in surveillance and attention to adversarial attacks on these spectrum beyond visible is emerging. 

\begin{figure}
    \centering
    \includegraphics[width=\columnwidth]{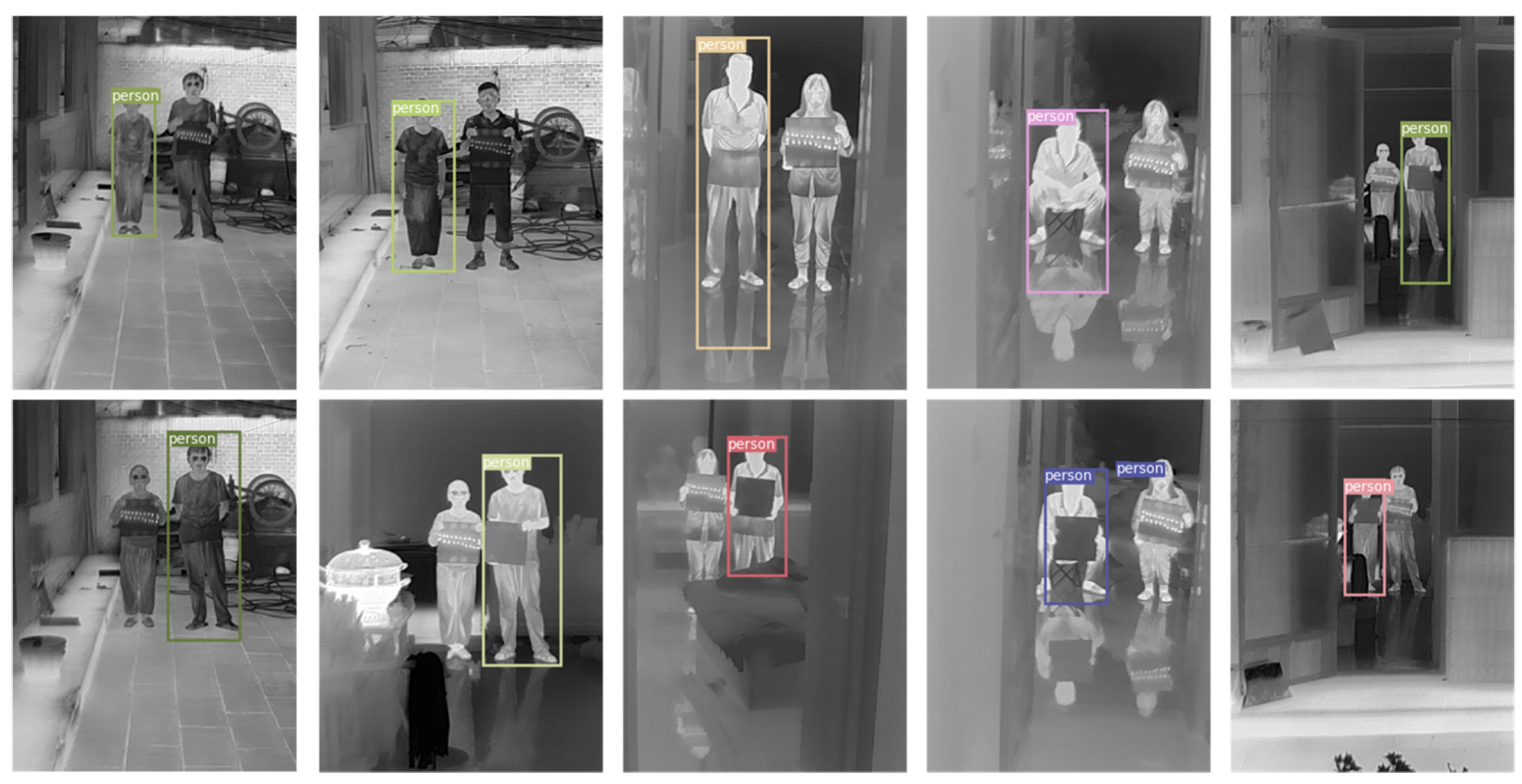}
    \caption{A cardboard with a specific layout of bulbs can function as an adversarial patch to mislead human detection in thermal images \cite{AdvThermal}.}
    \label{fig:AdvThermalPhysical}
\end{figure}

\vspace{6px}
\noindent\textbf{Infrared: } in surveillance, infrared imaging plays a similarly important role as visible imaging, especially for human detection and action recognition due to their advantages that enables 24/7 and adverse-imaging-condition surveillance.
There exist only a few works on attacking the thermal surveillance task of human detection and one of these methods is \cite{AdvThermal} in 2021. The authors proposed to use a set of small bulbs on a cardboard to generate infrared adversarial patches. Since the bulbs are visible in the thermal images, they function similar to an adversarial patch in the visible framework. The layout of bulbs is learned via a framework similar to Section~\ref{sec:PhysicalFramework} to mislead thermal human detectors, i.e. YOLOv3. They also employ a combined loss from the detection loss, $\mathcal{L}_T$, and the TV loss, $\mathcal{L}_{TV}$. Examples of the bulb cardboard learned and how they successfully thwarted the YOLOv3 detector are depicted in Fig.~\ref{fig:AdvThermalPhysical}.

\begin{figure}
    \centering
    \includegraphics[width=\columnwidth]{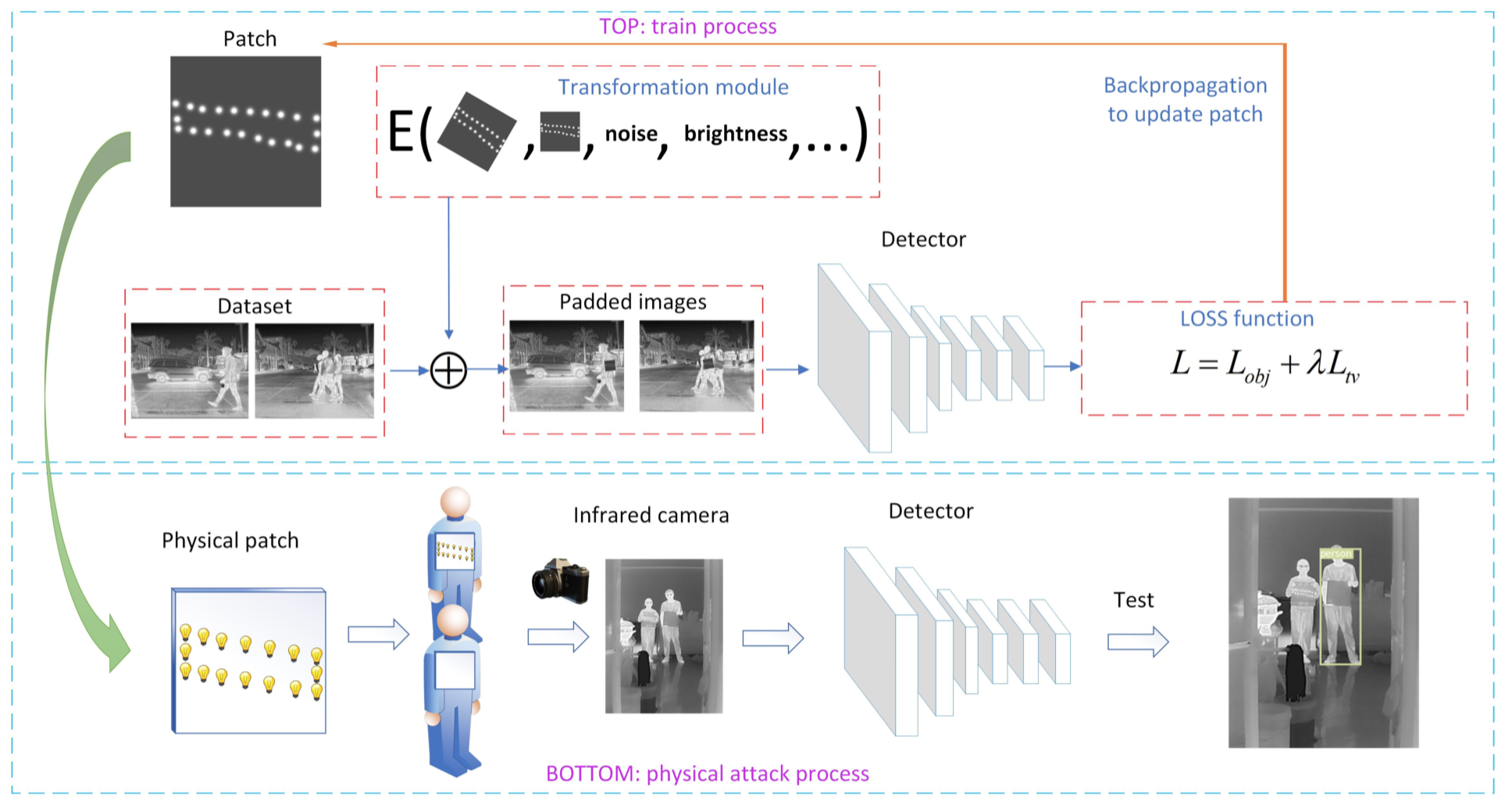}
    \caption{The pipeline to design a specific layout of a cardboard of bulbs to attack the human detection in thermal images \cite{AdvThermal}.}
    \label{fig:AdvThermal}
\end{figure}

Adversarial Infrared method proposed in \cite{AdvInfrared}. A basic pattern that can be expanded periodically is designed and the goal is to make the pattern retain the adversarial effect even after random cropping and deformations. First, the adversarial pattern is mapped to a cloth that a human wears using simulation and an  adversarial ``QR code'' pattern is learned. Then the adversarial cloth is manufactured using a material called aerogel which has good thermal insulation stability at room temperature. Their overall loss is defined as,
\begin{equation}
L = L_{obj} + \lambda L_{black},
\end{equation}
where $L_{obj}$ denotes the object score of the object detector and $L_{black}$ is the average probability of black pixels appearing in patch. The $L_{black}$ is used to control the amount of heat-insulating material (i.e aerogel) used in the clothing. The authors show that in addition to saving material it also improves the air permeability and comfort of the clothes. $\lambda$ is a parameter that controls the contribution from $L_{black}$. This $L_{black}$ term can be seen as the physical loss while the $L_{obj}$ is the task loss.  

The authors of \cite{wei2023physically} propose a method for designing thermal radiation insulation patches to manipulate the emitted thermal distribution. The optimization procedure involves manipulating the shape and location of infrared patches on the target object. To ensure the practicality of the attack, certain constraints are imposed on the optimization process: (i) all pixels in the mask must be connected, and (ii) the mask can have either a value of one (indicating the presence of insulating material) or zero. To achieve this, the authors introduce aggregation regularization to measure the proximity of an activated point to its neighbors, and a binary regularization term is added to calculate the distance between pixel values within the mask and a matrix of ones, for values above a specific threshold. The overall loss of their framework can be defined as,
\begin{equation}
L = L_{obj} + \lambda_1 L_{binary} +  \lambda_2 L_{agg},
\end{equation}
where  $L_{obj}$ denotes the object score of the object detector, $L_{binary}$ is the binary regularization loss and $L_{agg}$ is the aggregation regularization loss. $\lambda_1$ and $ \lambda_2$ are weights controlling the contribution of the specific loss terms. 



\vspace{6px}
\noindent\textbf{LiDAR: } Due to the emerging of affordable LiDARs, LiDARs are emerging into surveillance since they can enhance security by reducing false alarms, allowing for real time tracking of intruders, and enabling automated PTZ camera control for a more comprehensive security system \cite{church2018aerial}.  Cao \emph{et al.} has shown how the objects in the physical world can be 3D printed such that they would mislead the LiDAR-based detection \cite{AdvLiDAR} and Tu \emph{et al.} \cite{AdvLiDAR2} have shown success in learning 3D adversarial objects which can be placed on roadside to mislead Baidu Apollo’s LiDAR-based detection system. Tu \emph{et al.} \cite{AdvLiDAR2} modeled a 3D adversarial object as a mesh and learned it by minimizing the detection results and the Laplacian loss for mesh smoothness. The adversarial objects managed to mislead state of the art object detectors for LiDARs such as PIXOR \cite{PIXOR}, PointRCNN \cite{PointRCNN} and PointPillars \cite{PointPillars}. The adversarial 3D objects produced in \cite{AdvLiDAR} succeeded in avoid being detected while being placed on the side of a road by the LiDAR-based detection system in the Baidu Apollo autonomous driving platform. However, it should be noted that \cite{AdvLiDAR, AdvLiDAR2} frameworks have only been tested for 3D printed adversarial objects and have not been evaluated in the human surveillance setting. 

Furthermore, we would like to acknowledge the emerging works on attacking multimodal perception systems used in autonomous vehicles. For instance, in \cite{abdelfattah2021adversarial} the authors propose to render a mesh-based 3D adversarial object that can fool both RGB and LiDAR perception systems. To demonstrate the applicability of this system for both cascaded and fusion-based multimodal perception frameworks the authors have used Frustum-PointNet (F-PN) and EPNet victim models. The evaluations done using KITTI benchmark shows the viability of this framework, however, the evaluations are limited to digital attacks and this method has not been validated for attacking human surveillance. In \cite{tu2021exploring} authors propose to fool both RGB and LiDAR perception systems of autonomous vehicles. They introduce two adversarial objectives: one for suppressing true bounding boxes and the other for generating false bounding box proposals that avoid overlapping with any ground truth boxes in the scene. Additionally, a regularization loss is incorporated to promote smooth object surfaces. We note that the evaluations are limited to digital attacks and this method has not been validated for attacking human surveillance. 

\begin{figure}
    \centering
    \includegraphics[width=\columnwidth]{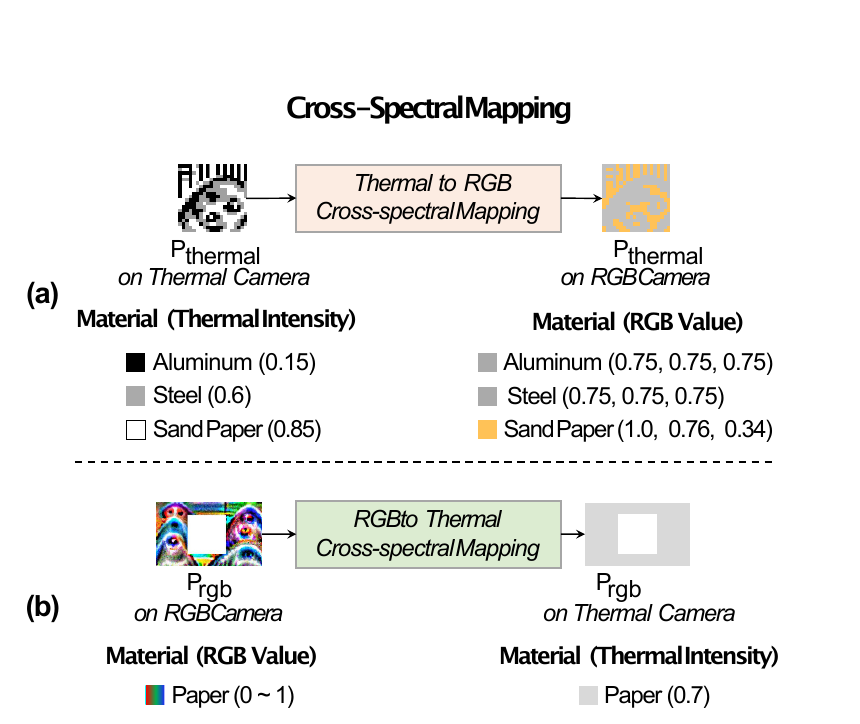}
    \caption{ Visual explanation of Cross-spectral Mapping \cite{kim2022map}.}
    \label{fig:Advmultispectral2}
\end{figure}

\begin{table*}[htbp]
\caption{A summary of the analysis physical adversarial attack methods that are discussed in the paper using the proposed analysis framework.}
\resizebox{\textwidth}{!}{%
\begin{tabular}{|p{2cm}|p{4cm}|p{2cm}|p{2cm}|p{2cm}|p{6cm}|p{6cm}|}
\hline
\textbf{Surveillance Task}                 & \textbf{Method}                                                        & \textbf{Accessory}                & \textbf{Adversarial Mapping}                    & \textbf{Physical Imaging Emulation} & \textbf{Task Loss}                                                                                                                                                                         & \textbf{Physical Loss}                                                                                                                                                  \\ \hline
\multirow{10}{*}{Human Detection} & Thys et al. \cite{AdvPatches}                                          & Patch                    & No                                     & Yes                        & detection loss of the YOLO detector                                                                                                                                               & non-printability score + total variation                                                                                                                       \\ \cline{2-7} 
                                  & ShapeShifter \cite{ShapeShifter}                                         & Patch                    & No                                     & Yes                        & loss of the classifier                                                                                                                                                            & None                                                                                                                                                           \\ \cline{2-7} 
                                  & UPC \cite{UPC}                                                  & Patch                    & No                                     & No                         & detection loss of the Faster RCNN detector                                                                                                                                        & total variation                                                                                                                                                \\ \cline{2-7} 
                                  & Facebook Invisibility Cloak  \cite{AdvTshirtFB}                        & t-shirt                  & No                                     & Yes                        & detection loss of the object detector                                                                                                                                             & total variation                                                                                                                                                \\ \cline{2-7} 
                                  & Baidu Invisible Cloak \cite{BaiduAdvTShirt}                               & Patch                    & No                                     & Yes                        & detection loss of the object detector + regulization loss (in the C\&W optimization method {[}17{]})                                                                              & total variation                                                                                                                                                \\ \cline{2-7} 
                                  & IBM T-shirt \cite{AdvTShirtIBM}                                         & t-shirt                  & deformation                            & Yes                        & detection loss of the object detector                                                                                                                                             & total variation                                                                                                                                                \\ \cline{2-7} 
                                  & Naturalistic Physical Adversarial Patch \cite{hu2021naturalistic}              & Patch                    & No                                     & Yes                        & detection loss of the object detector                                                                                                                                             & total variation                                                                                                                                                \\ \cline{2-7} 
                                  & Adversarial Texture  \cite{AdvTexture}                                         & clothing                 & No                                     & Yes                        & detection loss of the object detector                                                                                                                                             & No         
                                   \\ \cline{2-7} 
                                  & Adversarial Camouflage Texture \cite{hu2023physically}                                           & 3D clothing                 & Yes                                     & Yes                        & detection loss of the object detector                                                                                                                                             & Concentration loss                                                                     
                                  \\ \cline{2-7} 
                                  & Infrared Bulbs  \cite{AdvThermal}                                              & bulbs                    & No                                     & No                         & detection loss of the object detector                                                                                                                                             & total variation                                                                                                                                                \\ \cline{2-7} 
                                  & Adversarial Infrared  \cite{AdvInfrared}                                        & clothing                 & 3D clothing                            & yes                        & detection loss of the object detector                                                                                                                                             & probability of black pixels appearing in patch  
                                  \\ \cline{2-7} 
                                  & Physically Adversarial Infrared Patches \cite{wei2023physically}                                         & Patch                 & No                            & No                        & detection loss of the object detector                                                                                                                                             & binary regularization loss and aggregation regularization loss
                                  \\ \hline
\multirow{8}{*}{Face Recognition} & Adversarial Glasses \cite{advGlasses}                                  & Eye glasses              & No                                     & No                         & generator loss (i.e., inconspicuous outputs that mislead the discriminator)                                                                                                       & identification loss (i.e. untargeted misclassification or targeted misidentification)                                                                          \\ \cline{2-7} 
                                  & Adversarial Hat \cite{AdvHat}                                      & Hat                      & No                                     & Yes                        & cosine similarity between input and target face embeddings                                                                                                                        & total variation                                                                                                                                                \\ \cline{2-7} 
                                  & Adversarial Patches \cite{AdvFacePatches}                                  & Patch                    & 2D nonlinear projective transformation & No                         & cosine similarity loss between an embedding of the input face with the adversarial patch and initial embedding of the face                                                        & total variation                                                                        \\ \cline{2-7} 
                                  & 3D mesh    \cite{yang2023towards}                             & Patch                    &  3D Morphable Model & No                         & identification loss                                                        & No                                                                                                                                                                                 \\ \cline{2-7} 
                                  & Adversarial Masks \cite{AdvMask}                                   & Mask                     & 3D face reconstruction                 & Yes                        & cosine similarity loss between an embedding of the input face with the adversarial patch and initial embedding of the face                                                        & total variation                                                                                                                                                \\ \cline{2-7} 
                                  & Adversarial Makeup \cite{yin2021adv}                                           & Makeup                   & Makeup Blending                        & Yes                        & generator loss (i.e., inconspicuous outputs that mislead the discriminator) + misidentification loss                                                                              & style loss and content loss (i.e pushing the style of eye-shadow patches closer to the source image and also preserving the content of the synthesized region) \\ \cline{2-7} 
                                  & Adversarial Sticker \cite{guo2021meaningful}                                          & Sticker                  & 3D blending transformations            & Yes                        & identification loss (i.e. untargeted misclassification or targeted misidentification)                                                                                             & None                                                                                                                                                           \\ \cline{2-7} 
                                  & Non-texture Attacks - Adversarial Cap:\cite{AdvCap}             & Cap                      & No                                     & No                         & loss of the face recognition model                                                                                                                                                & None                                                                                                                                                           \\ \cline{2-7} 
                                  & 3D Face Recognition Attacks \cite{Adv3Dface}                                  & adversarial illumination & 3D face model                          & No                         & loss of the face recognition model                                                                                                                                                & Penalty on perturbations in sensitive areas                                                                                                                    \\ \hline
Person Re-ID                      & AdvPatterns: Wang et al. \cite{AdvPersonReID}                            & Patch                    & No                                     & Yes                        & minimize the similarity between two images of the same subject from two different cameras and maximize the similarity between two images of the same subject from the same camera & total variation                                                                                                                                                \\ \hline
\multirow{2}{*}{Human Tracking}   & PAT - Adversarial Posters: Wiyatno et al. \cite{AdvTracking}           & Poster                   & No                                     & Yes                        & non-targeted loss (victim model’s confusion) + targeted loss (distance to an adversarial target output)                                                                           & Lagrangian-relaxed loss (perceptual similarity)                                                                                                                \\ \cline{2-7} 
                                  & Universal Physical Attacks on Single Object Tracking \cite{ding2021towards} & Patch                    & No                                     & Yes                        & Maximum Textural Discrepancy (MTD) loss + shape attack loss (i.e difference between the shape of the selected top-K bounding boxes and the shape of the targeted bounding boxes)  & total variation                                                                                                                                                \\ \hline
Action Recognition                & Over-the-Air Adversarial Flickering Attacks \cite{PhysicalAdvAction}          & led bulb                 & No                                     & Yes                        & loss of the action classifier                                                                                                                                                     & total distortion introduced by the adversarial perturbations                                                                                                   \\ \hline
\end{tabular}}
\label{tab:summary_attack}
\end{table*}

\vspace{6px}
\noindent\textbf{Multispectral Adversarial Patch: } In recent studies, multispectral person detection has shown great promise especially, in scenarios such as illumination variations and occlusions. In \cite{kim2022map} the authors propose a Multispectral Adversarial Patch (MAP) generation framework to evaluate the robustness of multispectral person detectors in the physical world, which is the first work towards this direction.  The authors use the Cross-spectral Mapping (CSM) technique to translate a patch from a particular sensor to the other modality. For instance, a patch from thermal modality, $p_{thermal}$ is mapped across to RGB, $p_{rgb}$ and vise versa. To perform this the authors have predefined three materials (Aluminum, Steel, and Sand paper) and investigated the intensity and RGB values of those materials. Using the guidance from these pre-defined materials a transformation function is defined to transfer pixel values from one modality to another.  


The overall loss function that needs to be optimized for MAP generation is defined as,

\begin{equation}
    \mathcal{L} = \mathcal{L}_{thermal} +\mathcal{L}_{rgb} + \mathcal{L}_{obj},
\end{equation}
where $\mathcal{L}_{obj}$ is the loss function that minimize the objectness scores of the Faster-RCNN network. Therefore, we can denote $\mathcal{L}_T = \mathcal{L}_{obj}$ and $\mathcal{L}_P = \mathcal{L}_{thermal} +\mathcal{L}_{rgb}$.

The loss function $L_{thermal}$ is derived as,
\begin{equation}
 \mathcal{L}_{thermal} = \sum_{p^{i, j} \in p_{thermal}} \min_{m \in M} |p^{i, j} - m| + \mathcal{L}_{TV},
\end{equation}
where $p^{i, j}$ denotes the (i, j) pixel of $p_{thermal}$ and $M$ represents the intensity levels of the materials. While minimizing this loss the authors encourage the generated thermal patch to have intensity values so that it is possible to manufacture using the predefined materials. A similar loss function is defined for the RGB case (i.e. $\mathcal{L}_{rgb}$). Using $\mathcal{L}_{rgb}$ the authors ensure that the RGB patch contains pixel values that can be printed on paper. A visual explanation of this Cross-spectral Mapping is given in Fig.~\ref{fig:Advmultispectral2}.

The authors have collected a multispectral person detection dataset with 1,500 rgb-thermal image pairs. This dataset captures different illumination conditions as well as different backgrounds. Using this dataset the authors have demonstrated the possibility of generating physical and digital attacks toward multispectral person detection. 


\subsection{Summary of Different Adversarial Attack Methods}

In Tab. \ref{tab:summary_attack} we provide a summary of the analysis that we conducted using the proposed framework. We observe similarities across different attack methods in terms of utilised task and physical losses as well as the utilised accessories.

\section{Defending physical adversarial attacks}\label{sec:Defence}
In this section we discuss the methods introduced for defending physical adversarial attacks on human surveillance. We discuss the methods introduced for defending human detection attacks, human recognition attacks and human action recognition attacks. It should be noted that, to the best of our knowledge, there is no method proposed for defending human tracking attacks.

\subsection{Defending Surveillance Human Detection Attacks}

\noindent\textbf{Universal Defensive Frame \cite{yu2022defending}:} Yu \emph{et al.} proposed an iteratively competing optimization process where the adversarial patch and the defensive frame are competing against each other. Specifically, the authors first create sub-image sets consisting of $M$ images sampled from the data distribution of the original image. Then $M$ adversarial images are created using these images and defensive pattern is then optimized to reduce the difference between the $M$ adversarial images and the $M$ clean images. As such, the defensive frame is universal for different adversarial attack types and image-agnostic for images sampled from the input data distribution. However, the evaluations are limited to defending digital attacks.

\noindent\textbf{Adversarial YOLO (Ad-YOLO) \cite{ji2021adversarial}:} In a different line of work, a plug-in defense component on the YOLO detection system is proposed in \cite{ji2021adversarial}. Specifically, the authors have added a new category `adversarial patch' to YOLOv2 such that it can directly detect adversarial objects in the input. The proposed Ad-YOLO method has all the layers of YOLOv2 and the authors have only modified the last layer by adding a `patch category' output which recognises the objects and patches from the input image at the same time. For training the Ad-YOLO method the authors have created an augmented Pascal VOC dataset where bonafide and adversarial patch images are inserted to Pascal VOC dataset. This method has been evaluated for defending physical adverarila attacks.  

\subsection{Defending Surveillance Human Identification Attacks}

A few defence mechanisms have been proposed for defending face recognition systems from adversarial attacks. 

\noindent\textbf{Adversarial Patch Detector \cite{xie2023random}:} The method in \cite{xie2023random} operated by inspecting the input images before feeding them to the face recognition system. The authors have first split the input image into multiple patches ( which can be either even or random patches) and then a detector is trained to detect suspicious which are subsequently filtered before feeding the face image to the face recognition system. This system is trained on bonafide face images and the adversarial images generated using white-box attacks and has only been tested on digital white-box defense-model-leaked attacks.

\noindent\textbf{Adversarial Image Purifiers \cite{theagarajan2020defending} :} Theagarajan \emph{et al.} proposes to iteratively purify the adversarial images using an ensemble of purifiers. Specifically, a Bayesian CNN is designed to quantify whether the input image is adversarial and if it is detected as adversarial the input is passed through an ensemble of purifiers which generates the purified image. This purified image is passed back to the Bayesian CNN and if it is detected as not adversarial it is passed as input to the facial recognition model. If the input is still adversarial it is passed back to the purifiers for purification an this process is iteratively applied for either pre-defined number of iterations or until the input is detected as bonafide. This system has only being validated for defending digital adversarial attacks on facial recognition. 

In \cite{wang2022defensive}, a framework for generating defensive patches is introduced. The authors suggest injecting class-specific identifiable patterns into these patches to enhance their effectiveness. Additionally, they focus on ensuring the transferability of this defense mechanism across multiple models by encouraging the defensive patches to capture more global feature correlations within a class during the patch optimization process. The proposed method is successfully validated in defending object recognition systems against both digital and physical attacks. However, it's important to note that this framework has not been specifically tested for defending against attacks on human surveillance systems.

\noindent\textbf{Adversarial Occlusions \cite{wu2019defending}:} Wu \emph{et al.} have first proposed occlusion based adversarial attack method where the attacker introduces a fixed-dimension rectangle which can be placed arbitrarily in the input image and can have adversarial perturbations. Then the attack method iteratively search region to place the rectangle in the input image and generates adversarial perturbations such that the input image becomes adversarial. Once the image is adversarial this image is used for standard adversarial training. This method has been validated to defend a variety of attacks, including, eyeglass attacks, adversarial patch attacks and stop sign attacks. This system has only being validated for defending digital adversarial attacks on facial recognition.

Two methods have been proposed for defending person Re-ID Attacks.

\noindent\textbf{Multi-Expert Adversarial Attack Detection (MEAAD) \cite{wang2021multi}}: This method defends person Re-ID systems by detecting context inconsistencies. Specifically, multiple person Re-ID networks with different structures are used as expert models and a support set is defined as the top-K retrievals output by a single expert model. Then the MEAAD computes context as i) relationships between the query and its support samples returned by a single expert, ii) relationships among the support samples retrieved by a single expert and iii) relationships between the support samples returned by one expert and those returned by another expert. Using these relations as a feature representation a detector is trained to detect the context that belong to an adversarial sample. This system has only being validated for defending digital adversarial attacks on person Re-ID.  

\noindent\textbf{Adversarial Metric Defence \cite{bai2020adversarial}:} Bai \emph{et al.} propose to attack the distance metric used by the person Re-ID and corrupts the pairwise distance between images. The authors have validated their methods using both cross entropy loss and triplet loss with a variety of distance metrics, including, Euclidean distance and Mahalanobis distance. Once the metric attack is defined, an adversarial version of training set is generated by augmenting each example in the training set using the adversarial perturbations generated by the proposed metric attack method. Then using adversarial training the person Re-ID framework is re-trained to defend the metric attacks. 

\vspace{-9px}
\subsection{Defending Surveillance Action Recognition Attacks}
\noindent\textbf{Inpainting with Laplacian Prior \cite{anand2020adversarial}:} The authors have detected the  high-frequency adversarial noise patches in the image gradient domain and the patch is localised using a thresholding operation. Then instead of suppressing the adversarial noise, the authors propose to innpaint the region. However, this sytem has been validated only for optical flow-based action recognition frameworks and for defending only digital adversarial attacks.



\section{Discussion and Conclusions}
\label{sec:Conclusions}

Deep learning algorithms' great success is marred by their vulnerability to adversarial examples, which poses real threats in surveillance scenarios, both digitally and physically. Physical adversarial attacks have proven effective against all surveillance models, irrespective of tasks (detection, identification, tracking, and action recognition), data type (images or videos), or imaging spectrum (visible and infrared).

Compared with digital adversarial attacks, attacks in the physical domain of surveillance have been less explored as the high dimensions of varying factors of real-world conditions and they require manipulation of the actual subjects, objects, or scenes. In the literature, only adversarial attacks for human detection have been thoroughly investigated, research on other tasks are emerging. Based on our review and analysis, we identify the following challenges and milestones for the community to tackle.

\emph{Video-based tasks} such as gait biometrics and human action recognition are still largely unexplored, but they are critical tasks in surveillance. There is very few work on the video-based inputs. The video base method such as gait biometrics models and action recognition models leverage both spatial information as well as the temporal progression of the spatial features in their recognition pipeline. As such, the adversarial attack methods can leverage both of these domains to thwart the surveillance. However, to the best of our knowledge, there is no system that investigates utilizing temporal domain for generating adversarial attacks, as such, generating adversarial behaviors is area for future investigation. 

\emph{Beyond visible} footage from spectrum other than visible such as infrared, hyperspectral, radar is important in surveillance; however, the impact of adversarial attacks on these spectrum is mostly unexplored. An investigation could be conducted to evaluate the effectiveness of the adversarial attack methods proposed for visible spectrum under different surveillance settings that are beyond visible. Furthermore, pipelines such as universal adversarial patch generation pipelines can be leveraged to generate adversarial patches that are effective under different different sending modalities. Such methods can be beneficial when attacking multispectral and hyperspectral surveillance systems considering the fact that their observations span across multiple modalities.   

\emph{Multimodal surveillance} many surveillance systems employ a mixture of modalities to improve performance of surveillance. However, adversarial attack approaches that operate with these multimodal systems are very limited. The multimodality input generates a unique perspective and challenges with respect to the adversarial attack as the generated perturbations should be visible and effective under all the perceived modalities. For instance, when attacking visible + LiDAR-based multimodal surveillance framework the attack method should consider a diverse set of feature spaces for the attack generation. As such, direct extension of the unimodal adversarial attacks to multimodal setting would be less effective and further investigation is required in order to generate robust and effective multimodal physical adversarial examples. 

\emph{Defending physical adversarial attacks} is another area which is less explored. For instance, there is on method proposed for defending surveillance human tracking attacks. Furthermore, most of the adversarial defence frameworks proposed are being evaluated only under digital attack settings and only a limited number of works have been tested using physical adversarial examples. Therefore, the validity of these defence mechanisms under different capture conditions, illumination conditions, diverse attack types should be validated in order to verify their effectiveness.


%



\section*{Acknowledgment}
This research was supported by Discovery Project Grant DP200101942 awarded by the Australian Research Council (ARC). The authors would like to thank editors and reviewers for their invaluable comments.

\ifCLASSOPTIONcaptionsoff
  \newpage
\fi

\bibliographystyle{ieee}
\bibliography{PAAS}

\begin{thebibliography}{100}\itemsep=-1pt

\bibitem{AdversarialFashion}
Adversarial fashion.
\newblock \url{https://adversarialfashion.com/}, 2020.

\bibitem{CVDazzle}
Computer vision dazzle camouflage.
\newblock \url{https://cvdazzle.com}, 2020.

\bibitem{FaceJewellery}
Igconito face jewellery to avoid facial recognition.
\newblock
  \url{https://www.dezeen.com/2019/07/30/ewa-nowak-anti-ai-mask-protects-wearers-from-mass-surveillance/},
  2020.

\bibitem{LEDglasses}
Led light privacy visor.
\newblock \url{https://www.nii.ac.jp/userimg/press_details_20121212.pdf/},
  2020.

\bibitem{FaceProjector}
A projector can be used to project faces and images onto your own face to
  disguise surveillance face recognition software.
\newblock \url{https://www.instagram.com/p/BSlVHpIBxp7/?utm_source=ig_embed/},
  2020.

\bibitem{Reflectacleglasses}
Reflectacles privacy eyewear.
\newblock \url{https://www.reflectacles.com/}, 2020.

\bibitem{Scarf}
This scarf scrambles fascists.
\newblock
  \url{https://www.inverse.com/article/30589-facial-recognition-scrambling-scarf-surveillance-earrings/},
  2020.

\bibitem{3DFaceMask}
Urme personal surveillance identity prosthetic.
\newblock \url{http://www.urmesurveillance.com/urme-prosthetic/}, 2020.

\bibitem{abdelfattah2021adversarial}
M.~Abdelfattah, K.~Yuan, Z.~J. Wang, and R.~Ward.
\newblock Adversarial attacks on camera-lidar models for 3d car detection.
\newblock In {\em 2021 IEEE/RSJ International Conference on Intelligent Robots
  and Systems (IROS)}, pages 2189--2194. IEEE, 2021.

\bibitem{AdvSurvey}
N.~{Akhtar} and A.~{Mian}.
\newblock Threat of adversarial attacks on deep learning in computer vision: A
  survey.
\newblock {\em IEEE Access}, 6:14410--14430, 2018.

\bibitem{akhtar2018threat}
N.~Akhtar and A.~Mian.
\newblock Threat of adversarial attacks on deep learning in computer vision: A
  survey.
\newblock {\em IEEE Access}, 6:14410--14430, 2018.

\bibitem{YOLOv4}
B.~Alexey, W.~Chien-Yao, and L.~Hong-Yuan.
\newblock Yolov4: Optimal speed and accuracy of object detection.
\newblock {\em CoRR}, abs/2004.10934, 2020.

\bibitem{anand2020adversarial}
A.~P. Anand, H.~Gokul, H.~Srinivasan, P.~Vijay, and V.~Vijayaraghavan.
\newblock Adversarial patch defense for optical flow networks in video action
  recognition.
\newblock In {\em IEEE International Conference on Machine Learning and
  Applications (ICMLA)}, pages 1289--1296. IEEE, 2020.

\bibitem{bai2020adversarial}
S.~Bai, Y.~Li, Y.~Zhou, Q.~Li, and P.~H. Torr.
\newblock Adversarial metric attack and defense for person re-identification.
\newblock {\em IEEE Transactions on Pattern Analysis and Machine Intelligence},
  43(6):2119--2126, 2020.

\bibitem{DigitalAdvPersonReID}
S.~Bai, Y.~Li, Y.~Zhou, Q.~Li, and P.~H.~S. Torr.
\newblock Adversarial metric attack for person re-identification.
\newblock {\em IEEE Transactions on Pattern Analysis and Machine Intelligence},
  pages 1--8, 2020.

\bibitem{DigitalAdvPersonReID3}
Q.~Bouniot, R.~Audigier, and A.~Loesch.
\newblock Vulnerability of person re-identification models to metric
  adversarial attacks.
\newblock In {\em IEEE/CVF Conference on Computer Vision and Pattern
  Recognition Workshops (CVPRW)}, pages 3450--3459, 2020.

\bibitem{CircularAdvPatches}
T.~B. Brown, D.~Man{\'{e}}, A.~Roy, M.~Abadi, and J.~Gilmer.
\newblock Adversarial patch.
\newblock {\em CoRR}, abs/1712.09665, 2017.

\bibitem{TrackingSurvey}
A.~Brunetti, D.~Buongiorno, G.~F. Trotta, and V.~Bevilacqua.
\newblock Computer vision and deep learning techniques for pedestrian detection
  and tracking: A survey.
\newblock {\em Neurocomputing}, 300:17 -- 33, 2018.

\bibitem{CascadeRCNN}
Z.~Cai and N.~Vasconcelos.
\newblock Cascade r-cnn: Delving into high quality object detection.
\newblock In {\em CVPR}, 2018.

\bibitem{AdvLiDAR}
Y.~Cao, C.~Xiao, D.~Yang, J.~Fang, R.~Yang, M.~Liu, and B.~Li.
\newblock Adversarial objects against lidar-based autonomous driving systems.
\newblock {\em CoRR}, abs/1907.05418, 2019.

\bibitem{CandW}
N.~Carlini and D.~Wagner.
\newblock Towards evaluating the robustness of neural networks.
\newblock In {\em IEEE Symposium on Security and Privacy (SP)}, pages 39--57,
  2017.

\bibitem{I3D}
J.~{Carreira} and A.~{Zisserman}.
\newblock Quo vadis, action recognition? a new model and the kinetics dataset.
\newblock In {\em IEEE/CVF Conference on Computer Vision and Pattern
  Recognition (CVPR)}, pages 4724--4733, 2017.

\bibitem{SiamMaskE}
B.~X. {Chen} and J.~{Tsotsos}.
\newblock Fast visual object tracking using ellipse fitting for rotated
  bounding boxes.
\newblock In {\em IEEE/CVF International Conference on Computer Vision Workshop
  (ICCVW)}, pages 2281--2289, 2019.

\bibitem{ShapeShifter}
S.-T. Chen, C.~Cornelius, and {et al}.
\newblock Shapeshifter: Robust physical adversarial attack on faster r-cnn
  object detector.
\newblock In {\em Machine Learning and Knowledge Discovery in Databases}, pages
  52--68, 2019.

\bibitem{church2018aerial}
P.~Church, C.~Grebe, J.~Matheson, and B.~Owens.
\newblock Aerial and surface security applications using lidar.
\newblock {\em Laser Radar Technology and Applications XXIII}, 10636:27--38,
  2018.

\bibitem{AdvTransferability2}
A.~Demontis, M.~Melis, M.~Pintor, M.~Jagielski, B.~Biggio, A.~Oprea,
  C.~Nita-Rotaru, and F.~Roli.
\newblock Why do adversarial attacks transfer? explaining transferability of
  evasion and poisoning attacks.
\newblock In {\em {USENIX} Security Symposium}, pages 321--338, 2019.

\bibitem{ArcFace}
J.~{Deng}, J.~{Guo}, N.~{Xue}, and S.~{Zafeiriou}.
\newblock Arcface: Additive angular margin loss for deep face recognition.
\newblock In {\em IEEE/CVF Conference on Computer Vision and Pattern
  Recognition (CVPR)}, pages 4685--4694, 2019.

\bibitem{deng2019accurate}
Y.~Deng, J.~Yang, S.~Xu, D.~Chen, Y.~Jia, and X.~Tong.
\newblock Accurate 3d face reconstruction with weakly-supervised learning: From
  single image to image set.
\newblock In {\em Proceedings of the IEEE/CVF conference on computer vision and
  pattern recognition workshops}, pages 0--0, 2019.

\bibitem{ding2021towards}
L.~Ding, Y.~Wang, K.~Yuan, M.~Jiang, P.~Wang, H.~Huang, and Z.~J. Wang.
\newblock Towards universal physical attacks on single object tracking.
\newblock In {\em AAAI Conference on Artificial Intelligence (AAAI)}, pages
  1236--1245, 2021.

\bibitem{QUTAdvLED}
M.~Frearson and K.~Nguyen.
\newblock Adversarial attack on facial recognition using visible light.
\newblock {\em CoRR}, abs/2011.12680, 2020.

\bibitem{FastRCNN}
R.~Girshick.
\newblock {Fast R-CNN [Region-based Convolutional Neural Network]}.
\newblock In {\em IEEE International Conference on Computer Vision (ICCV)},
  pages 1440 -- 8, 2015.

\bibitem{FGSM}
I.~J. Goodfellow, J.~Shlens, and C.~Szegedy.
\newblock Explaining and harnessing adversarial examples.
\newblock In {\em International Conference on Learning Representations (ICLR)},
  2015.

\bibitem{guo2021meaningful}
Y.~Guo, X.~Wei, G.~Wang, and B.~Zhang.
\newblock Meaningful adversarial stickers for face recognition in physical
  world.
\newblock {\em arXiv preprint arXiv:2104.06728}, 2021.

\bibitem{MaskRCNN}
K.~He, G.~Gkioxari, P.~Dollar, and R.~Girshick.
\newblock Mask {R-CNN}.
\newblock In {\em IEEE International Conference on Computer Vision (ICCV)},
  pages 2980 -- 8, 2017.

\bibitem{GOTURN}
D.~Held, S.~Thrun, and S.~Savarese.
\newblock Learning to track at 100 fps with deep regression networks.
\newblock In {\em European Conference on Computer Vision (ECCV)}, pages
  749--765, 2016.

\bibitem{ActionRecognitionSurvey}
S.~Herath, M.~Harandi, and F.~Porikli.
\newblock Going deeper into action recognition: A survey.
\newblock {\em Image and Vision Computing}, 60:4 -- 21, 2017.
\newblock Regularization Techniques for High-Dimensional Data Analysis.

\bibitem{hu2021naturalistic}
Y.-C.-T. Hu, B.-H. Kung, D.~S. Tan, J.-C. Chen, K.-L. Hua, and W.-H. Cheng.
\newblock Naturalistic physical adversarial patch for object detectors.
\newblock In {\em IEEE/CVF International Conference on Computer Vision (CVPR)},
  pages 7848--7857, 2021.

\bibitem{hu2023physically}
Z.~Hu, W.~Chu, X.~Zhu, H.~Zhang, B.~Zhang, and X.~Hu.
\newblock Physically realizable natural-looking clothing textures evade person
  detectors via 3d modeling.
\newblock In {\em Proceedings of the IEEE/CVF Conference on Computer Vision and
  Pattern Recognition}, pages 16975--16984, 2023.

\bibitem{AdvTexture}
Z.~Hu, S.~Huang, X.~Zhu, F.~Sun, B.~Zhang, and X.~Hu.
\newblock Adversarial texture for fooling person detectors in the physical
  world.
\newblock In {\em IEEE/CVF Conference on Computer Vision and Pattern
  Recognition (CVPR)}, pages 13307--13316, June 2022.

\bibitem{LFW}
G.~B. Huang, N.~Ramesh, T.~Berg, and E.~Learned-Miller.
\newblock Labeled faces in the wild: A database for studying face recognition
  in unconstrained environments.
\newblock Technical report, University of Massachusetts, 2007.

\bibitem{UPC}
L.~{Huang}, C.~{Gao}, and {et al}.
\newblock Universal physical camouflage attacks on object detectors.
\newblock In {\em IEEE/CVF Conference on Computer Vision and Pattern
  Recognition (CVPR)}, pages 717--726, 2020.

\bibitem{AdvExistence2}
A.~Ilyas, S.~Santurkar, D.~Tsipras, L.~Engstrom, B.~Tran, and A.~Madry.
\newblock Adversarial examples are not bugs, they are features.
\newblock In {\em Advances in Neural Information Processing Systems (NeurIPS)},
  volume~32, pages 125--136, 2019.

\bibitem{STN}
M.~Jaderberg, K.~Simonyan, A.~Zisserman, and k.~kavukcuoglu.
\newblock Spatial transformer networks.
\newblock In {\em Advances in Neural Information Processing Systems (NeurIPS)},
  volume~28, pages 2017--2025, 2015.

\bibitem{ji2021adversarial}
N.~Ji, Y.~Feng, H.~Xie, X.~Xiang, and N.~Liu.
\newblock Adversarial yolo: Defense human detection patch attacks via detecting
  adversarial patches.
\newblock {\em arXiv preprint arXiv:2103.08860}, 2021.

\bibitem{AdvMOT}
Y.~Jia, Y.~Lu, J.~Shen, Q.~A. Chen, H.~Chen, Z.~Zhong, and T.~Wei.
\newblock Fooling detection alone is not enough: Adversarial attack against
  multiple object tracking.
\newblock In {\em International Conference on Learning Representations (ICLR)},
  2020.

\bibitem{R21DBERT}
M.~E. Kalfaoglu, S.~Kalkan, and A.~A. Alatan.
\newblock Late temporal modeling in 3d cnn architectures with bert for action
  recognition.
\newblock In {\em European Conference on Computer Vision (ECCV)}, pages 21 --
  37, 2020.

\bibitem{MegaFace}
I.~Kemelmacher-Shlizerman, S.~M. Seitz, D.~Miller, and E.~Brossard.
\newblock The megaface benchmark: 1 million faces for recognition at scale.
\newblock In {\em IEEE/CVF Conference on Computer Vision and Pattern
  Recognition (CVPR)}, pages 4873--4882, 2016.

\bibitem{kim2022map}
T.~Kim, H.~J. Lee, and Y.~M. Ro.
\newblock Map: Multispectral adversarial patch to attack person detection.
\newblock In {\em IEEE International Conference on Acoustics, Speech and Signal
  Processing (ICASSP)}, pages 4853--4857. IEEE, 2022.

\bibitem{AdvHat}
S.~Komkov and A.~Petiushko.
\newblock Advhat: Real-world adversarial attack on arcface face {ID} system.
\newblock {\em CoRR}, abs/1908.08705, 2019.

\bibitem{CellphoneAttacks}
A.~Kurakin, I.~J. Goodfellow, and S.~Bengio.
\newblock Adversarial examples in the physical world.
\newblock In {\em International Conference on Learning Representations (ICLR)},
  2019.

\bibitem{PointPillars}
A.~H. {Lang}, S.~{Vora}, H.~{Caesar}, L.~{Zhou}, J.~{Yang}, and O.~{Beijbom}.
\newblock Pointpillars: Fast encoders for object detection from point clouds.
\newblock In {\em IEEE/CVF Conference on Computer Vision and Pattern
  Recognition (CVPR)}, pages 12689--12697, 2019.

\bibitem{DPatch2}
M.~Lee and J.~Z. Kolter.
\newblock On physical adversarial patches for object detection.
\newblock {\em CoRR}, abs/1906.11897, 2019.

\bibitem{SiamRPNpp}
B.~{Li}, W.~{Wu}, Q.~{Wang}, F.~{Zhang}, J.~{Xing}, and J.~{Yan}.
\newblock Siamrpn++: Evolution of siamese visual tracking with very deep
  networks.
\newblock In {\em IEEE/CVF Conference on Computer Vision and Pattern
  Recognition (CVPR)}, pages 4277--4286, 2019.

\bibitem{Adv3Dface}
Y.~Li, Y.~Li, X.~Dai, S.~Guo, and B.~Xiao.
\newblock Physical-world optical adversarial attacks on 3d face recognition.
\newblock In {\em IEEE/CVF Conference on Computer Vision and Pattern
  Recognition (CVPR)}, 2023.

\bibitem{MSCOCO}
T.-Y. Lin, M.~Maire, S.~Belongie, J.~Hays, P.~Perona, D.~Ramanan,
  P.~Doll{\'a}r, and C.~L. Zitnick.
\newblock Microsoft coco: Common objects in context.
\newblock In {\em European Conference on Computer Vision (ECCV)}, pages
  740--755, 2014.

\bibitem{AdaptiveFace}
H.~{Liu}, X.~{Zhu}, Z.~{Lei}, and S.~Z. {Li}.
\newblock Adaptiveface: Adaptive margin and sampling for face recognition.
\newblock In {\em IEEE/CVF International Conference on Computer Vision and
  Pattern Recognition (CVPR)}, pages 11939--11948, 2019.

\bibitem{DPatch}
X.~Liu, H.~Yang, Z.~Liu, L.~Song, H.~Li, and Y.~Chen.
\newblock Dpatch: An adversarial patch attack on object detectors.
\newblock In {\em AAAI Workshop on Artificial Intelligence Safety}, volume
  2301, 2019.

\bibitem{AdvTransferability}
Y.~Liu, X.~Chen, C.~Liu, and D.~Song.
\newblock Delving into transferable adversarial examples and black-box attacks.
\newblock In {\em International Conference on Learning Representations (ICLR)},
  2017.

\bibitem{AdvFaceMask}
B.~MacDonald.
\newblock Fooling facial detection with fashion.
\newblock
  \url{https://towardsdatascience.com/fooling-facial-detection-with-fashion-d668ed919eb},
  2019.

\bibitem{Deepfool}
S.-M. Moosavi-Dezfooli, A.~Fawzi, and P.~Frossard.
\newblock Deepfool: A simple and accurate method to fool deep neural networks.
\newblock In {\em IEEE/CVF Conference on Computer Vision and Pattern
  Recognition (CVPR)}, pages 2574 -- 82, 2016.

\bibitem{CamouflageHistory}
T.~Newark.
\newblock {\em Camouflage / introduction by Jonathan Miller.}
\newblock Thames and Hudson, London, 2007.

\bibitem{JSMA}
N.~Papernot, P.~McDaniel, S.~Jha, M.~Fredrikson, Z.~Celik, and A.~Swami.
\newblock The limitations of deep learning in adversarial settings.
\newblock In {\em IEEE European Symposium on Security and Privacy (EuroSPP)},
  pages 372 -- 87, 2016.

\bibitem{AdvFacePatches}
M.~Pautov, G.~Melnikov, E.~Kaziakhmedov, K.~Kireev, and A.~Petiushko.
\newblock On adversarial patches: real-world attack on arcface-100 face
  recognition system.
\newblock {\em CoRR}, abs/1910.07067, 2019.

\bibitem{PhysicalAdvAction}
R.~Pony, I.~Naeh, and S.~Mannor.
\newblock Over-the-air adversarial flickering attacks against video recognition
  networks.
\newblock In {\em IEEE/CVF Conference on Computer Vision and Pattern
  Recognition (CVPR)}, 2021.

\bibitem{YOLOv2}
J.~Redmon and A.~Farhadi.
\newblock {YOLO9000: better, faster, stronger}.
\newblock In {\em IEEE/CVF Conference on Computer Vision and Pattern
  Recognition (CVPR)}, pages 6517 -- 25, 2017.

\bibitem{YOLOv3}
J.~Redmon and A.~Farhadi.
\newblock Yolov3: An incremental improvement.
\newblock {\em CoRR}, abs/1804.02767, 2018.

\bibitem{FasterRCNN}
S.~Ren, K.~He, R.~Girshick, and J.~Sun.
\newblock {Faster R-CNN: Towards real-time object detection with region
  proposal networks}.
\newblock In {\em Advances in Neural Information Processing Systems (NeurIPS)},
  pages 91 -- 99, 2015.

\bibitem{FaceNet}
F.~Schroff, D.~Kalenichenko, and J.~Philbin.
\newblock Facenet: A unified embedding for face recognition and clustering.
\newblock In {\em IEEE/CVF International Conference on Computer Vision and
  Pattern Recognition (CVPR)}, pages 815--823, 2015.

\bibitem{AdvExistence}
A.~Shafahi, W.~R. Huang, C.~Studer, S.~Feizi, and T.~Goldstein.
\newblock Are adversarial examples inevitable?
\newblock In {\em International Conference on Learning Representations (ICLR)},
  2019.

\bibitem{advGlasses}
M.~Sharif, S.~Bhagavatula, L.~Bauer, and M.~K. Reiter.
\newblock A general framework for adversarial examples with objectives.
\newblock {\em ACM Transactions on Privacy and Security}, 2019.

\bibitem{PointRCNN}
S.~{Shi}, X.~{Wang}, and H.~{Li}.
\newblock Pointrcnn: 3d object proposal generation and detection from point
  cloud.
\newblock In {\em IEEE/CVF Conference on Computer Vision and Pattern
  Recognition (CVPR)}, pages 770--779, 2019.

\bibitem{UCF101}
K.~Soomro, A.~R. Zamir, and M.~Shah.
\newblock {UCF101:} {A} dataset of 101 human actions classes from videos in the
  wild.
\newblock {\em CoRR}, abs/1212.0402, 2012.

\bibitem{OnePixelAttack}
J.~Su, D.~Vargas, and K.~Sakurai.
\newblock One pixel attack for fooling deep neural networks.
\newblock {\em IEEE Transactions on Evolutionary Computation}, 23(5):828 -- 41,
  2019.

\bibitem{AdvOriginal}
C.~Szegedy, W.~Zaremba, I.~Sutskever, J.~Bruna, D.~Erhan, I.~Goodfellow, and
  R.~Fergus.
\newblock Intriguing properties of neural networks.
\newblock In {\em International Conference on Learning Representation (ICLR)},
  2014.

\bibitem{LBFGS}
C.~Szegedy, W.~Zaremba, I.~Sutskever, J.~Bruna, D.~Erhan, I.~Goodfellow, and
  R.~Fergus.
\newblock Intriguing properties of neural networks.
\newblock In {\em International Conference on Learning Representations (ICLR)},
  2014.

\bibitem{tang2019augmentation}
Z.~Tang, K.~Chen, M.~Pan, M.~Wang, and Z.~Song.
\newblock An augmentation strategy for medical image processing based on
  statistical shape model and 3d thin plate spline for deep learning.
\newblock {\em IEEE Access}, 7:133111--133121, 2019.

\bibitem{theagarajan2020defending}
R.~Theagarajan and B.~Bhanu.
\newblock Defending black box facial recognition classifiers against
  adversarial attacks.
\newblock In {\em IEEE/CVF conference on computer vision and pattern
  recognition workshops (CVPRW)}, pages 812--813, 2020.

\bibitem{AdvPatches}
S.~Thys, W.~V. Ranst, and T.~Goedem{\'e}.
\newblock Fooling automated surveillance cameras: adversarial patches to attack
  person detection.
\newblock In {\em IEEE/CVF Conference on Computer Vision and Pattern
  Recognition Workshops (CVPRW)}, 2019.

\bibitem{tu2021exploring}
J.~Tu, H.~Li, X.~Yan, M.~Ren, Y.~Chen, M.~Liang, E.~Bitar, E.~Yumer, and
  R.~Urtasun.
\newblock Exploring adversarial robustness of multi-sensor perception systems
  in self driving.
\newblock {\em Conference on Robot Learning}, 2021.

\bibitem{AdvLiDAR2}
J.~Tu, M.~Ren, S.~Manivasagam, M.~Liang, B.~Yang, R.~Du, F.~Cheng, and
  R.~Urtasun.
\newblock Physically realizable adversarial examples for lidar object
  detection.
\newblock In {\em IEEE/CVF Conference on Computer Vision and Pattern
  Recognition (CVPR)}, pages 13713 -- 22, 2020.

\bibitem{ScaledYOLO}
C.-Y. Wang, A.~Bochkovskiy, and H.-Y.~M. Liao.
\newblock Scaled-yolov4: Scaling cross stage partial network, 2021.

\bibitem{stReID}
G.~Wang, J.~Lai, P.~Huang, and X.~Xie.
\newblock Spatial-temporal person re-identification.
\newblock pages 8933 -- 8940, 2019.

\bibitem{TCIAdvPersonReID}
H.~Wang, G.~Wang, Y.~Li, D.~Zhang, and L.~Lin.
\newblock Transferable, controllable, and inconspicuous adversarial attacks on
  person re-identification with deep mis-ranking.
\newblock In {\em IEEE/CVF Conference on Computer Vision and Pattern
  Recognition (CVPR)}, 2020.

\bibitem{CosFace}
H.~Wang, Y.~Wang, Z.~Zhou, X.~Ji, D.~Gong, J.~Zhou, Z.~Li, and W.~Liu.
\newblock Cosface: Large margin cosine loss for deep face recognition.
\newblock In {\em IEEE/CVF International Conference on Computer Vision and
  Pattern Recognition (CVPR)}, pages 5265--5274, 2018.

\bibitem{wang2022defensive}
J.~Wang, Z.~Yin, P.~Hu, A.~Liu, R.~Tao, H.~Qin, X.~Liu, and D.~Tao.
\newblock Defensive patches for robust recognition in the physical world.
\newblock In {\em Proceedings of the IEEE/CVF Conference on Computer Vision and
  Pattern Recognition}, pages 2456--2465, 2022.

\bibitem{DFR_survey}
M.~Wang and W.~Deng.
\newblock Deep face recognition: {A} survey.
\newblock {\em CoRR}, abs/1804.06655, 2018.

\bibitem{wang2021multi}
X.~Wang, S.~Li, M.~Liu, Y.~Wang, and A.~K. Roy-Chowdhury.
\newblock Multi-expert adversarial attack detection in person re-identification
  using context inconsistency.
\newblock In {\em IEEE/CVF International Conference on Computer Vision (ICCV)},
  pages 15097--15107, 2021.

\bibitem{SVXSoftmax}
X.~Wang, S.~Zhang, S.~Wang, T.~Fu, H.~Shi, and T.~Mei.
\newblock Mis-classified vector guided softmax loss for face recognition.
\newblock {\em AAAI Conference on Artificial Intelligence (AAAI)},
  34(07):12241--12248, 2020.

\bibitem{wang2021towards}
Y.~Wang, H.~Lv, X.~Kuang, G.~Zhao, Y.-a. Tan, Q.~Zhang, and J.~Hu.
\newblock Towards a physical-world adversarial patch for blinding object
  detection models.
\newblock {\em Information Sciences}, 556:459--471, 2021.

\bibitem{AdvPersonReID}
Z.~Wang, S.~Zheng, M.~Song, Q.~Wang, A.~Rahimpour, and H.~Qi.
\newblock advpattern: Physical-world attacks on deep person re-identification
  via adversarially transformable patterns.
\newblock pages 8340 -- 9, 2019.

\bibitem{wei2022physically}
X.~Wei, B.~Pu, J.~Lu, and B.~Wu.
\newblock Physically adversarial attacks and defenses in computer vision: A
  survey.
\newblock {\em arXiv preprint arXiv:2211.01671}, 2022.

\bibitem{wei2023physically}
X.~Wei, J.~Yu, and Y.~Huang.
\newblock Physically adversarial infrared patches with learnable shapes and
  locations.
\newblock In {\em Proceedings of the IEEE/CVF Conference on Computer Vision and
  Pattern Recognition}, pages 12334--12342, 2023.

\bibitem{CenterFace}
Y.~Wen, K.~Zhang, Z.~Li, and Y.~Qiao.
\newblock A discriminative feature learning approach for deep face recognition.
\newblock In {\em European Conference on Computer Vision (ECCV)}, pages
  499--515, 2016.

\bibitem{CTL}
M.~Wieczorek, B.~Rychalska, and J.~Dabrowski.
\newblock On the unreasonable effectiveness of centroids in image retrieval.
\newblock {\em CoRR}, abs/2104.13643, 2021.

\bibitem{AdvTracking}
R.~{Wiyatno} and A.~{Xu}.
\newblock Physical adversarial textures that fool visual object tracking.
\newblock In {\em IEEE/CVF International Conference on Computer Vision (ICCV)},
  pages 4821--4830, 2019.

\bibitem{wiyatno2019adversarial}
R.~R. Wiyatno, A.~Xu, O.~Dia, and A.~De~Berker.
\newblock Adversarial examples in modern machine learning: A review.
\newblock {\em arXiv preprint arXiv:1911.05268}, 2019.

\bibitem{wu2019defending}
T.~Wu, L.~Tong, and Y.~Vorobeychik.
\newblock Defending against physically realizable attacks on image
  classification.
\newblock {\em arXiv preprint arXiv:1909.09552}, 2019.

\bibitem{AdvTshirtFB}
Z.~Wu, S.-N. Lim, L.~S. Davis, and T.~Goldstein.
\newblock Making an invisibility cloak: Real world adversarial attacks on
  object detectors.
\newblock In {\em European Conference on Computer Vision (ECCV)}, pages 1--17,
  2020.

\bibitem{xie2023random}
J.~Xie, Y.~Luo, and J.~Lu.
\newblock A random-patch based defense strategy against physical attacks for
  face recognition systems.
\newblock {\em arXiv preprint arXiv:2304.07822}, 2023.

\bibitem{AdvTShirtIBM}
K.~Xu, G.~Zhang, S.~Liu, Q.~Fan, M.~Sun, H.~Chen, P.~Chen, Y.~Wang, and X.~Lin.
\newblock Adversarial t-shirt! evading person detectors in a physical world.
\newblock In {\em European Conference on Computer Vision (ECCV)}, pages 21 --
  37, 2020.

\bibitem{STGCN}
S.~Yan, Y.~Xiong, and D.~Lin.
\newblock Spatial temporal graph convolutional networks for skeleton-based
  action recognition.
\newblock In {\em AAAI Conference on Artificial Intelligence (AAAI)}, pages
  7444 -- 7452, 2018.

\bibitem{HijackingTracker}
X.~Yan, X.~Chen, Y.~Jiang, S.-T. Xia, Y.~Zhao, and F.~Zheng.
\newblock Hijacking tracker: a powerful adversarial attack on visual tracking.
\newblock In {\em IEEE International Conference on Acoustics, Speech and Signal
  Processing (ICASSP)}, pages 2897 -- 901, 2020.

\bibitem{PIXOR}
B.~{Yang}, W.~{Luo}, and R.~{Urtasun}.
\newblock Pixor: Real-time 3d object detection from point clouds.
\newblock In {\em IEEE/CVF Conference on Computer Vision and Pattern
  Recognition}, pages 7652--7660, 2018.

\bibitem{BaiduAdvTShirt}
D.~Y. {Yang}, J.~{Xiong}, X.~{Li}, X.~{Yan}, J.~{Raiti}, Y.~{Wang}, H.~{Wu},
  and Z.~{Zhong}.
\newblock Building towards "invisible cloak": Robust physical adversarial
  attack on yolo object detector.
\newblock In {\em IEEE Annual Ubiquitous Computing, Electronics Mobile
  Communication Conference (UEMCON)}, pages 368--374, 2018.

\bibitem{yang2023towards}
X.~Yang, C.~Liu, L.~Xu, Y.~Wang, Y.~Dong, N.~Chen, H.~Su, and J.~Zhu.
\newblock Towards effective adversarial textured 3d meshes on physical face
  recognition.
\newblock In {\em Proceedings of the IEEE/CVF Conference on Computer Vision and
  Pattern Recognition}, pages 4119--4128, 2023.

\bibitem{PersonReIDsurvey}
M.~Ye, J.~Shen, G.~Lin, T.~Xiang, L.~Shao, and S.~C.~H. Hoi.
\newblock Deep learning for person re-identification: A survey and outlook.
\newblock {\em IEEE Transactions on Pattern Analysis and Machine Intelligence},
  pages 1--20, 2021.

\bibitem{yin2021adv}
B.~Yin, W.~Wang, T.~Yao, J.~Guo, Z.~Kong, S.~Ding, J.~Li, and C.~Liu.
\newblock Adv-makeup: A new imperceptible and transferable attack on face
  recognition.
\newblock {\em International Joint Conference on Artificial Intelligence
  (IJCAI)}, 2021.

\bibitem{yu2022defending}
Y.~Yu, H.~J. Lee, H.~Lee, and Y.~M. Ro.
\newblock Defending person detection against adversarial patch attack by using
  universal defensive frame.
\newblock {\em IEEE Transactions on Image Processing}, 31:6976--6990, 2022.

\bibitem{yuan2019adversarial}
X.~Yuan, P.~He, Q.~Zhu, and X.~Li.
\newblock Adversarial examples: Attacks and defenses for deep learning.
\newblock {\em IEEE transactions on neural networks and learning systems},
  30(9):2805--2824, 2019.

\bibitem{AdvSurvey2}
X.~{Yuan}, P.~{He}, Q.~{Zhu}, and X.~{Li}.
\newblock Adversarial examples: Attacks and defenses for deep learning.
\newblock {\em IEEE Transactions on Neural Networks and Learning Systems},
  30(9):2805--2824, 2019.

\bibitem{TripletLossReID}
Y.~Yuan, W.~Chen, Y.~Yang, and Z.~Wang.
\newblock In defense of the triplet loss again: Learning robust person
  re-identification with fast approximated triplet loss and label distillation.
\newblock In {\em IEEE/CVF Conference on Computer Vision and Pattern
  Recognition Workshops (CVPRW)}, pages 1454--1463, 2020.

\bibitem{Ocean}
Z.~Zhang, H.~Peng, J.~Fu, B.~Li, and W.~Hu.
\newblock Ocean: Object-aware anchor-free tracking.
\newblock In {\em European Conference on Computer Vision (ECCV)}, pages
  771--787. Springer, 2020.

\bibitem{zheng2022robust}
X.~Zheng, Y.~Fan, B.~Wu, Y.~Zhang, J.~Wang, and S.~Pan.
\newblock Robust physical-world attacks on face recognition.
\newblock {\em Pattern Recognition}, page 109009, 2022.

\bibitem{DigitalAdvPersonReID2}
Y.~{Zheng}, Y.~{Lu}, and S.~{Velipasalar}.
\newblock An effective adversarial attack on person re-identification in video
  surveillance via dispersion reduction.
\newblock {\em IEEE Access}, 8:183891--183902, 2020.

\bibitem{zheng2016discriminatively}
Z.~Zheng, L.~Zheng, and Y.~Yang.
\newblock A discriminatively learned cnn embedding for person
  re-identification.
\newblock {\em ACM Transactions on Multimedia Computing Communications and
  Applications}, 2017.

\bibitem{OSNetPAMI}
K.~Zhou, Y.~Yang, A.~Cavallaro, and T.~Xiang.
\newblock Learning generalisable omni-scale representations for person
  re-identification.
\newblock {\em IEEE Transactions on Pattern Analysis and Machine Intelligence},
  pages 1--1, 2021.

\bibitem{AdvCap}
Z.~Zhou, D.~Tang, X.~Wang, W.~Han, X.~Liu, and K.~Zhang.
\newblock Invisible mask: Practical attacks on face recognition with infrared.
\newblock {\em CoRR}, abs/1803.04683, 2018.

\bibitem{AdvInfrared}
X.~Zhu, Z.~Hu, S.~Huang, J.~Li, and X.~Hu.
\newblock Infrared invisible clothing: Hiding from infrared detectors at
  multiple angles in real world.
\newblock In {\em IEEE/CVF Conference on Computer Vision and Pattern
  Recognition (CVPR)}, pages 13307--13316, 2022.

\bibitem{AdvThermal}
X.~Zhu, X.~Li, J.~Li, Z.~Wang, and X.~Hu.
\newblock Fooling thermal infrared pedestrian detectors in real world using
  small bulbs.
\newblock In {\em AAAI Conference on Artificial Intelligence (AAAI)}, 2021.

\bibitem{AdvMask}
A.~Zolfi, S.~Avidan, Y.~Elovici, and A.~Shabtai.
\newblock Adversarial mask: Real-world adversarial attack against face
  recognition models.
\newblock {\em CoRR}, abs/2111.10759, 2021.

\end{thebibliography}

\newpage
\appendix
\section{Evaluation Methods}
In this section we summarise the evaluation metrics that have been used to evaluate the physical adversarial attack methods proposed for thwarting human detection, identification tracking and action recognition.  

\subsubsection{Evaluation of Attacks on Human Detection}
The change in the detection precision of the object detector has been the primary evaluation metric among the methods \cite{UPC, AdvTshirtFB, BaiduAdvTShirt} that are proposed to thwart human detection. For instance, the authors on \cite{UPC} measure the precision $p_{0.5}$ with respect to the probability of whether the detector can hit the true category under a distinct set of camera viewpoints, brightness and scenes. This can be written as,

\begin{equation}
    p_{0.5} = \frac{1}{|X|} \sum_{v \sim V, b \sim B, s \in S} \{C(x) = y, C(\hat{(x)} = y)\},
\end{equation}
where $x$ is the original image and $\hat{x}$ is the adversarial image. $V, L, S$ denote the sets of camera viewpoints, brightness and scenes, respectively. The detector is denoted by $C$ and $y$ denotes the ground truth label of the object. 
The authors of \cite{AdvTShirtIBM} evaluates attack success rate (ratio of successfully attacked testing frames over the total number of testing frames) and this metric has also been used in \cite{AdvTshirtFB} as an additional metric. In contrast, Naturalistic Physical Adversarial Patch \cite{hu2021naturalistic} method has been evaluated with respect to the reduction in the detection recall.

\subsubsection{Evaluation of Attacks on Human Identification}
Within the literature on physical adversarial attacks on face recognition, The Attack Success Rate (ASR) has been a popular evaluation method \cite{AdvCap, yin2021adv, AdvMask, advGlasses}. ASR can be calculated as,
\begin{equation}
     ASR = \frac{\sum_{i}^{N}1_{\tau}(cos[F(I^i_v), F(I^i_a)] > \tau)}{N} \times 100\%,
\end{equation}
where $1_{\tau}$ denotes the indicator function and for impersonating attack the proportion of examples with similarity larger than ${\tau}$ will be obtained as ASR.
 
In addition to this popular metric some studies have used specific metrics. For instance, the number of  queries from the attacked face recognition model are required for successful attacks is measured in \cite{guo2021meaningful}. Adversarial Glasses \cite{advGlasses} method measures the mean probability assigned to the correct class. The authors of Adversarial Masks \cite{AdvMask} also measures the persistence detection which they calculate as the number of frames in which the attacker was recognised with respect to the number of frames in a sliding window.
 
In contrast, Adversarial Hat \cite{AdvHat} and Adversarial Patches \cite{AdvFacePatches} methods report the reduction in cosine similarity as their evaluation metric. Specifically, in \cite{AdvFacePatches} the authors report the difference in cosine similarity between the target face embedding and the adversarial face. Komkov et al. \cite{AdvHat} have measured the difference between baseline similarity and final similarity as follows: The baseline similarity: Cosine similarity between ground truth embedding and embedding for a photo with a hat;  Final similarity: Cosine similarity between ground truth embedding and embedding for a photo with an adversarial sticker.  

When considering the attacks on person Re-ID the AdvPatterns method proposed in \cite{AdvPersonReID} is the only method to demonstrate physical adversarial attacks against person Re-ID. The authors used the reduction in rank-k re-id accuracy and mean average precision of the Re-ID to quantify the effectiveness of the AdvPatterns method. 

\subsubsection{Evaluation of Attacks on Human Tracking}

 Wiyatno \emph{et al.} proposed the PAT - Adversarial Posters \cite{AdvTracking} method to thwart the human tracking and its effectiness is evaluated using mean-Intersection-Over-Union-difference ($ \mu IOUd$) evaluation metric. This measure can be evaluated as,  
 \begin{equation}
    \begin{split}
    IOU(l_j, \hat{l}_j) &= \frac{A(l_j \cap	 \hat{l}_j)}{A(l_j) + A(\hat{l}_j) - A(l_j \cap	 \hat{l}_j)} \\
    \mu IOUd &= \frac{1}{N-1} \sum_{j \in |2, N|, f_j \in F}  IOU(l_j (f_{j-1}, f_j), \hat{l}_j) \\
    & - \frac{1}{N-1} \sum_{j \in |2, N|, f^*_j \in F^*}  IOU(l_j (f^*_{j-1}, f_j), \hat{l}_j),
    \end{split}
 \end{equation}
where $F^*$ is the adversarial sequence of frames and $F$ is another sequence of frames where adversarial texture is replaced by an inert source texture. $\cap$ denotes the intersection of two bounding boxes and $A(·)$ denotes the area of the bounding box $l$. The ground truth bounding box is denoted by $l$ while $\hat{l}$ denotes the predicted bounding box. 
 
 In a similar line of work, the authors of Universal Physical Attacks on Single Object Tracking \cite{ding2021towards} method have used the success rate which is computed as the Intersection-over-Union (IoU) between the predicted bounding box and the ground truth as one of their evaluation methods. In addition they have used precision, which is measured using the distance between predicted bounding box and the ground truth bounding box in pixels. Furthermore, they have also calculated the normalized precision, which is computed with the Area Under Curve (AUC) between 0 and 0.5. 

\subsubsection{Evaluation of Attacks on Human Action Recognition}
Over-the-Air Adversarial Flickering Attacks \cite{PhysicalAdvAction} is the only adversarial attack method that has demonstrated physical capabilities to thwart human action recognition. The authors have utilised a series of evaluation metrics to evaluate their framework. Specifically, Fooling ratio, Mean Absolute Perturbation per-pixel, and Mean Absolute Temporal-diff Perturbation per-pixel are used. These can be defined as:
\begin{enumerate}
    \item Fooling ratio: the percentage of adversarial videos that are successfully misclassified (higher is better). 
    \item Mean Absolute Perturbation per-pixel: \begin{equation}
        \textrm{MAPer} = \frac{1}{3T}||\delta||_1,
    \end{equation}
    where $\delta$ is the adversarial perturbation and $T$ is the total duration of the video.
    
    \item Mean Absolute Temporal-diff Perturbation per-pixel:
    
    \begin{equation}
        \textrm{MATPer} = \frac{1}{3T}||\frac{\partial \delta}{\partial t}||_1.
    \end{equation}
    
\end{enumerate}

\end{document}